%% file: acl_latex_new.tex
\PassOptionsToPackage{table}{xcolor}
\documentclass[11pt]{article}

\usepackage[final]{acl}
\usepackage{placeins}
\usepackage{needspace}

\usepackage{times}
\usepackage{latexsym}
\usepackage[table]{xcolor}
\usepackage{listings}
\usepackage[T1]{fontenc}

\usepackage[utf8]{inputenc}

\usepackage{microtype}

\usepackage{inconsolata}

\usepackage{graphicx}
\graphicspath{{../figures/}{figures/}{./figures/}{../}}
\usepackage{tabularx}
\usepackage{multirow}
\usepackage{enumitem}
\usepackage{booktabs}
\usepackage{arydshln}
\usepackage{pifont}

\usepackage{amsmath}
\usepackage{amssymb}
\definecolor{memoraBlue}{HTML}{1f4e79}
\definecolor{memoraBlueLight}{HTML}{eef3f7}
\definecolor{memoraBlueMid}{HTML}{5a8ec3}
\definecolor{memoraOrange}{HTML}{ff7f0e}
\definecolor{memoraOrangeDark}{HTML}{cc6600}
\definecolor{memoraOrangeMid}{HTML}{ffb066}
\definecolor{graph2dPurple}{HTML}{7b3fa0}
\definecolor{socraticGrey}{HTML}{7f7f7f}
\definecolor{nomemCoral}{HTML}{e08e8e}  
\definecolor{cogsciOrange}{HTML}{a05a3a}
\definecolor{cogsciTint}{HTML}{f6efee}
\definecolor{parametricTint}{HTML}{F7E8E4}
\definecolor{rawBaselineTint}{HTML}{E5E7EB}
\definecolor{processedBaselineTint}{HTML}{D8F0E7}
\definecolor{memoraResultTint}{HTML}{FCE6CC}  

\title{MEMORA: Embodied Action Memory from Egocentric Videos for Reasoning and Planning}

\author{
\textbf{Zihao Yu}\textsuperscript{1}
\quad \quad \quad \quad \quad
\textbf{Xiu Yuan}\textsuperscript{1} \quad \quad \quad \quad \quad
\textbf{Chongjie Zhang}\textsuperscript{1}
\\
\textsuperscript{1}Washington University in St. Louis \\
\texttt{\{yu.zihao,xiu,chongjie\}@wustl.edu}
}

\begin{document}
\maketitle
\begin{abstract}
Embodied agents accumulate experience over time. We study how accumulated experience can be formed into persistent memory for future reasoning and action. We formulate Embodied Action Memory (EAM) as the capability to form and use memory over embodied experience, together with the persistent memory state produced by that process. We introduce MEMORA, a framework that instantiates EAM through a formation--consolidation--retrieval lifecycle and a multi-store world-memory architecture. MEMORA organizes experience into participant-specific Environment, Entity, Activity, and Inferred Knowledge stores: online editing revises memory as new evidence arrives, while offline consolidation abstracts repeated experience into reusable routines, habits, and preferences. We evaluate MEMORA with MEMORA-Bench, a 45-hour egocentric-video suite that measures both retrospective memory faithfulness and prospective memory-grounded planning. Across four open-weight answer models, MEMORA achieves the strongest aggregate planning performance among the evaluated memory interfaces, with its largest gains on out-of-distribution planning. On these tasks, MEMORA improves Robot-Grounded Plan score by up to 16.6\%, suggesting that memory formed and consolidated across experience can support planning for new goals beyond directly observed episodes. A physical-robot demonstration further shows that memory formed solely from human egocentric video can ground high-level robot plans in participant-specific objects and preferences.
\end{abstract}

\input{sections/memora_types}
\input{sections_new/Introduction}
\input{sections_new/Related_Works}
\input{sections_new/Problem_Formulation}
\input{sections_new/Method}
\input{sections_new/Experiments}
\input{sections_new/Discussion}
\input{sections_new/Conclusion}


\input{sections_new/Limitations}
\input{sections_new/Ethical_Considerations}
\FloatBarrier
\bibliography{references}

\newpage
\input{sections_new/Appendix}

\end{document}

%% file: sections/memora_types.tex
\providecommand{\ELoc}{\textsc{ELoc}}
\providecommand{\EOrder}{\textsc{EOrder}}
\providecommand{\ERecall}{\textsc{ERecall}}
\providecommand{\SPref}{\textsc{SPref}}
\providecommand{\SHabit}{\textsc{SHabit}}
\providecommand{\SRoutine}{\textsc{SRoutine}}

\providecolor{envGreen}{HTML}{2E7D5B}
\providecolor{entBlue}{HTML}{2F5AA8}
\providecolor{actOrange}{HTML}{B75C1B}
\providecolor{infPurple}{HTML}{7B3FA0}
\providecolor{envTint}{HTML}{EEF7F2}
\providecolor{entTint}{HTML}{EEF3FB}
\providecolor{actTint}{HTML}{FFF3EA}
\providecolor{infTint}{HTML}{F4EEF8}

\providecommand{\EnvMem}{\textcolor{envGreen}{\textbf{Environment Memory}}}
\providecommand{\EntMem}{\textcolor{entBlue}{\textbf{Entity Memory}}}
\providecommand{\ActMem}{\textcolor{actOrange}{\textbf{Activity Memory}}}
\providecommand{\InfMem}{\textcolor{infPurple}{\textbf{Inferred Knowledge}}}

\providecommand{\EnvTok}[1]{\textcolor{envGreen}{\texttt{#1}}}
\providecommand{\EntTok}[1]{\textcolor{entBlue}{\texttt{#1}}}
\providecommand{\ActTok}[1]{\textcolor{actOrange}{\texttt{#1}}}
\providecommand{\InfTok}[1]{\textcolor{infPurple}{\texttt{#1}}}

\providecommand{\Menv}[1][]{\ensuremath{\textcolor{envGreen}{\mathcal{M}_{#1}^{\mathrm{env}}}}}
\providecommand{\Ment}[1][]{\ensuremath{\textcolor{entBlue}{\mathcal{M}_{#1}^{\mathrm{ent}}}}}
\providecommand{\Mact}[1][]{\ensuremath{\textcolor{actOrange}{\mathcal{M}_{#1}^{\mathrm{act}}}}}
\providecommand{\Minf}[1][]{\ensuremath{\textcolor{infPurple}{\mathcal{M}_{#1}^{\mathrm{inf}}}}}

\providecommand{\Zenv}[1][]{\ensuremath{\textcolor{envGreen}{z_{#1}^{\mathrm{env}}}}}
\providecommand{\Zent}[1][]{\ensuremath{\textcolor{entBlue}{z_{#1}^{\mathrm{ent}}}}}
\providecommand{\Zact}[1][]{\ensuremath{\textcolor{actOrange}{z_{#1}^{\mathrm{act}}}}}

%% file: sections_new/Introduction.tex
\section{Introduction}

Human intelligence is not only a response to the present; it is shaped by experience accumulated across time.
Memory turns such experience into a persistent basis for future reasoning and action~\citep{tulving1972episodic,mcclelland1995there,stickgold2005sleep}.
We study embodied agents through the same lens: their later decisions can depend not only on the current observation, but also on memory formed from prior embodied experience.

\begin{figure*}[!t]
\centering
\includegraphics[width=0.98\textwidth]{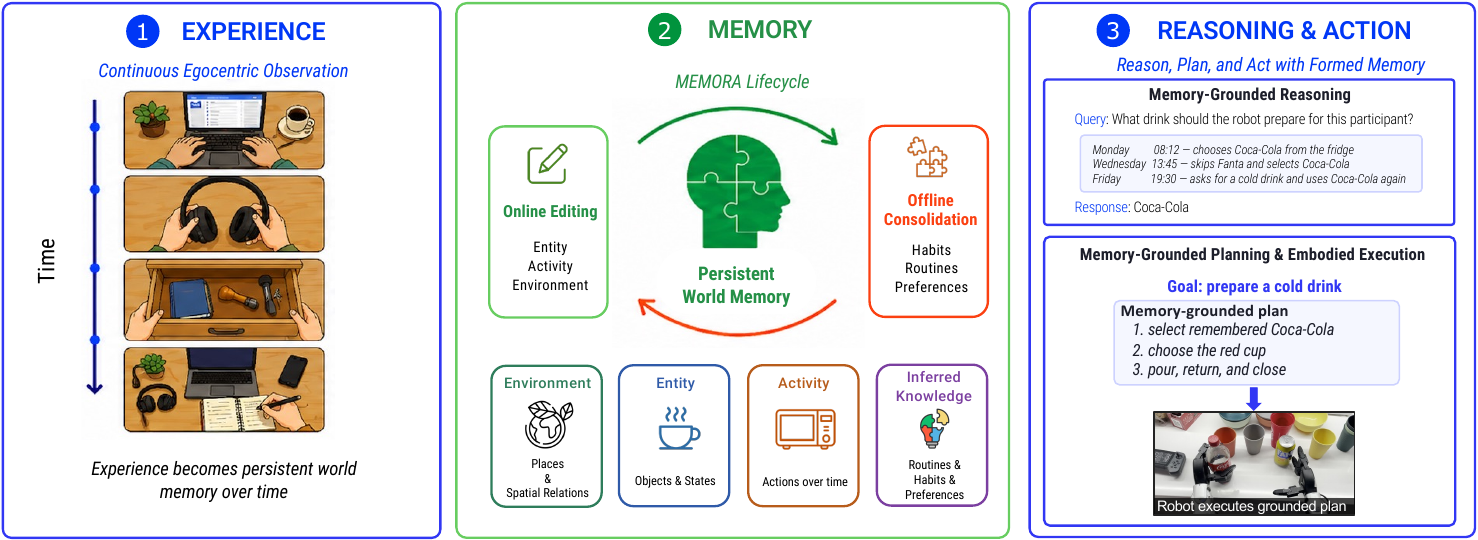}
\caption{\textbf{MEMORA studies embodied memory as a lifecycle from experience to action.}
An agent observes continuous egocentric experience, forms typed memory online, and consolidates repeated experience into reusable routines, habits, and preferences.
The resulting persistent world memory supports memory-grounded reasoning, planning, and physical robot execution from accumulated experience.}
\label{fig:memora-teaser}
\end{figure*}

A long tradition in cognitive and computational models treats memory not as a passive archive, but as a staged process of encoding, consolidation, and retrieval, often across functionally distinct forms of memory~\citep{Atkinson1968,tulving1972episodic,mcclelland1995there}.
This lifecycle view has shaped classical cognitive architectures~\citep{anderson1983actr,laird2012soar}, neural agents with differentiable read--write memory~\citep{weston2015memorynetworks,graves2016hybrid}, and recent LLM-agent systems that explicitly model memory formation and consolidation~\citep{ma2026brainmembraininspiredevolvingmemory,mody2026cranimemcranialinspiredgated,ye2026autodreamerlearningofflinememory,du2026memoryautonomousllmagentsmechanisms}.
Embodied settings complicate this lifecycle because a continuous perception stream interleaves different kinds of information that persist, change, and become useful in different ways, including spatial relations among places and objects, entities tracked across changing states, temporally ordered actions, and regularities that emerge only across episodes.
Forming memory from this stream involves revising earlier content as evidence arrives and consolidating repeated experience while preserving distinctions that later reasoning and action may draw upon.
We call this capability \textbf{Embodied Action Memory (EAM)}: the ability to form and use memory over embodied experience, together with the persistent memory state produced by that process.

We therefore study embodied memory as a formation--consolidation lifecycle (Figure~\ref{fig:memora-teaser}) that maintains changing experience online and abstracts repeated experience offline.
Egocentric video provides a longitudinal first-person record of how a participant interacts with familiar environments.
Whereas prior work often uses this signal for imitation or cross-embodiment policy learning~\citep{kareer2024egomimicscalingimitationlearning,punamiya2026egoverseegocentrichumandataset,zheng2026egoscalescalingdexterousmanipulation}, MEMORA uses it to form participant-specific persistent memory.

We propose \textbf{MEMORA} (Figure~\ref{fig:memora-teaser}), an embodied memory framework for forming, consolidating, and using persistent world memory from continuous egocentric experience.
To preserve distinctions among these forms of embodied information, MEMORA organizes memory into four typed stores: \EnvMem{} for spatial context, \EntMem{} for object identity and state, \ActMem{} for temporally ordered action traces, and \InfMem{} for regularities consolidated across repeated experience.
During online formation, an active Memory Editor applies LLM judgment to entity observations, selecting \textsc{Add}, \textsc{Update}, \textsc{Delete}, or \textsc{Noop} and yielding a median $18\times$ compression while preserving state histories.
After offline consolidation, typed retrieval at use time separates reusable procedure from physical grounding, letting MEMORA compose reusable routines with participant-specific entities into grounded language plans, even for goals never directly observed.

To evaluate this view, we introduce \textbf{MEMORA-Bench}, a 45-hour egocentric-video evaluation suite built from an EPIC-KITCHENS extension split~\citep{damen2022epic}.
It pairs \textbf{MEMORA-Embodied Memory Assessment}, which tests whether constructed memory is faithful to prior experience, with \textbf{MEMORA-Planning}, which tests whether memory supports future action.
Planning includes \textbf{Replay} tasks with previously observed workflows and \textbf{Generalize} tasks requiring transfer, composition, or new goals.
Across four open-weight backbones, MEMORA improves most on out-of-distribution planning, reaching a $+16.4\%$ relative gain in Robot-Grounded Plan score (RGP) on Gemma-4-31B-it.
A physical-robot demonstration further shows that memory formed solely from human egocentric video can ground high-level robot plans in participant-specific objects and preferences.
MEMORA therefore complements the robot's execution system: memory helps determine and ground what to do, while an embodiment-specific controller carries out the resulting plan.

\noindent\textbf{Contributions.}
(1) A formulation of \textbf{Embodied Action Memory (EAM)} (\S\ref{sec:problem}) as the formation and use of persistent memory over embodied experience, separating how memory is formed from how it later supports reasoning and action.
(2) \textbf{MEMORA} (\S\ref{sec:method}), an embodied memory framework that instantiates the formation--consolidation--retrieval lifecycle on real egocentric video through four typed stores---\EnvMem{}, \EntMem{}, \ActMem{}, and \InfMem{}---specialized for different forms of embodied information and their update dynamics, together with write-time active editing (median $18\times$ compression), offline consolidation, and typed retrieval at use time.
(3) \textbf{MEMORA-Bench}, an evaluation suite for EAM over 45 hours of egocentric video, pairing memory faithfulness (\textbf{MEMORA-Embodied Memory Assessment}) with memory-grounded action (\textbf{MEMORA-Planning}).
(4) A \textbf{systematic empirical study} of complete memory interfaces across four answer backbones, with within-family contrasts isolating online editing and offline consolidation; the largest planning gains occur out of distribution.

%% file: sections_new/Related_Works.tex
\section{Related Work}

\paragraph{Memory formation.}
Agent memory has been modeled as parametric storage in model weights~\citep{zhang2024surveymemorymechanismlarge} or as explicit stores retrieved as documents, embeddings, summaries, graphs, or edited records~\citep{lewis2021retrievalaugmentedgenerationknowledgeintensivenlp,zhong2023memorybankenhancinglargelanguage,packer2024memgptllmsoperatingsystems,park2023generativeagentsinteractivesimulacra,chhikara2025mem0buildingproductionreadyai,yan2026memoryr1enhancinglargelanguage}. M3-Agent, for example, organizes multimodal video experience as a graph for long-term remembering and reasoning~\citep{long2025seeinglisteningrememberingreasoning}.
Recent work introduces explicit encoding and consolidation over text or simulated experience~\citep{ma2026brainmembraininspiredevolvingmemory,mody2026cranimemcranialinspiredgated,ye2026autodreamerlearningofflinememory,du2026memoryautonomousllmagentsmechanisms}, reflecting cognitive distinctions between episodic, semantic, and consolidative memory~\citep{tulving1972episodic,mcclelland1995there,stickgold2005sleep,nader2003memory}.
MEMORA brings this direction to continuous egocentric video, where memory is revised online and consolidated into persistent state.
MEMORA thus differs not only in how past experience is retrieved, but also in treating memory formation itself as part of the embodied problem.

\paragraph{Embodied memory benchmarks.}
Long-video QA benchmarks evaluate comprehension of video supplied at query time~\citep{fu2025videommefirstevercomprehensiveevaluation,mangalam2023egoschemadiagnosticbenchmarklongform,li2024mvbenchcomprehensivemultimodalvideo}, while conversational-memory benchmarks isolate memory in text-only interactions~\citep{maharana2024evaluatinglongtermconversationalmemory,wu2025longmemevalbenchmarkingchatassistants}.
Egocentric datasets and instruction-following suites provide rich embodied activity, but these resources do not evaluate whether an agent can form memory across sessions and later use it without receiving the relevant video at test time~\citep{damen2022epic,grauman2022ego4dworld3000hours,shridhar2020alfredbenchmarkinterpretinggrounded,padmakumar2021teachtaskdrivenembodiedagents}.
MEMORA-Bench targets this experience-to-memory setting: questions and plans are grounded in a participant's accumulated world and require evidence about preferences, routines, object choices, and spatial context across sessions.

\paragraph{Memory-grounded embodied action.}
Language-conditioned planning uses prompted LLMs, code, or skill libraries to connect perception, reasoning, and action~\citep{zeng2022socraticmodelscomposingzeroshot,ahn2022icanisay,liang2023codepolicieslanguagemodel,huang2022innermonologueembodiedreasoning}.
Other embodied agents maintain spatial memory or learn reactive vision--language--action policies for grounding and control~\citep{yang20253dmem3dscenememory,zheng2026spatialmemmetricalignedlonghorizonvideo,lei2026robomemorybraininspiredmultimemoryagentic,brohan2023rt2visionlanguageactionmodelstransfer,black2026pi0visionlanguageactionflowmodel}.
Egocentric video also supports imitation, manipulation, and cross-embodiment learning~\citep{kareer2024egomimicscalingimitationlearning,punamiya2026egoverseegocentrichumandataset,zheng2026egoscalescalingdexterousmanipulation}; MEMORA instead uses the same longitudinal signal to form persistent participant memory for downstream reasoning and action. It supplies participant-specific context to downstream planners and policies, complementing spatial maps and visuomotor capabilities without replacing their metric grounding or control machinery.

%% file: sections_new/Problem_Formulation.tex
\section{Problem Formulation: Embodied Action Memory}
\label{sec:problem}

Embodied intelligence unfolds over time: observations arrive before their future relevance is known.
\textbf{Embodied Action Memory (EAM)} studies how temporally extended embodied experience can be formed into a persistent, evolving memory state that can later support reasoning and action.
Let $o_t$ be the observation at time $t$, $\mathcal{M}_t$ the memory state after processing experience through $t$, and $x_\tau$ a later question, goal, or decision context.
We model experience formation and later use as
\begin{equation}
\mathcal{M}_t = F(\mathcal{M}_{t-1}, o_t), \qquad
d_\tau = G(\mathcal{M}_\tau, x_\tau),
\end{equation}
\noindent Here, $F$ maps incoming embodied experience to an evolving memory state, while $G$ uses the memory available at time $\tau$ together with a later decision context to produce an answer, plan, or other action-relevant decision.
This formulation separates how memory is formed from how it is later used, without prescribing a single memory representation or decision interface.
In this work, we instantiate EAM with longitudinal egocentric experience, typed memory stores, and memory-grounded reasoning and planning.

%% file: sections_new/Method.tex
\section{Method}
\label{sec:method}
\label{sec:agent}


We present MEMORA along the EAM lifecycle: \S\ref{sec:overview} describes online formation and offline consolidation, \S\ref{sec:memory-retrieval-planning} exposes the formed memory for retrieval and planning, and \S\ref{sec:bench} introduces MEMORA-Bench.
Online encoding and active editing instantiate the online formation function $F$, while offline consolidation abstracts repeated experience into reusable knowledge~\citep{ma2026brainmembraininspiredevolvingmemory,mody2026cranimemcranialinspiredgated,ye2026autodreamerlearningofflinememory}. To address heterogeneous forms of information and their update dynamics in embodied experience, MEMORA combines four typed stores with \emph{write-time active editing}.

\subsection{Memory Formation: Encoding, Editing, and Consolidation}
\label{sec:overview}
\label{sec:workflow}
\label{sec:layered-memory}
\label{sec:consolidation}

MEMORA instantiates the EAM formulation in Section~\ref{sec:problem} with four typed stores: \EnvMem{}, \EntMem{}, \ActMem{}, and \InfMem{} (Table~\ref{tab:eam-memory-stores}).

The design distinguishes four forms of embodied information: recurring spatial context, persistent object identities and changing states, temporally ordered action evidence, and routines or preferences that emerge across repeated experience.
A single append-only store flattens these distinctions, while a single graph-structured store can blur them.
Motivated by both the multi-store view of memory~\citep{Atkinson1968} and the functional distinction between episodic traces and semantic generalizations~\citep{tulving1972episodic,mcclelland1995there}, MEMORA therefore maintains a \emph{multi-store memory state} $\mathcal{M}_t = (\Menv[t],\, \Ment[t],\, \Mact[t],\, \Minf{})$ (Table~\ref{tab:eam-memory-stores}), with stores specialized for these forms of information and their update dynamics; this typing is not an additional supervision signal but determines the update rule each observation receives, concentrating LLM judgment where semantic revision is necessary and using deterministic operations where the rule is stable.

\begin{table*}[t]
\centering
\footnotesize
\setlength{\tabcolsep}{3pt}
\renewcommand{\arraystretch}{1.14}
\begin{tabularx}{0.99\textwidth}{@{}>{\raggedright\arraybackslash}p{0.10\textwidth}
>{\raggedright\arraybackslash}p{0.22\textwidth}
>{\raggedright\arraybackslash}p{0.33\textwidth}
>{\raggedright\arraybackslash}X@{}}
\toprule
\textbf{Store} & \textbf{Record unit} & \textbf{Example} & \textbf{Read-time role} \\
\midrule
\rowcolor{envTint}
\EnvMem{} &
Place records with layout descriptions, named zones, and spatial relations. &
\EnvTok{sink\_area}: zones $\{$\EnvTok{sink}, \EnvTok{dish\_rack}, \EnvTok{stove}, \EnvTok{faucet}$\}$; \EnvTok{dish\_rack} to the right of \EnvTok{sink}. &
Ground actions and objects in remembered places. \\
\addlinespace[0.2em]
\rowcolor{entTint}
\EntMem{} &
Object records with visual attributes, location, current state, and \texttt{state\_history}. &
\EntTok{plate\_with\_food\_residue} (white ceramic, dirty): \textsc{Upd} \emph{idle@countertop} $\to$ \emph{being\_washed@sink}; prior state retained in \texttt{state\_history}. &
Preserve object identity through changing state. \\
\addlinespace[0.2em]
\rowcolor{actTint}
\ActMem{} &
Segment-aligned action records (10\,s segments) with summary, narrative, and \texttt{action\_breakdown}. &
\ActTok{10--20\,s} ``picks up pan and washes it in the sink''; \texttt{action\_breakdown}: 10--12\,s picks up pan, 12--14\,s moves to sink, 14--20\,s washes pan. &
Recover event evidence, temporal order, and procedure traces. \\
\addlinespace[0.2em]
\rowcolor{infTint}
\InfMem{} &
Routines, habits, preferences, and workflow patterns linked to supporting episodes. &
Workflow pattern \emph{``dirty dishes and used containers are washed in the sink, then placed on the countertop''} (supports: \EntTok{pan\_black}, \EntTok{plate\_with\_food\_residue}; conf.\ 0.95). &
Support cross-event generalization and planning routines. \\
\bottomrule
\end{tabularx}
\caption{\textbf{Typed memory stores in MEMORA.}
Examples are drawn from a real kitchen recording (participant P09).}
\label{tab:eam-memory-stores}
\end{table*}

During \emph{online memory formation}, the write path first encodes the current segment with working context, then revises the persistent memory state via \textsc{Edit}, which instantiates the formation function $F$ from Section~\ref{sec:problem}:
\begin{equation}
\begin{aligned}
(z_t, c_t) &= \phi(v_t, c_{t-1}), \\
\mathcal{M}_t &= \textsc{Edit}(\mathcal{M}_{t-1}, z_t).
\end{aligned}
\end{equation}
Here $\phi$ is the \emph{Segment Encoder}, a multimodal model that converts each 10\,s video segment $v_t$ into text-structured observations $z_t = (\Zenv[t], \Zent[t], \Zact[t])$, and $c_t$ carries a short working-memory summary across segment boundaries.
Inside \textsc{Edit}, following recent active memory-editing systems~\citep{chhikara2025mem0buildingproductionreadyai,yan2026memoryr1enhancinglargelanguage}, the \emph{Memory Editor} selects $\delta_j \in \Delta = \{\textsc{Add},\, \textsc{Upd},\, \textsc{Del},\, \textsc{Noop}\}$ per object observation while preserving previous states in \texttt{state\_history}; \EnvMem{} and \ActMem{} are updated deterministically (merge by detected place; append by segment), and \InfMem{} is populated offline by consolidation~\citep{stickgold2005sleep} over repeated entity and activity evidence.
Later evidence can update histories or delete inconsistent identifications, so editing revises earlier memory rather than only compressing it.
JSON-style records, entity-edit diagnostics, and memory-store composition are in Appendix~\ref{app:kb-construction}.

\begin{figure*}[t]
\centering
\includegraphics[width=0.98\textwidth]{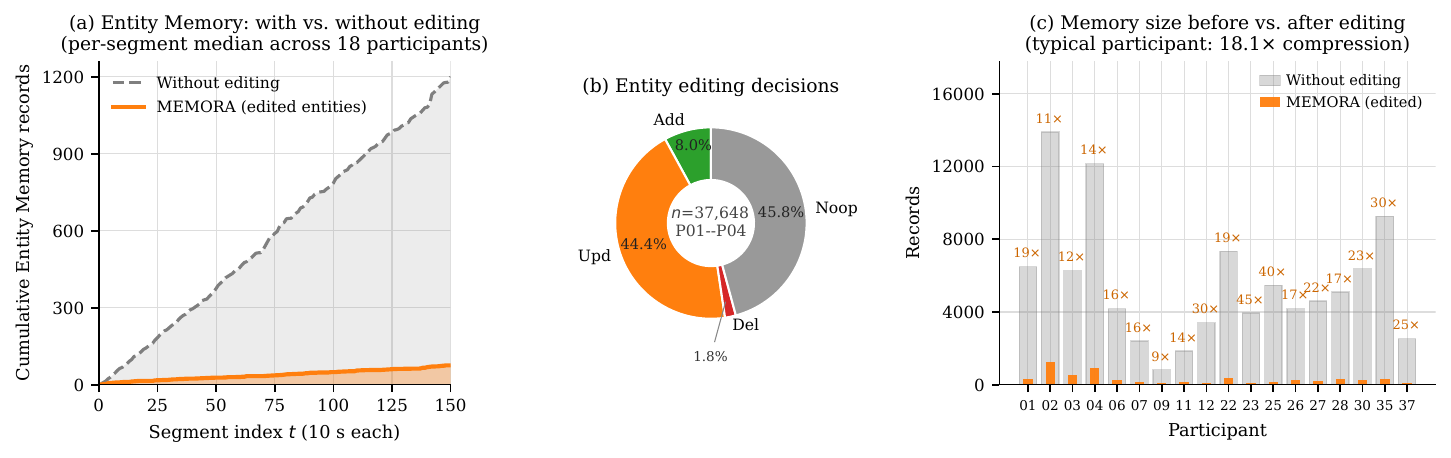}
\caption{\textbf{Active memory formation in Entity Memory (18 participants).}
\textbf{(a)}~Cumulative records with and without the editor (per-segment median, first 150 segments).
\textbf{(b)}~Editor decision mix ($n{=}37{,}648$, P01--P04): \textsc{Noop} ($45.8\%$) and \textsc{Upd} ($44.4\%$) dominate \textsc{Add} ($8.0\%$) with rare \textsc{Del} ($1.8\%$) --- the editor mostly \emph{maintains} rather than \emph{expands}.
\textbf{(c)}~Per-participant size before vs.\ after editing; median reduction \textbf{$\approx 18\times$}, minimum $9\times$.
Per-PID counts and editor ablations are in Appendices~\ref{app:editor-activity}, \ref{app:ablation-a1-shape}.}
\label{fig:memory-maintenance}
\end{figure*}

MEMORA thus maintains memory selectively rather than storing every observation as an independent record; Figure~\ref{fig:memory-maintenance} quantifies a median $18\times$ reduction relative to the unedited observation stream on the 18-participant EPIC-Kitchens corpus (state histories preserved).

\subsection{Memory-Guided Retrieval and Planning}
\label{sec:memory-retrieval-planning}

Once memory has been formed, MEMORA exposes it through two uses of the
decision function $G$: retrospective assessment and prospective planning.
\emph{Retrieval} is an evaluation interface: given a question $q$ and the
completed memory bank $\mathcal{M}^{(p)}$ for its target participant $p$, the
agent is asked to recover participant-specific evidence and select an answer $a$ only
when that memory supports it.
\emph{Planning} is an application interface: given a goal $g$, the agent is asked to produce a grounded plan $\pi = ((a_1, e_1, \ell_1), \ldots, (a_n, e_n, \ell_n))$ from the memory state, where each step specifies an action $a_k$, an entity $e_k$, and a location $\ell_k$ drawn from memory.

Both evaluations use an iterative ReAct loop~\citep{yao2023reactsynergizingreasoningacting}, with read interfaces matched to each memory representation. Each condition exposes a common search entry point, while Graph-2D and MEMORA additionally provide representation-specific temporal or typed-store helpers; MEMORA's planning interface also exposes consolidated routines and preferences. Cross-family results compare complete memory interfaces; within-family controls are specified in \S\ref{sec:setup}. At read time, planners bind procedural scaffolds from \InfMem{}/\ActMem{} (``how to do it'') to participant-specific entities, states, and locations from \EntMem{}/\EnvMem{} (``which object and where''), especially on Generalize, where planning composes evidence across memories.
The released tool surface, participant routing, helper semantics, retrieval settings, prompts, and iteration budgets are specified in Appendix~\ref{app:impl}; memory-bank construction is detailed in Appendix~\ref{app:kb-construction}, and an end-to-end Planning trace appears in Appendix~\ref{app:planning}.


\subsection{MEMORA-Bench: An Evaluation Suite for Embodied Action Memory}
\label{sec:bench}

We evaluate embodied action memory formed from continuous embodied experience along two complementary dimensions: \emph{faithfulness to past experience} and \emph{support for future action} --- a combination not jointly evaluated by these benchmark families~\citep{fu2025videommefirstevercomprehensiveevaluation,mangalam2023egoschemadiagnosticbenchmarklongform,maharana2024evaluatinglongtermconversationalmemory,wu2025longmemevalbenchmarkingchatassistants}.
MEMORA-Bench therefore pairs two interfaces over the same participant-specific memory state, formed from 45 hours of EPIC-KITCHENS extension video across 18 participants.
\textbf{MEMORA-Embodied Memory Assessment} (\textbf{EAM-QA}) probes retrospective faithfulness with multiple-choice questions over four EAM types (\textsc{SPref}: preferences; \textsc{SHabit}: habits; \textsc{SRoutine}: ordered procedures; \textsc{ERecall}: within-video recall). It includes answerable items and cross-participant E-correct controls, all with an explicit \emph{information not available} option E to separate missing target-participant evidence from forced guessing.
\textbf{MEMORA-Planning} probes prospective utility by producing grounded plans from participant-specific goals, split into \textbf{Replay} (observed workflows) and \textbf{Generalize} (transfer, composition, or fully novel goals).
Both interfaces are automatically constructed without inspecting any memory system's output, with ground truth derived from the participant's recordings rather than MEMORA's memory.
Released benchmark items are static artifacts and the MEMORA pipeline itself runs entirely on open-weight models (Appendix~\ref{app:ablation-enrichment}), so reproducing the headline results does not require closed-weight LLM access.
Construction and quality-control details are in Appendices~\ref{app:bench-quality} and~\ref{app:planning-denovo}; dataset, conditions, and metrics follow in \S\ref{sec:setup}.

%% file: sections_new/Experiments.tex
\section{Experiments}
\label{sec:experiments}



\subsection{Experimental Setup}
\label{sec:setup}

\paragraph{Evaluation logic.}
The experiments separate three questions: whether online editing improves the memory written from perception, whether offline consolidation adds reusable cross-event knowledge, and whether the resulting complete memory interfaces support retrospective assessment and prospective planning. Controlled raw/processed and episodic/full pairs address the first two questions; cross-representation comparisons test the system-level value of each representation together with its read interface, and the robot study checks whether the resulting language plans can support physical execution.

\paragraph{Data and protocols.}
We evaluate 45 hours from the EPIC-KITCHENS-100 extension~\citep{damen2022epic}, comprising 201 egocentric videos from all $18$ participants for whom consolidated \InfMem{} memory banks are built ($8$--$15$ sessions each). The recordings span participant-specific kitchens, objects, and recurring routines; memory construction and evaluation remain participant-scoped throughout.
\textbf{EAM-QA} contains $N{=}2{,}763$ multiple-choice items---$2{,}212$
answerable questions and $551$ E-correct controls---with four content options
A--D plus an explicit information-not-available option E, across the four EAM
types defined in \S\ref{sec:bench}.
\textbf{MEMORA-Planning} contains $207$ \textbf{Replay} (in-distribution) and $153$ \textbf{Generalize} (out-of-distribution) tasks; tier-stratified subsets are reported in Appendix~\ref{app:planning-denovo}.

\paragraph{Conditions and metrics.}
We compare seven conditions organized into controlled within-family contrasts: \textbf{Parametric} (No Memory), \textbf{Flat-1D}, a Socratic-style chronological text interface~\citep{zeng2022socraticmodelscomposingzeroshot}, and \textbf{Graph-2D}, an entity-relation memory adapted from M3-Agent~\citep{long2025seeinglisteningrememberingreasoning}, in both \emph{raw} and MEMORA-edited (\emph{processed}) variants, \textbf{MEMORA-Episodic} (no offline consolidation), and full \textbf{MEMORA}.
All conditions fix the perception stream, answer backbone, and agent budget; each representation uses its intended read interface. Raw/processed pairs within Flat-1D and Graph-2D, and the MEMORA-Episodic/full pair, keep that interface fixed and isolate online editing or offline consolidation. Tasks are generated by pipelines that never inspect memory-system outputs.
EAM-QA reports a conditional diagnostic on questions missed by Parametric for
which MEMORA selects a content option A--D rather than option E; all conditions
are evaluated on this same subset. This operational filter does not condition
on MEMORA being correct and is reported alongside, not instead of, full-benchmark
accuracy and E-selection behavior in Appendix~\ref{app:eamqa-evaluation-views}.
Planning keeps Parametric as a real prior baseline and reports \textbf{Robot-Grounded Plan score (RGP)}, the mean of the condition-level \textbf{OrderExec}, \textbf{KeyObj}, and \textbf{PrefAdh} scores; auxiliary planning metrics are in Appendices~\ref{app:planning-full} and~\ref{app:planning-grounding-diag}.
Unless noted, perception and editing use Qwen2.5-Omni-7B and Qwen3-30B-A3B-Instruct-2507; the four answer backbones and implementation details are listed in Appendix~\ref{app:impl}.

\begin{table*}[t]
\centering
\caption{\textbf{From memory formation to planning use.}
Three cases link formed memory to planning; full traces are in Appendix Table~\ref{tab:benchmark-examples-expanded}.}
\label{tab:benchmark-examples}
\footnotesize
\setlength{\tabcolsep}{2pt}
\renewcommand{\arraystretch}{0.96}
\begin{tabularx}{\textwidth}{@{}>{\raggedright\arraybackslash}p{0.23\textwidth}
>{\raggedright\arraybackslash}X
 >{\raggedright\arraybackslash}p{0.25\textwidth}@{}}
\toprule
\textbf{Capability tested} & \textbf{Real MEMORA evidence/output} & \textbf{Why it matters} \\
\midrule
\cellcolor{entTint!10}
\textbf{Track one object through state changes.}
A dirty plate appears at 40\,s, 50\,s, and 540\,s. &
\cellcolor{entTint!10}
\textbf{Entity Memory:} \texttt{plate\_with\_food\_residue};
\texttt{idle@countertop} $\rightarrow$ \texttt{being\_used@countertop}, preserving identity and state history. &
\cellcolor{entTint!10}
Planning needs the current state rather than duplicate sightings; editing reduces entity records by median $\approx18\times$ (Figure~\ref{fig:memory-maintenance}). \\
\cellcolor{memoraResultTint}
\textbf{Ground remembered evidence in action.}
Goal: help P03 clean the washing-up bowl. &
\cellcolor{memoraResultTint}
\textbf{Typed retrieval:} retrieve ``wash bowl''; ground the black bowl, green sponge, chrome faucet, and metal drying rack.
\textbf{Plan:} wash with the sponge under the faucet, then place on the rack. &
\cellcolor{memoraResultTint}
The plan binds a reusable routine to participant-specific objects and places; No Memory and Flat-1D retrieve generic or mismatched evidence. \\
\cellcolor{processedBaselineTint}
\textbf{Consolidate repeated experience.}
\textsc{SPref}: does P04 prefer a spatula or fork for eggs? Gold: spatula. &
\cellcolor{processedBaselineTint}
\textbf{Evidence:} P04 repeatedly uses a wooden spatula for yellow mixtures and eggs (P04\_116, turns 28 and 39).
\textbf{Answer:} spatula. &
\cellcolor{processedBaselineTint}
Repeated evidence captures a stable preference; No Memory, Flat-1D, and Graph-2D return information unavailable. \\
\bottomrule
\end{tabularx}
\end{table*}


\subsection{MEMORA-Embodied Memory Assessment Results}
\label{sec:main-results}
\label{sec:eamqa-evaluation}
\label{sec:memorization-results}

\begin{table*}[t]
\centering
\caption{%
\textbf{MEMORA-Embodied Memory Assessment per-type accuracy (\%) on Gemma-4-31B-it} ($N{=}1{,}122$ EAM-QA questions under the parametric-missed, MEMORA-committed diagnostic over the full $18$ participants).
\textsc{SPref}/\textsc{SHabit}/\textsc{SRoutine} aggregate evidence across sessions; \textsc{ERecall} is within a single video.
$\Delta$ is MEMORA minus MEMORA-Episodic; \textbf{bold} marks the best per column.
Parametric is omitted because this view focuses on questions it answered incorrectly; see Appendix~\ref{app:eamqa-evaluation-views}.
Per-backbone reproductions are in Appendix Table~\ref{tab:eamqa-per-backbone}.
}
\label{tab:memorization-main}
\label{tab:eamqa-main}
\small
\setlength{\tabcolsep}{6pt}
\renewcommand{\arraystretch}{1.12}
\begin{tabular}{@{}l ccc c c@{}}
\toprule
                                                      & \multicolumn{3}{c}{\textbf{Cross-video}}                                                           & \textbf{Single-video}      & \\
\cmidrule(lr){2-4}\cmidrule(lr){5-5}
\textbf{Memory condition}                             & \textbf{\textsc{SPref}}    & \textbf{\textsc{SHabit}}    & \textbf{\textsc{SRoutine}}     & \textbf{\textsc{ERecall}}    & \textbf{Overall}     \\
\midrule
\rowcolor{rawBaselineTint}
\multicolumn{6}{@{}l}{\textit{Baselines built directly from the perception stream}} \\
Flat-1D~\citep{zeng2022socraticmodelscomposingzeroshot} (raw)                & 63.2                       & 36.8                        & 53.8                           & 51.2                         & 52.0                 \\
Graph-2D~\citep{long2025seeinglisteningrememberingreasoning} (raw)                 & 50.9                       & 21.4                        & 37.1                           & 55.4                         & 40.0                 \\
\addlinespace[0.15em]
\rowcolor{processedBaselineTint}
\multicolumn{6}{@{}l}{\textit{Baselines rebuilt after MEMORA's online editing}} \\
Flat-1D~\citep{zeng2022socraticmodelscomposingzeroshot} (processed)          & 62.4                       & 41.4                        & 56.8                           & 53.0                         & 54.0                 \\
Graph-2D~\citep{long2025seeinglisteningrememberingreasoning} (processed)           & 57.1                       & 27.0                        & 48.6                           & 57.1                         & 47.0                 \\
\midrule
\rowcolor{memoraResultTint}
\multicolumn{6}{@{}l}{\textit{MEMORA memory stores}} \\
\textbf{MEMORA-Episodic} (no consol.)                 & 67.4                       & 47.0                        & 56.8                           & 73.8                         & 60.1                 \\
\textbf{MEMORA} (full)                                & \textbf{76.8}              & \textbf{60.7}               & \textbf{81.5}                  & \textbf{79.8}                & \textbf{74.5}        \\
\midrule
$\Delta$ from consolidation (points)                      & \textbf{+9.4}              & \textbf{+13.7}              & \textbf{+24.6}                 & \textbf{+6.0}                & \textbf{+14.4}       \\
\bottomrule
\end{tabular}

\end{table*}

\paragraph{MEMORA improves both cross-video and single-video memory probes.}
Table~\ref{tab:memorization-main} reports per-type accuracy under the parametric-missed, MEMORA-committed diagnostic; Appendix~\ref{app:eamqa-evaluation-views} reports the full benchmark and both conditional views.
Under this diagnostic view, MEMORA is best on all four question types and overall, leading the strongest non-MEMORA condition (\textsc{Flat-1D} after online editing) by $+20.5$ points on Gemma-4-31B-it; the lead holds on both the cross-video semantic probes (\textsc{SPref}/\textsc{SHabit}/\textsc{SRoutine}) and the single-video episodic probe (\textsc{ERecall}), showing gains on both cross-session semantic probes and within-video episodic recall.

\paragraph{Both online editing and offline consolidation contribute.}
Online editing improves Graph-2D on all four question types ($+7.0$ points overall), whereas its effect on Flat-1D is smaller and mixed ($+2.0$ overall, with a slight drop on \textsc{SPref}).
Even without consolidation, MEMORA-Episodic is particularly strong on within-video \textsc{ERecall} ($73.8$), indicating that typed episodic organization already yields strong within-video recall.
Offline consolidation adds $+14.4$ points over MEMORA-Episodic, with its largest gain on \textsc{SRoutine} ($+24.6$), followed by \textsc{SHabit} ($+13.7$), \textsc{SPref} ($+9.4$), and \textsc{ERecall} ($+6.0$).
This pattern is consistent with consolidation contributing more to cross-event than within-video memory; additional store/editing and consolidation ablations are reported in Appendix~\ref{app:embodied-memory}.

\paragraph{Cross-backbone consistency.}
Across the other three open-weight backbones under the same parametric-missed, MEMORA-committed diagnostic, MEMORA leads the strongest non-MEMORA condition on every question type, and the offline-consolidation lift remains positive for every type and backbone.
Per-backbone tables and complementary evaluation views are in Appendices~\ref{app:per-backbone-per-type} and~\ref{app:eamqa-evaluation-views}.


\subsection{MEMORA-Planning Results}
\label{sec:planning-results}

Planning tests whether formed memory can guide future action rather than only answer retrospective questions.
Replay tasks contain a previously observed workflow, while Generalize tasks require transfer or composition when no single stored episode matches the goal.

\begin{figure*}[t]
\centering
\includegraphics[width=1\linewidth]{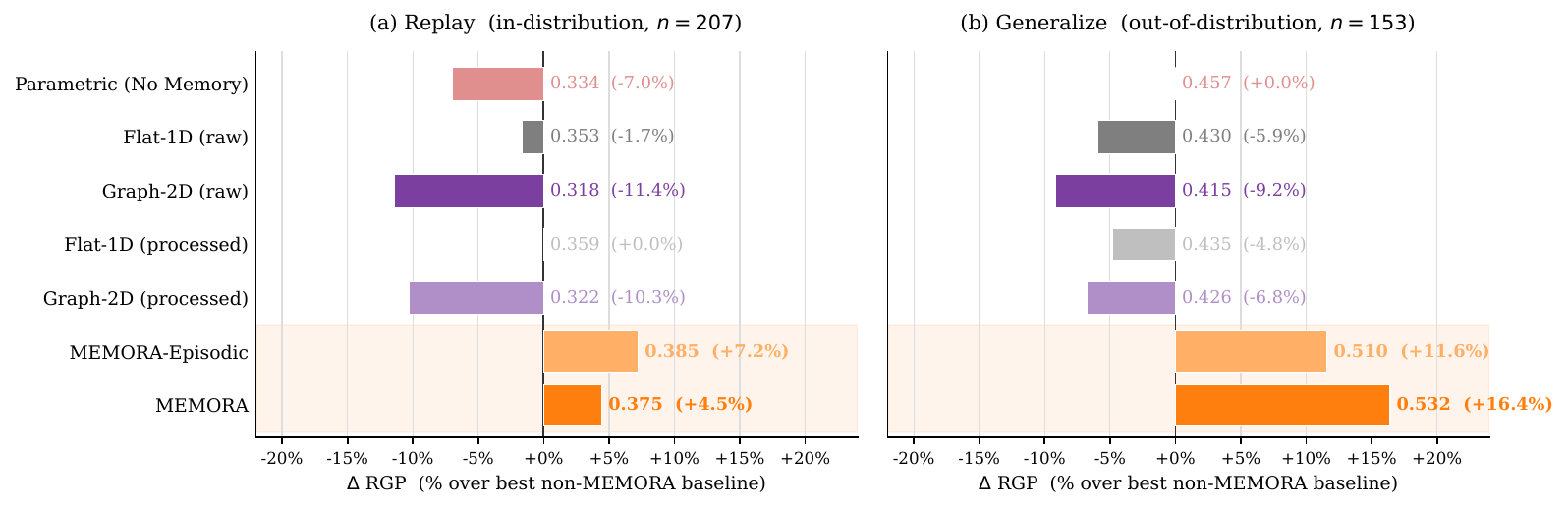}
\caption{%
\textbf{RGP on Gemma-4-31B-it} (highlighted Generalize case).
Bars show absolute RGP and $\Delta\%$ vs.\ the strongest non-MEMORA baseline ($0\%$ line).
\textbf{(a)}~Replay ($n{=}207$): MEMORA-Episodic $+7.2\%$.
\textbf{(b)}~Generalize ($n{=}153$): both MEMORA variants exceed the parametric prior; full MEMORA reaches $+16.4\%$, while non-MEMORA retrieval baselines fall below no memory.
Cross-backbone and per-axis details are in Appendix~\ref{app:planning-full}.
}
\label{fig:planning-main}
\end{figure*}

\begin{figure*}[!t]
\centering
\includegraphics[width=1\linewidth,keepaspectratio]{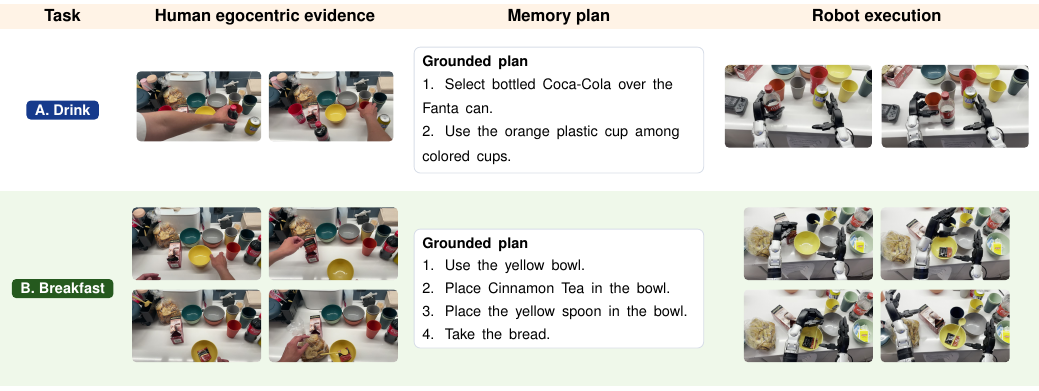}
\caption{\textbf{Physical robot demonstration.}
MEMORA transfers memory formed from human egocentric video to a rule-based Unitree~G1 for \emph{Prepare Drink} and \emph{Breakfast}, grounding preferences, objects, and sequence.}
\label{fig:robot-demo-main}
\end{figure*}

\paragraph{MEMORA's strongest planning gains occur when exact replay is unavailable.}
Across the four answer backbones, a MEMORA-family variant is best in all eight (backbone, benchmark) cells.
For three of the four backbones, the gain is larger on Generalize than on Replay, suggesting that MEMORA is especially useful when the model composes beyond a matching observed episode.
In the Gemma-4-31B-it case, the contrast is informative: MEMORA-Episodic is strongest on Replay, where a matched workflow can suffice, whereas full MEMORA is strongest on Generalize, where consolidated knowledge supports composition. All non-MEMORA retrieval baselines fall below the no-memory prior on Generalize, while MEMORA improves RGP by $+16.4\%$ over the strongest non-MEMORA baseline, suggesting that access to past experience alone is insufficient without an interface that exposes useful evidence for transfer.
This is the planning behavior illustrated in Table~\ref{tab:benchmark-examples}: reusable procedure is retrieved first and then grounded in remembered entities and places.
Per-cell results and cross-backbone scaling are in Appendix~\ref{app:planning-full} (Figure~\ref{fig:planning-xbackbone-appendix}).

\Needspace{8\baselineskip}
\paragraph{Offline consolidation benefits stronger planners on Generalize.}
On Replay, MEMORA-Episodic often matches or exceeds full MEMORA, consistent with the relevant routine already being available as episodic evidence.
On Generalize, full MEMORA has higher RGP than MEMORA-Episodic on the two strongest reasoning backbones (Q35 $+.025$; G31 $+.022$), while smaller backbones are neutral or negative under the same comparison.
On Gemma-4-31B-it, per-axis tables show the largest relative gains on procedure order ($+43\%$) and preference adherence ($+14\%$), while key-object coverage remains competitive with the parametric prior (Appendices~\ref{app:planning-full},~\ref{app:planning-grounding-diag}).

\paragraph{Tool use and ablations provide supporting evidence.}
Flat-1D uses keyword \texttt{search}, whereas MEMORA provides typed access to routines, objects, and preferences (Appendix~\ref{app:planning-full}; Fig.~\ref{fig:store-contribution}).
Replacing the main editor with Qwen3-14B and Omni perception with VL-only perception reduces EAM-QA by $5.8$ and $2.7$ points, respectively; removing \InfMem{} shifts $35\%$ of planning calls toward episodic activity search.
The editor ablation also raises E-selection from $42.8\%$ to $62.3\%$, consistent with reduced usability rather than only reduced completeness (Appendices~\ref{app:ablation-results},~\ref{app:ablation-master}).
Additional breakdowns, grounding metrics, and worked examples are in Appendix~\ref{app:planning-full}.


\subsection{Physical Robot Execution}
\label{sec:robot-demo}

We include a physical demonstration to test whether memory formed from human egocentric experience can guide a robot acting in the same scene.
Its participant-specific language plans are executed on a Unitree~G1 by a fixed rule-based controller, isolating grounding without training on robot demonstrations.
Baseline plans often select wrong objects, miss remembered preferences, or produce incomplete sequences; with MEMORA plans, the robot succeeds on both tasks.
We treat this as a qualitative execution check, not a standalone robot benchmark; Figure~\ref{fig:robot-demo-main} summarizes both tasks, with full traces and protocol in Appendix~\ref{app:robot}.

%% file: sections_new/Discussion.tex
\section{Discussion}
\label{sec:discussion}

\paragraph{Memory as a substrate for reasoning and planning.}
MEMORA separates reusable procedural evidence from participant-specific grounding.
This separation is most consequential on \textsc{Generalize}, where no single stored episode fully specifies a new goal and planning depends on composing the two.
\textsc{Replay} can often recover a matched episodic workflow, whereas \textsc{Generalize} asks the planner to compose knowledge accumulated across prior experience and reinstantiate that knowledge with remembered entities, states, and locations.

\paragraph{What the ablations isolate.}
Within-family contrasts isolate online editing and offline consolidation by holding the read interface fixed; cross-family comparisons instead evaluate complete memory interfaces and are best interpreted at the system level because representation, organization, and access change together.

\paragraph{Why memory can hurt.}
The Generalize results show that memory presence alone is insufficient: flat retrieval can fall below the parametric prior when large, weakly structured passages introduce irrelevant or conflicting evidence.
Online editing reduces this context pollution, while typed retrieval limits the evidence exposed to the answerer or planner.
The baseline pattern is also representation-dependent: Graph-2D is relatively stronger on within-video \textsc{ERecall}, while Flat-1D is stronger on cross-session semantic probes, suggesting that memory organization changes which kinds of evidence are easiest to access.
Offline consolidation contributes a different benefit by converting repeated activity into reusable regularities; its larger gains on the two higher-performing reasoning backbones are consistent with those models making more effective use of the regularities alongside entity and location evidence.
Together, these results suggest that memory quality depends on how experience is formed, organized, and exposed at use time, not only on how much history is retained.

\paragraph{Scope within an embodied-agent stack.}
MEMORA addresses long-horizon context accumulated through a participant's ongoing experience.
Vision--language--action policies map instructions and observations toward actions, world or action models reason over possible futures, and low-level controllers execute motions on a particular embodiment.
MEMORA instead exposes prior participant-specific places, state changes, procedures, and regularities as semantic--procedural context.
Egocentric video is particularly useful here because its longitudinal first-person record captures how a particular person actually uses a familiar environment.
Goals such as ``prepare a drink'' or ``prepare breakfast'' are under-specified by the current instruction and scene alone: remembered choices of beverage, cup, objects, and order determine what the goal means for that participant.
The same egocentric signal can therefore support both action-capability learning and memory formation for later grounding.
Closed-loop deployment could further pair this memory-grounded context with an execution policy and a progress monitor that verifies intermediate state transitions and triggers replanning~\citep{skreta2024replanroboticreplanningperception,dai2026seeplanrewindprogressaware,liu2026checkvlaexecutiontimeverificationactionconditioned}; Appendix~\ref{app:robot} discusses practical configurations.
Accordingly, the Unitree demonstration evaluates grounding for downstream execution, not a new controller or closed-loop robot policy.

%% file: sections_new/Conclusion.tex
\section{Conclusion}
\label{sec:conclusion}

We introduce \textbf{MEMORA}, which forms longitudinal egocentric experience into typed, editable, and consolidated memory, and \textbf{MEMORA-Bench} for retrospective faithfulness and prospective planning. Across four backbones, MEMORA achieves the strongest aggregate planning performance among the evaluated complete memory interfaces, with gains up to +16.6\% relative RGP on \textsc{Generalize}. A physical-robot demonstration shows participant-specific memory can ground robot plans. Together, these results show that actively formed memory can turn accumulated embodied experience into a persistent basis for future reasoning and action.

%% file: sections_new/Limitations.tex
\section*{Limitations}
\label{sec:limitations}

\paragraph{Domain and horizon.}
Our evaluation is limited to kitchen activity in the EPIC-KITCHENS-100 extension videos~\citep{damen2022epic}, with at most $15$ sessions per participant.
This horizon exposes recurring places, objects, and routines, but not months-long drift, seasonal reorganization, or rare events.
Extending MEMORA to other environments and months-to-years horizons may require stronger capacity control, forgetting, and consolidation policies.

\paragraph{Pipeline dependence.}
MEMORA depends on the quality of perception and write-time memory editing.
The ablations in Appendix~\ref{app:ablation-results} show that weaker editors or perception backbones reduce EAM-QA performance, so improving memory formation remains important for both retrieval and planning.
Because memory is formed from perceived segments, missed objects can be recovered only if they appear in later evidence, and repeated perception errors may enter consolidated knowledge if subsequent observations do not contradict them.

\paragraph{Evaluation scope.}
The Flat-1D, Graph-2D, and MEMORA comparisons evaluate complete memory interfaces, including their representations, retrieval tools, and instructions; only the within-family contrasts isolate online editing or offline consolidation.
Generalize tasks are proposed and filtered relative to each participant's recorded experience and scored against verified task-level references.
They test transfer and composition beyond matched Replay episodes, but do not substitute for an independently authored distribution of external goals.

\paragraph{Physical deployment.}
The Unitree~G1 result in \S\ref{sec:robot-demo} is a qualitative transfer check rather than a closed-loop robot benchmark; we do not claim a new controller or learned policy, only that the memory state formed from human egocentric video is usable as language-level grounding for downstream physical action.
Privacy, fairness, and dual-use considerations associated with persistent participant-specific memory are discussed in the Ethical Considerations section.

%% file: sections_new/Ethical_Considerations.tex
\section*{Ethical Considerations}
\label{sec:ethics}

\paragraph{Participant consent and data provenance.}
MEMORA-Bench is built exclusively on EPIC-KITCHENS-100~\citep{damen2022epic}, a publicly released egocentric dataset whose recordings were obtained with participant consent and under institutional ethics review by the original dataset authors.
We add no new recordings, no additional identifiers, and no information that was not already present in the source dataset; the participant-specific memories we construct are derived representations of the same released video.
Participant identifiers (e.g.\ \texttt{P01}, \texttt{P09}) follow the dataset's pseudonymous labels.

\paragraph{Privacy implications of persistent participant-specific memory.}
A practical embodied agent that maintains months- or years-long memory of an individual's home, routines, preferences, and object choices creates real privacy exposure beyond the present research setting.
We frame MEMORA as a research substrate for memory \emph{construction and assessment}, not as a deployment-ready system.
Any downstream deployment of similar memory machinery should, at minimum, obtain explicit informed consent (including for downstream uses such as planning or robot transfer), store participant memories under appropriate access control with on-device or end-to-end-encrypted storage where feasible, support per-participant inspection and deletion, and avoid sharing per-participant memories across users.
These are engineering and governance prerequisites that lie outside the present paper but which we view as binding conditions for any real deployment.

\paragraph{Population coverage and fairness.}
EPIC-KITCHENS-100 participants are demographically concentrated and recorded under a single activity protocol; routines, preferences, object inventories, and verbal/non-verbal cues may differ substantially across cultures, household structures, abilities, and economic contexts.
We therefore do not claim that MEMORA's behavior generalises to populations outside this protocol, and we treat cross-cultural and cross-ability extensions as both a fairness concern and the cleanest empirical test of the typed-store formulation; we flag this explicitly so that downstream users do not over-generalise from our reported numbers.

\paragraph{Physical deployment.}
The Unitree~G1 demonstration in \S\ref{sec:robot-demo} is a qualitative transfer check, not a closed-loop robot benchmark, and uses a rule-based controller rather than a learned visuomotor policy.
Real physical deployment of any system that converts human egocentric memory into robot behavior introduces safety considerations (workspace design, operator supervision, failure modes under perception error) that are beyond the scope of this paper but which any practical instantiation must address before unsupervised use.

\paragraph{Dual use and intended use.}
Persistent first-person memory of an individual could in principle be repurposed for surveillance or behavioural monitoring.
We explicitly position MEMORA as supporting \emph{participant-aligned} assistive applications --- where the memory is constructed for, with the consent of, and for the benefit of the same person whose experience it represents --- and not as a third-party monitoring tool.
We discourage applications in which a memory of an individual is constructed or queried without that individual's informed, ongoing consent.

\paragraph{Reproducibility and openness.}
The MEMORA pipeline uses only publicly released open-weight language and multimodal models (full list and identifiers in Appendix~\ref{app:impl}); we do not depend on closed-weight APIs for any headline result.
This is intended both to support independent reproduction and to keep the cost of replicating, scrutinising, or auditing the system within reach of academic groups.

%% file: sections_new/Appendix.tex

\providecolor{promptBg}{HTML}{F3F7FC}
\providecolor{promptFrame}{HTML}{7FA6D8}
\providecolor{jsonBg}{HTML}{F2F8F5}
\providecolor{jsonFrame}{HTML}{7DB39D}
\providecolor{guardBg}{HTML}{FFF7E6}
\providecolor{guardFrame}{HTML}{D9A441}
\providecolor{listingRule}{HTML}{B8C2CC}
\providecolor{flowInputBg}{HTML}{EEF4FF}
\providecolor{flowInputFrame}{HTML}{6E9BD6}
\providecolor{flowRuleBg}{HTML}{EEF8F1}
\providecolor{flowRuleFrame}{HTML}{6FAE7D}
\providecolor{flowLLMBg}{HTML}{F7F0FF}
\providecolor{flowLLMFrame}{HTML}{A783D8}
\providecolor{flowHybridBg}{HTML}{FFF6E8}
\providecolor{flowHybridFrame}{HTML}{D9A441}
\providecolor{flowOutputBg}{HTML}{EEF7F8}
\providecolor{flowOutputFrame}{HTML}{5AA7B3}
\providecolor{flowNeutralFrame}{HTML}{B8C2CC}

\lstdefinestyle{compactjson}{
  basicstyle=\ttfamily\scriptsize,
  breaklines=true,
  breakatwhitespace=false,
  columns=fullflexible,
  keepspaces=true,
  showstringspaces=false,
  frame=single,
  framerule=0.4pt,
  rulecolor=\color{listingRule},
  backgroundcolor=\color{jsonBg},
  aboveskip=3pt,
  belowskip=3pt,
  captionpos=b,
  literate={_}{{\_}}1,
}

\lstdefinestyle{promptskel}{
  basicstyle=\ttfamily\scriptsize,
  breaklines=true,
  breakatwhitespace=false,
  columns=fullflexible,
  keepspaces=true,
  frame=single,
  framerule=0.4pt,
  rulecolor=\color{listingRule},
  backgroundcolor=\color{promptBg},
  aboveskip=3pt,
  belowskip=3pt,
  captionpos=b,
  literate={_}{{\_}}1,
}

\lstdefinestyle{promptwide}{
  basicstyle=\ttfamily\footnotesize,
  breaklines=true,
  breakatwhitespace=false,
  columns=fullflexible,
  keepspaces=true,
  showstringspaces=false,
  frame=single,
  framerule=0.55pt,
  rulecolor=\color{promptFrame},
  backgroundcolor=\color{promptBg},
  aboveskip=5pt,
  belowskip=4pt,
  captionpos=b,
  xleftmargin=0.5em,
  xrightmargin=0.5em,
  literate={_}{{\_}}1,
}

\lstdefinestyle{jsonwide}{
  basicstyle=\ttfamily\footnotesize,
  breaklines=true,
  breakatwhitespace=false,
  columns=fullflexible,
  keepspaces=true,
  showstringspaces=false,
  frame=single,
  framerule=0.55pt,
  rulecolor=\color{jsonFrame},
  backgroundcolor=\color{jsonBg},
  aboveskip=5pt,
  belowskip=4pt,
  captionpos=b,
  xleftmargin=0.5em,
  xrightmargin=0.5em,
  literate={_}{{\_}}1,
}

\lstdefinestyle{guardwide}{
  basicstyle=\ttfamily\footnotesize,
  breaklines=true,
  breakatwhitespace=false,
  columns=fullflexible,
  keepspaces=true,
  showstringspaces=false,
  frame=single,
  framerule=0.55pt,
  rulecolor=\color{guardFrame},
  backgroundcolor=\color{guardBg},
  aboveskip=5pt,
  belowskip=4pt,
  captionpos=b,
  xleftmargin=0.5em,
  xrightmargin=0.5em,
  literate={_}{{\_}}1,
}

\newlength{\flowboxwd}
\setlength{\flowboxwd}{0.62\textwidth}
\newcommand{\flowstage}[3]{%
  \fbox{\begin{minipage}{\flowboxwd}
    \centering
    \textbf{#1}\hfill{\footnotesize\textit{#2}}\par\vspace{0.2em}
    {\footnotesize #3}
  \end{minipage}}%
}
\newcommand{\flowtag}[1]{\textit{#1}}
\newcommand{\flowpill}[2]{%
  \begingroup
  \setlength{\fboxsep}{1.5pt}%
  \colorbox{#1}{\scriptsize\strut\textsf{#2}}%
  \endgroup
}
\newcommand{\flowphase}[1]{%
  \vspace{0.12em}
  {\footnotesize\bfseries #1}\par\vspace{0.12em}
}
\newcommand{\flowarrow}{\\[-0.08em]{\color{flowNeutralFrame}\footnotesize$\downarrow$}\\[-0.08em]}
\newcommand{\flowstagecolor}[5]{%
  \begingroup
  \setlength{\fboxsep}{4pt}%
  \fcolorbox{#2}{#1}{%
    \begin{minipage}{0.76\textwidth}
      \textbf{#3}\hfill #4\par\vspace{0.12em}
      {\footnotesize #5}
    \end{minipage}}%
  \endgroup
}
\newcommand{\flowcard}[6]{%
  \begingroup
  \setlength{\fboxsep}{4pt}%
  \fcolorbox{#2}{#1}{%
    \begin{minipage}{#3}
      \textbf{#4}\hfill #5\par\vspace{0.12em}
      {\footnotesize #6}
    \end{minipage}}%
  \endgroup
}
\newcommand{\promptpanelbox}[2]{%
  \fcolorbox{promptFrame}{promptBg}{%
    \begin{minipage}[t]{#1}
    \ttfamily\scriptsize\raggedright
    #2
    \end{minipage}}%
}
\newcommand{\jsonpanelbox}[2]{%
  \fcolorbox{jsonFrame}{jsonBg}{%
    \begin{minipage}[t]{#1}
    \ttfamily\scriptsize\raggedright
    #2
    \end{minipage}}%
}

\appendix

\makeatletter
\setlength{\@fptop}{0pt}
\setlength{\@fpsep}{8pt plus 2pt minus 2pt}
\setlength{\@fpbot}{0pt plus 1fil}
\setlength{\@dblfptop}{0pt}
\setlength{\@dblfpsep}{8pt plus 2pt minus 2pt}
\setlength{\@dblfpbot}{0pt plus 1fil}
\makeatother

\paragraph{Appendix roadmap.}
The appendix is organized around construction, evaluation, and controlled analysis.
Appendix~\ref{app:kb-construction} specifies the memory-bank construction pipeline from egocentric video.
Appendix~\ref{app:embodied-memory} defines the retrospective Embodied Memory assessment, participant scope, and unsupported-answer controls.
Appendix~\ref{app:planning} specifies the MEMORA-Planning benchmark and rule-based evaluation panel.
Appendix~\ref{app:robot} documents the physical robot demonstration used as a qualitative deployment check.
Appendix~\ref{app:ablation} defines controlled ablations over processor and answer-agent choices.
Appendix~\ref{app:impl} consolidates the model registry, inference stack, perception preprocessing, and retrieval/agent-loop settings shared across all of the above.

\begin{figure*}[t]
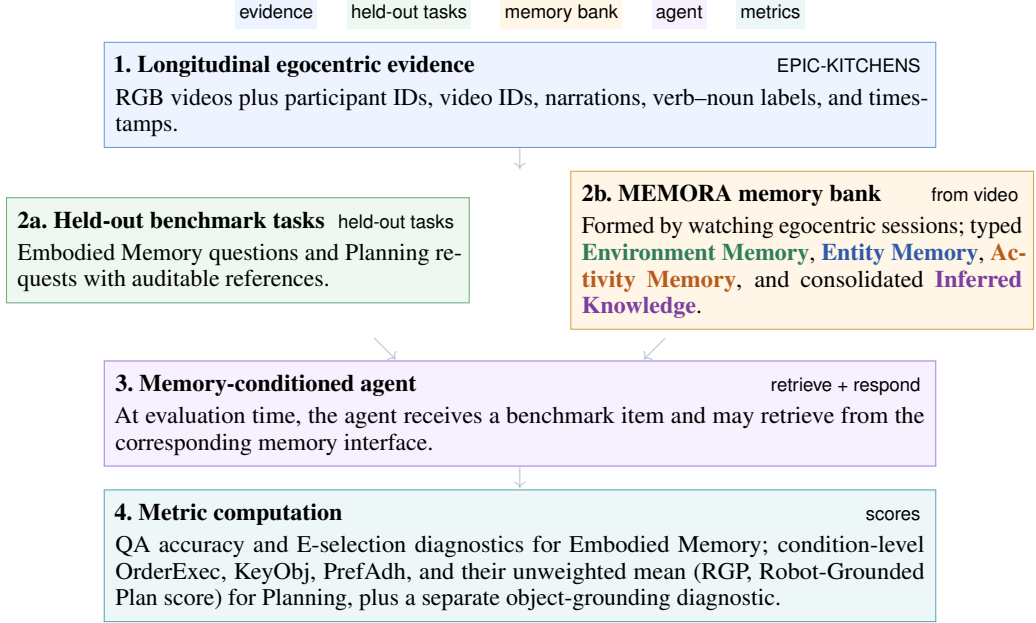

\centering
\small
\setlength{\fboxsep}{5pt}
\begin{minipage}{0.88\textwidth}
\centering
{\footnotesize
\flowpill{flowInputBg}{evidence}
\quad \flowpill{flowRuleBg}{held-out tasks}
\quad \flowpill{flowHybridBg}{memory bank}
\quad \flowpill{flowLLMBg}{agent}
\quad \flowpill{flowOutputBg}{metrics}
}
\\[0.45em]

\flowstagecolor{flowInputBg}{flowInputFrame}{1.\ Longitudinal egocentric evidence}
  {\flowpill{flowInputBg}{EPIC-KITCHENS}}
  {RGB videos plus participant IDs, video IDs, narrations, verb--noun labels, and timestamps.}
\flowarrow
\begin{minipage}[t]{0.47\textwidth}
\centering
\flowcard{flowRuleBg}{flowRuleFrame}{0.88\linewidth}{2a.\ Held-out benchmark tasks}
  {\flowpill{flowRuleBg}{held-out tasks}}
  {Embodied Memory questions and Planning requests with auditable references.}
\end{minipage}
\hfill
\begin{minipage}[t]{0.47\textwidth}
\centering
\flowcard{flowHybridBg}{flowHybridFrame}{0.88\linewidth}{2b.\ MEMORA memory bank}
  {\flowpill{flowHybridBg}{from video}}
  {Formed by watching egocentric sessions; typed \EnvMem{}, \EntMem{}, \ActMem{}, and consolidated \InfMem{}.}
\end{minipage}

\vspace{0.25em}
{\color{flowNeutralFrame}\footnotesize$\searrow$ \qquad\qquad\qquad\qquad\qquad $\swarrow$}\\[-0.08em]
\flowstagecolor{flowLLMBg}{flowLLMFrame}{3.\ Memory-conditioned agent}
  {\flowpill{flowLLMBg}{retrieve + respond}}
  {At evaluation time, the agent receives a benchmark item and may retrieve from the corresponding memory interface.}
\flowarrow
\flowstagecolor{flowOutputBg}{flowOutputFrame}{4.\ Metric computation}
  {\flowpill{flowOutputBg}{scores}}
  {QA accuracy and E-selection diagnostics for Embodied Memory; condition-level OrderExec, KeyObj, PrefAdh, and their unweighted mean (RGP, Robot-Grounded Plan score) for Planning, plus a separate object-grounding diagnostic.}
\end{minipage}
\caption{\textbf{Appendix visual roadmap for MEMORA-Bench.}
The appendix separates four concepts that are easy to conflate: longitudinal evidence, held-out benchmark tasks, memory-bank construction from video, and memory-conditioned evaluation.
Subsequent figures expand the memory-bank construction pipeline (Figure~\ref{fig:memory-bank-flow}), the Embodied Memory benchmark generator (Figure~\ref{fig:gen-pipeline-overview}), the Planning benchmark generator (Figure~\ref{fig:planning-benchmark-flow}), and the final retrieval-and-scoring loop (Figure~\ref{fig:evaluation-agent-flow}).}
\label{fig:memora-bench-overview}
\end{figure*}


\input{sections_new/Appendix_sections/pipeline}
\FloatBarrier
\raggedbottom

\input{sections_new/Appendix_sections/MEMORA_Embodied_Memory}
\FloatBarrier

\input{sections_new/Appendix_sections/MEMORA_Planning}
\FloatBarrier

\input{sections_new/Appendix_sections/robot}
\FloatBarrier

\input{sections_new/Appendix_sections/ablation}
\FloatBarrier

\input{sections_new/Appendix_sections/implementation_details}

%% file: sections_new/Appendix_sections/pipeline.tex
\section{Memory-Bank Construction Pipeline}
\label{app:kb-construction}

Appendix~\ref{app:kb-construction} specifies how MEMORA constructs the typed memory bank used by the Embodied Memory and Planning evaluations.
The pipeline maps egocentric EPIC-KITCHENS videos into the four stores defined in Section~\ref{sec:method}: \EnvMem{}, \EntMem{}, \ActMem{}, and offline \InfMem{}.
The presentation follows processor roles rather than engineering stage names.

\paragraph{Terminology (\InfMem{} and MEMORA-Episodic).}
In the main paper, cross-episode routines, habits, and preferences are typed as \InfMem{} and populated by offline consolidation (Table~\ref{tab:eam-memory-stores}).
This appendix decomposes that offline path into two processors: an \textbf{Offline Consolidation Processor} (per-video storage preferences, organizational habits, workflow patterns, and repeated action sequences) and an \textbf{\InfMem{} Enrichment Processor} (participant-level routine-skill indices, generated preferences, and retrieval text).
Together they form the consolidated \InfMem{} bank used at evaluation time.
Full \textbf{MEMORA} loads the enriched participant memory bank: EAM-QA reaches consolidation outputs through \texttt{search\_patterns} and, under full MEMORA, also indexes enriched routine skills and generated preferences in the same unified retrieval interface; Planning additionally advertises dedicated \texttt{get\_routine\_skill} and \texttt{get\_preferences} entry points over the enriched consolidated \InfMem{} slice.
The label \textbf{MEMORA-Episodic} matches the main-paper ablation on online \EnvMem{}/\EntMem{}/\ActMem{}: EAM-QA evaluates the per-video memory bank before enrichment through the standard layered search tools; Planning evaluates the same episodic stores through \texttt{search\_objects}, \texttt{get\_object\_history}, and \texttt{search\_activities}.

\begin{figure*}[t]
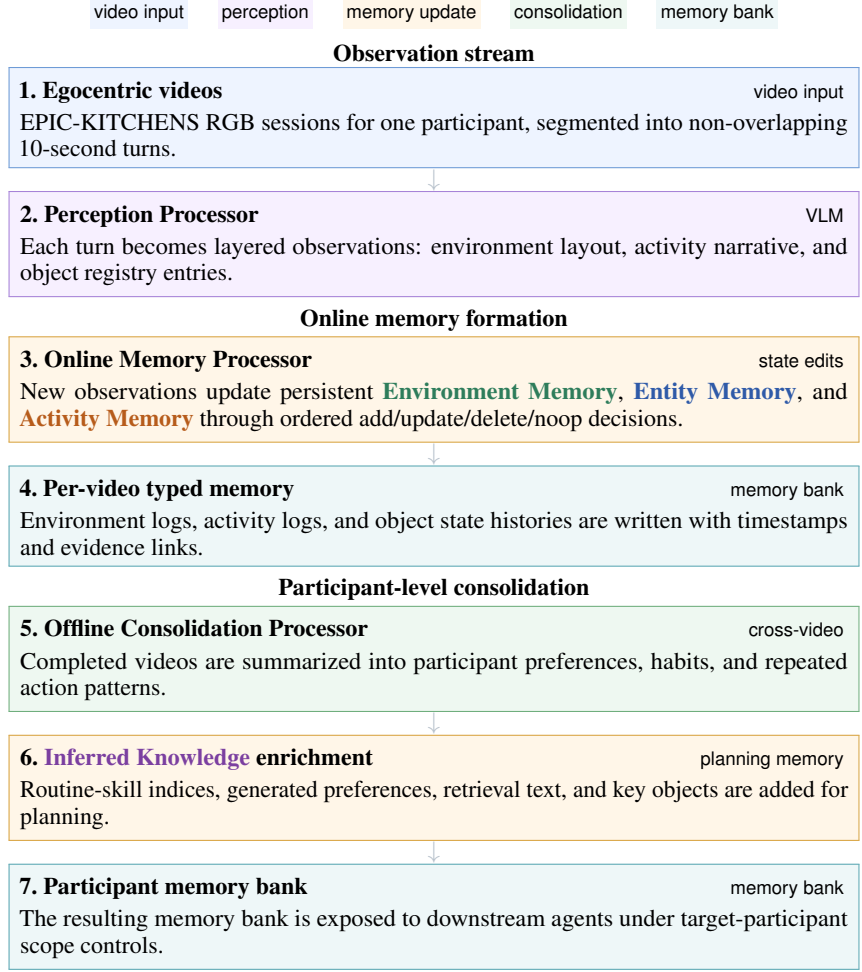

\centering
\small
\setlength{\fboxsep}{5pt}
\begin{minipage}{0.90\textwidth}
\centering
{\footnotesize
\flowpill{flowInputBg}{video input}
\quad \flowpill{flowLLMBg}{perception}
\quad \flowpill{flowHybridBg}{memory update}
\quad \flowpill{flowRuleBg}{consolidation}
\quad \flowpill{flowOutputBg}{memory bank}
}
\\[0.45em]

\flowphase{Observation stream}
\flowstagecolor{flowInputBg}{flowInputFrame}{1.\ Egocentric videos}
  {\flowpill{flowInputBg}{video input}}
  {EPIC-KITCHENS RGB sessions for one participant, segmented into non-overlapping 10-second turns.}
\flowarrow
\flowstagecolor{flowLLMBg}{flowLLMFrame}{2.\ Perception Processor}
  {\flowpill{flowLLMBg}{VLM}}
  {Each turn becomes layered observations: environment layout, activity narrative, and object registry entries.}

\vspace{0.28em}
\flowphase{Online memory formation}
\flowstagecolor{flowHybridBg}{flowHybridFrame}{3.\ Online Memory Processor}
  {\flowpill{flowHybridBg}{state edits}}
  {New observations update persistent \EnvMem{}, \EntMem{}, and \ActMem{} through ordered add/update/delete/noop decisions.}
\flowarrow
\flowstagecolor{flowOutputBg}{flowOutputFrame}{4.\ Per-video typed memory}
  {\flowpill{flowOutputBg}{memory bank}}
  {Environment logs, activity logs, and object state histories are written with timestamps and evidence links.}

\vspace{0.28em}
\flowphase{Participant-level consolidation}
\flowstagecolor{flowRuleBg}{flowRuleFrame}{5.\ Offline Consolidation Processor}
  {\flowpill{flowRuleBg}{cross-video}}
  {Completed videos are summarized into participant preferences, habits, and repeated action patterns.}
\flowarrow
\flowstagecolor{flowHybridBg}{flowHybridFrame}{6.\ \InfMem{} enrichment}
  {\flowpill{flowHybridBg}{planning memory}}
  {Routine-skill indices, generated preferences, retrieval text, and key objects are added for planning.}
\flowarrow
\flowstagecolor{flowOutputBg}{flowOutputFrame}{7.\ Participant memory bank}
  {\flowpill{flowOutputBg}{memory bank}}
  {The resulting memory bank is exposed to downstream agents under target-participant scope controls.}
\end{minipage}
\caption{\textbf{MEMORA memory-bank construction pipeline.}
The system watches egocentric videos turn by turn, converts each clip into structured observations, updates typed memory online, and then consolidates participant-level routines and preferences for downstream Embodied Memory and Planning evaluations.}
\label{fig:memory-bank-flow}
\end{figure*}

\subsection{Implementation Overview}

Each video is divided into non-overlapping 10-second turns.
In implementation terms, the \emph{Perception Processor} realizes the Segment
Encoder $\phi$, while the \emph{Online Memory Processor} realizes the
write-time \textsc{Edit} update described in Section~\ref{sec:method}.
For turn $t$, the Perception Processor emits a layered observation
$z_t=(z_t^{\mathrm{env}},z_t^{\mathrm{act}},z_t^{\mathrm{ent}})$.
The Online Memory Processor then updates a persistent memory state
$\mathcal{M}_t=(\mathcal{M}_t^{\mathrm{env}},\mathcal{M}_t^{\mathrm{act}},\mathcal{M}_t^{\mathrm{ent}},\mathcal{M}^{\mathrm{inf}})$
in timestamp order:
\begin{equation}
z_t = \phi(v_t, c_{t-1}), \qquad
\mathcal{M}_t = \textsc{Edit}(\mathcal{M}_{t-1}, z_t).
\end{equation}
In this notation, $v_t$ is the 10-second clip and $c_{t-1}$ is a bounded textual summary of earlier turns in the same video.
Previous video frames are not retained in the model KV cache across turns.

\begin{table*}[t]
\centering
\small
\setlength{\tabcolsep}{4pt}
\renewcommand{\arraystretch}{1.08}
\caption{Main-paper memory-bank construction processors.
The table specifies processor contracts (input scope, outputs, and construction notes); model identifiers appear in Table~\ref{tab:impl-models} and temporal preprocessing in Appendix~\ref{app:impl}.}
\label{tab:kb-processors}
\begin{tabularx}{\textwidth}{@{}p{0.18\textwidth}p{0.20\textwidth}p{0.22\textwidth}X@{}}
\toprule
\textbf{Processor} & \textbf{Input scope} & \textbf{Output} & \textbf{Construction notes} \\
\midrule
Perception Processor &
One 10-second egocentric clip &
Turn-local \texttt{environment}, \texttt{activity\_narrative}, and \texttt{object\_registry} records &
Run independently per turn; 2\,FPS, 24-frame cap, audio on (Omni's built-in audio path); vision-only ablation drops audio at decode time. \\
Online Memory Processor &
Ordered turns for one video &
Per-video \texttt{environment\_log}, \texttt{activity\_log}, and edited \texttt{object\_registry} &
Per-video scope; dense retrieval pre-filter when the entity registry exceeds the prompt budget. \\
Offline Consolidation Processor &
Completed videos for one participant &
Per-video consolidation summaries: storage preferences, organizational habits, workflow patterns, and action sequences &
Post-video offline pass; reuses the loaded online memory model. \\
\InfMem{} Enrichment Processor &
One participant memory slice &
Routine-skill indices, retrieval text, key objects, and evidence-grounded generated preferences &
Rule skeleton plus one LLM preference pass per participant; closed-weight alternative in Appendix~\ref{app:ablation-enrichment}. \\
\bottomrule
\end{tabularx}
\end{table*}

Each participant memory bank comprises four typed stores: turn-local perception records, per-video environment and activity logs, an edited entity registry with operation history, and a participant-level consolidated \InfMem{} slice.

\subsection{Memory Update Semantics}

The Perception Processor produces turn-local observations only; it does not reconcile duplicate entities or maintain long-horizon state.
Cross-turn memory formation is handled by the Online Memory Processor.
Activity memory is append-only because it represents the event stream.
Environment memory is merged by stable location identifiers and retains a history of updates.
Entity memory is edited by the online memory processor: for each incoming object observation, the model emits one of \textsc{Add}, \textsc{Update}, \textsc{Delete}, or \textsc{Noop} against the compact current registry.
When the entity registry is too large for the prompt budget, dense retrieval (Appendix~\ref{app:impl}) selects the most relevant existing entities for the edit decision.

The Offline Consolidation Processor reads completed memories for a participant and writes per-video consolidation summaries (storage preferences, habits, workflow patterns, and action sequences).
These summaries support preference, habit, and routine reasoning; in the released retrospective QA panel, they are exposed only within the complete fixed memory bank of the question's target participant.
The \InfMem{} Enrichment Processor subsequently derives planning-oriented memory from the participant slice.
It creates routine-skill structures and retrieval text deterministically, then uses the main-paper enrichment LLM in Table~\ref{tab:impl-models} to produce evidence-grounded generated preferences; a closed-weight alternative is reported as a sanity check in Appendix~\ref{app:ablation-enrichment}.

\subsection{Editor Activity and Memory Compression}
\label{app:editor-activity}

This subsection documents the data behind Figure~\ref{fig:memory-maintenance} in Method~\S\ref{sec:overview}.
The 18-participant figure traces the cumulative size of Entity Memory under two write policies on the same main-paper perception stream and online memory processor (Table~\ref{tab:impl-models}): a hypothetical \emph{without-editing} policy that would persist one record per entity observation (every \texttt{state\_history} event the perception layer emits with a turn-aligned timestamp), and the MEMORA policy that emits one record per unique entity and folds repeat encounters into its \texttt{state\_history} list.
The compression ratio reported per participant is the ratio of these two record counts at end-of-session; segment-level cumulative curves use turn-aligned events only (events without a resolved \texttt{turn\_id} are dropped from both numerator and denominator so they cannot inflate either side).
Panel~(b)'s editor decision distribution is read off the released editor operation logs; these decision counts are fully available for the P01--P04 calibration cohort and partially available for the 18-participant evaluation cohort, so the distribution panel is reported on P01--P04 alone while panels~(a) and~(c) use the full 18-participant cohort.

\paragraph{Per-participant size and compression.}
Figure~\ref{fig:memory-maintenance}, panels~(a) and~(c), report the per-participant figures.
Every participant achieves at least a $9\times$ reduction; the typical (median) participant achieves $18.1\times$, and three participants exceed $30\times$.
PIDs with longer sessions and higher entity reuse (P12, P23, P25, P35) reach the largest ratios because the editor turns repeated encounters with the same kitchen objects into state updates rather than new records; participants with shorter sessions (P09) sit at the floor of the range.

\paragraph{Editor decision distribution (P01--P04).}
The calibration-cohort editor logs (P01--P04) report the editor's decisions over $n{=}37{,}648$ object observations: \textsc{Add}~$=2{,}995$ (8.0\%), \textsc{Update}~$=16{,}718$ (44.4\%), \textsc{Delete}~$=691$ (1.8\%), \textsc{Noop}~$=17{,}244$ (45.8\%).
Two readings are worth recording.
First, \textsc{Noop}~$+$~\textsc{Update} together account for $90.2\%$ of decisions: most of the editor's work is recognizing that an entity is already in memory and either revising its state or letting it stand, not adding new records.
Second, \textsc{Delete} is rare (1.8\%) but never zero --- the editor does retract identifications that subsequent observations disconfirm, which a frozen, append-only memory could not.
This decision profile is the per-event mechanism behind the $18.1\times$ aggregate compression in Figure~\ref{fig:memory-maintenance}: every \textsc{Noop} on a previously-seen object is one entity observation that does not become a new record, and every \textsc{Update} is one observation that updates an existing record rather than creating one.

\subsection{Processor Prompt Templates}

The construction pipeline in Figure~\ref{fig:memory-bank-flow} uses fixed prompt templates with runtime variables filled from the current segment, current memory state, or participant memory slice.
Figure~\ref{fig:processor-prompt-templates} summarizes the four prompted processors, and Figure~\ref{fig:kb-memory-json-examples} gives compact records for the four MEMORA memory types.
Deterministic rule components in the \InfMem{} Enrichment Processor have no prompt.

\begin{figure*}[t]
\centering
\small
\textbf{A. Processor prompt templates}\par\vspace{0.35em}
\begin{minipage}[t]{0.48\textwidth}
\textbf{Perception Processor}\vspace{0.2em}\par
\promptpanelbox{0.95\linewidth}{%
Input: current 10-second clip; optional previous context and detected objects.\\
Instruction: analyze visible egocentric kitchen activity, use relative timestamps in [0,10]s, reuse object identifiers when available.\\
Output JSON: environment layout/zones/relations; activity summary and action\_breakdown; object\_registry with visual properties, spatial info, and state.}
\end{minipage}
\hfill
\begin{minipage}[t]{0.48\textwidth}
\textbf{Online Memory Processor}\vspace{0.2em}\par
\promptpanelbox{0.95\linewidth}{%
Input: compact current object\_registry plus full new object states.\\
Instruction: edit Entity Memory with ADD, UPDATE, DELETE, or NOOP; preserve movement\_trajectory; keep exact object\_id unless adding a new object.\\
Output JSON: object\_operations with event, object\_id, changes/data, and reason.}
\end{minipage}

\vspace{0.6em}

\begin{minipage}[t]{0.48\textwidth}
\textbf{Offline Consolidation Processor}\vspace{0.2em}\par
\promptpanelbox{0.95\linewidth}{%
Input: participant\_id, completed object\_registry, and activity\_log.\\
Instruction: infer only from completed memories; identify storage preferences, workflow preferences, organizational habits, and recurring task patterns.\\
Output JSON: storage\_preferences, organizational\_habits, workflow\_patterns.}
\end{minipage}
\hfill
\begin{minipage}[t]{0.48\textwidth}
\textbf{\InfMem{} Enrichment Processor}\vspace{0.2em}\par
\promptpanelbox{0.95\linewidth}{%
Input: participant, max\_skills, allowed\_subtypes, and evidence\_items.\\
Instruction: convert low-level evidence into reusable participant-level skill memories; use only supplied evidence IDs; prefer concrete object/action/context skills.\\
Output JSON: generated\_preferences with subtype, keywords, confidence, and supporting\_evidence\_ids.}
\end{minipage}
\caption{\textbf{Prompt templates for MEMORA memory-bank construction.}
Each box preserves the operational contract of a prompted processor: what enters the module, what instruction constrains it, and what structured fields it must emit.}
\label{fig:processor-prompt-templates}
\end{figure*}

\subsection{Memory Type Examples}

Figure~\ref{fig:kb-memory-json-examples} provides abbreviated records for the four memory types used by MEMORA: \EnvMem{}, \EntMem{}, \ActMem{}, and \InfMem{}.
Field names are preserved from the released records; values are shortened to make the representation readable in print.

\begin{figure*}[t]
\centering
\begin{minipage}[t]{0.48\textwidth}
\textbf{\EnvMem{} record}\vspace{0.2em}\par
\jsonpanelbox{0.95\linewidth}{%
environment\_log: [\\
\quad location\_id: sink\_area\\
\quad current\_state: layout\_description, zones=[sink\_basin, drying\_rack, prep\_counter], spatial\_relations=[drying\_rack RIGHT\_OF sink\_basin]\\
\quad history: turn\_id + update\\
]}
\end{minipage}
\hfill
\begin{minipage}[t]{0.48\textwidth}
\textbf{\ActMem{} record}\vspace{0.2em}\par
\jsonpanelbox{0.95\linewidth}{%
activity\_log: [\\
\quad turn\_id: 12; time\_window: 120.0--130.0\\
\quad summary: participant washes a knife with a sponge in the sink\\
\quad action\_breakdown: [0--3s scrubs knife\_metal; 3--6s rinses knife\_metal]\\
]}
\end{minipage}

\vspace{0.6em}

\begin{minipage}[t]{0.48\textwidth}
\textbf{\EntMem{} record}\vspace{0.2em}\par
\jsonpanelbox{0.95\linewidth}{%
object\_registry.knife\_metal: \\
\quad visual\_properties: material=metal, size=medium\\
\quad spatial\_info: location=in sink, zone=sink\_basin\\
\quad state: current\_state=being washed, held\_by=right\_hand\\
\quad state\_history: turn 11 on counter/idle; turn 12 in sink/being washed}
\end{minipage}
\hfill
\begin{minipage}[t]{0.48\textwidth}
\textbf{\InfMem{} record}\vspace{0.2em}\par
\jsonpanelbox{0.95\linewidth}{%
Consolidation (per-video): workflow pattern ``dirty utensils are washed at the sink before reuse''; confidence=0.82\\
Enrichment (participant-level): generated preference that participant often uses a spatula to stir potatoes or vegetables; keywords=[spatula, stir, potatoes, vegetables]; supporting\_evidence\_ids=[E001,E005]}
\end{minipage}
\caption{\textbf{Abbreviated JSON-style examples of MEMORA memory records.}
The panel keeps literal schema field names but compresses values so the four record types can be compared without a split listing block.}
\label{fig:kb-memory-json-examples}
\end{figure*}
\FloatBarrier

\raggedbottom
\subsection{Memory Stores and Evaluation Scope}

\paragraph{Evaluation scope.}
In retrospective Embodied Memory, each question is evaluated against only the
fixed memory bank of its target participant. Answerable items are derived from
that participant's observed corpus; E-correct controls are remapped from a
different donor participant and therefore lack supporting target-memory
evidence. The released headline panel has no per-question temporal anchor and
does not apply state rollback. In Planning, scope control restricts both
benchmark tasks and retrieval to videos already present in the participant's
consolidated memory slice, so the agent is not evaluated on episodes it never
encoded.
Table~\ref{tab:kb-artifacts} summarizes these controls for each memory store.

\begin{table*}[t]
\centering
\small
\setlength{\tabcolsep}{4pt}
\renewcommand{\arraystretch}{1.08}
\caption{Primary memory stores and evaluation scope.
The final column distinguishes construction-time locality from the
participant-level scope used by each evaluation.}
\label{tab:kb-artifacts}
\begin{tabularx}{\textwidth}{@{}l l X X@{}}
\toprule
\textbf{Store} & \textbf{Producer} & \textbf{Content} & \textbf{Evaluation scope} \\
\midrule
Per-segment perception record & Perception Processor & Turn-local \EnvTok{environment}, \ActTok{activity}, and \EntTok{object} observations & Generated from the current 10-second segment plus bounded text context only \\
\EnvTok{environment\_log}, \ActTok{activity\_log} & Online Memory Processor & Timestamped spatial and activity memories & Restricted to the target participant's fixed memory bank \\
\EntTok{object\_registry} & Online Memory Processor & Edited entity memory with add/update/delete/noop decisions and state history & Full target-participant registry is available in headline EAM-QA \\
Consolidation summaries & Offline Consolidation Processor & Per-video storage preferences, organizational habits, workflow patterns, and action sequences & Included according to the evaluated episodic/full condition \\
Consolidated \InfMem{} slice & \InfMem{} Enrichment Processor & Routine skills, generated preferences, retrieval text, and key objects & Planning retrieval limited to source videos in the participant memory slice \\
\bottomrule
\end{tabularx}
\end{table*}

\begin{figure*}[t]
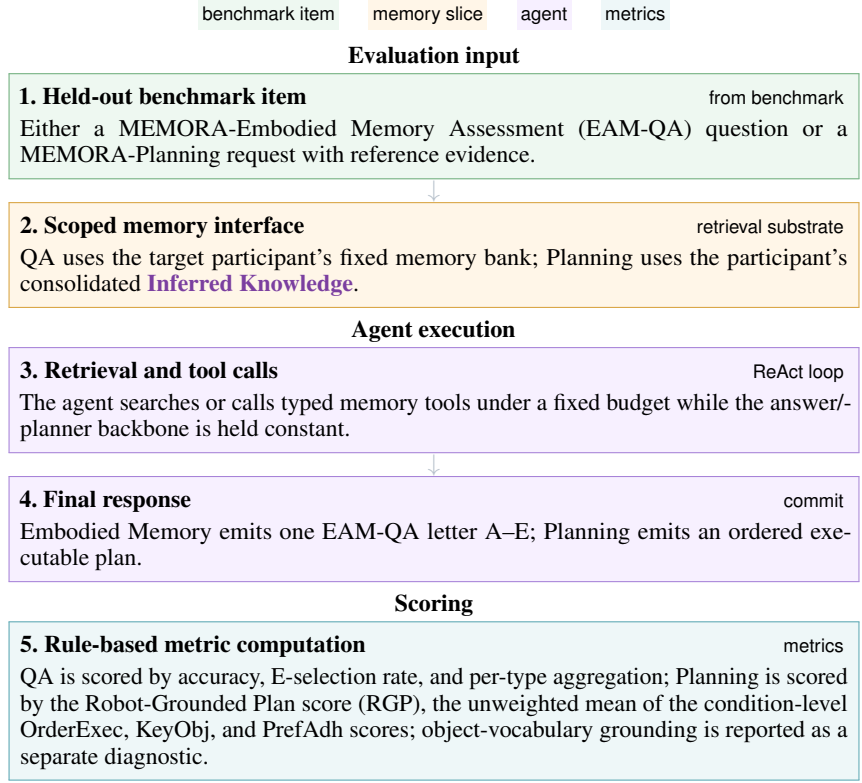

\centering
\small
\setlength{\fboxsep}{5pt}
\begin{minipage}{0.90\textwidth}
\centering
{\footnotesize
\flowpill{flowRuleBg}{benchmark item}
\quad \flowpill{flowHybridBg}{memory slice}
\quad \flowpill{flowLLMBg}{agent}
\quad \flowpill{flowOutputBg}{metrics}
}
\\[0.45em]

\flowphase{Evaluation input}
\flowstagecolor{flowRuleBg}{flowRuleFrame}{1.\ Held-out benchmark item}
  {\flowpill{flowRuleBg}{from benchmark}}
  {Either a MEMORA-Embodied Memory Assessment (EAM-QA) question or a MEMORA-Planning request with reference evidence.}
\flowarrow
\flowstagecolor{flowHybridBg}{flowHybridFrame}{2.\ Scoped memory interface}
  {\flowpill{flowHybridBg}{retrieval substrate}}
  {QA uses the target participant's fixed memory bank; Planning uses the participant's consolidated \InfMem{}.}

\vspace{0.28em}
\flowphase{Agent execution}
\flowstagecolor{flowLLMBg}{flowLLMFrame}{3.\ Retrieval and tool calls}
  {\flowpill{flowLLMBg}{ReAct loop}}
  {The agent searches or calls typed memory tools under a fixed budget while the answer/planner backbone is held constant.}
\flowarrow
\flowstagecolor{flowLLMBg}{flowLLMFrame}{4.\ Final response}
  {\flowpill{flowLLMBg}{commit}}
  {Embodied Memory emits one EAM-QA letter A--E; Planning emits an ordered executable plan.}

\vspace{0.28em}
\flowphase{Scoring}
\flowstagecolor{flowOutputBg}{flowOutputFrame}{5.\ Rule-based metric computation}
  {\flowpill{flowOutputBg}{metrics}}
  {QA is scored by accuracy, E-selection rate, and per-type aggregation; Planning is scored by the Robot-Grounded Plan score (RGP), the unweighted mean of the condition-level OrderExec, KeyObj, and PrefAdh scores; object-vocabulary grounding is reported as a separate diagnostic.}
\end{minipage}
\caption{\textbf{Memory-conditioned evaluation and scoring loop.}
The benchmark item and the scoped memory interface are kept separate: benchmark tasks define what should be answered, while memory substrates define what the agent can retrieve before committing to an answer or plan.}
\label{fig:evaluation-agent-flow}
\end{figure*}

\subsection{Limitations}

The memory bank inherits errors from the perception stream: missed objects, weak audio cues, and ambiguous hand-object interactions can propagate into later processors.
The Online Memory Processor can repair cross-turn state, but visually similar objects may still be duplicated or stale attributes may persist when evidence is weak.
The released retrospective panel evaluates a participant's completed memory bank and therefore does not test within-video prefix recall; a future temporally anchored variant would require explicit memory slicing or rollback.
Finally, consolidated \InfMem{} is optimized for planning and preference retrieval; it should not be interpreted as an exhaustive transcript of every observed event.

%% file: sections_new/Appendix_sections/MEMORA_Embodied_Memory.tex
\section{MEMORA-Embodied Memory Assessment}
\label{app:bench-quality}
\label{app:embodied-memory}

Appendix~\ref{app:embodied-memory} defines the retrospective QA arm of MEMORA-Bench.
The \textbf{MEMORA-Embodied Memory Assessment} measures whether participant-specific memory is faithful, accessible, and correctly scoped, including when the target participant's memory lacks supporting evidence.
The annotation-grounded release supports the headline four-way split (\ERecall, \SPref, \SHabit, \SRoutine) on 45 hours of EPIC-KITCHENS-100~\citep{damen2022epic} extension video across 18 participants.
Generation is architecture-agnostic: questions are mined from action annotations and do not presuppose MEMORA's four-substrate layout.
\S\ref{app:mem-bench-prep} specifies benchmark construction and question grounding; \S\ref{app:mem-bench-eval} specifies the evaluation protocol.
The EAM-QA setup evaluates retrieval from the constructed memory bank rather than generic video question answering.

\subsection{Benchmark Construction}
\label{app:mem-bench-prep}

\paragraph{EPIC-KITCHENS inputs.}
The construction builds on \textbf{EPIC-KITCHENS-100}~\citep{damen2022epic}, an egocentric kitchen corpus in which each \textbf{participant} records multiple \textbf{videos} in the same home over time.
The public release supplies:
\begin{itemize}\setlength{\itemsep}{0.1em}
\item \textbf{RGB video} for every session (\emph{not used in generation}; required at evaluation by the perception front-end);
\item \textbf{Dense action annotations}---one row per atomic action, time-aligned to the video (\emph{used in generation});
\item \textbf{Participant and session identifiers} (\texttt{participant\_id}, \texttt{video\_id}) so longitudinal behavior can be grouped (\emph{used in generation});
\item \textbf{Timestamps} for when each action occurs (\texttt{narration\_timestamp}, plus frame- and clock-based start/stop fields in the official CSV; \emph{used in generation}, with start/stop fields optional);
\item \textbf{Structured labels} (\texttt{verb}, \texttt{noun}, and their taxonomy classes) and a free-text \textbf{narration} describing the action in context (\emph{used in generation}).
\end{itemize}
The Embodied Memory generator consumes only the annotation table (not pixels): it needs a time-ordered log of \emph{who} did \emph{what}, \emph{when}, and \emph{in which video}.

The release is restricted to an \textbf{EPIC-KITCHENS-100 extension video whitelist} (45 hours, 201 videos across 18 participants), so every item remains within the same longitudinal setting.

\subsection{Question Types and Grounding}

\paragraph{Design.}
Each item is \emph{participant-specific}: it concerns one person's kitchen behavior across longitudinal sessions in the same home, not generic cooking knowledge, so the benchmark tests \emph{consolidated user models} rather than generic cooking priors.
Released questions require \emph{evidence-grounded} stems and keys: every item must cite observable actions (timestamps, \texttt{video\_id}s, verb--noun labels) drawn from EPIC narrations, so items remain auditable and cannot be answered from world knowledge alone.
Because LLM generation is noisy and many candidates fail rule checks, we \emph{overgenerate then curate} (target $\times 3$ per type before filtering) to hit per-type quotas without hand-writing questions; the surplus absorbs parse failures, weak evidence, and judge rejections.
The pipeline applies a \emph{cross-model judge} (generator \texttt{gpt-5.5} vs.\ verifier \texttt{gpt-4o-mini}) so acceptance is not decided by the same model that proposed the item.
Every released item must pass mandatory verifiability prompts at generation time plus post-hoc quality filtering (semantic deduplication, per-item verification, and answer re-derivation).

\begin{table*}[t]
\centering
\footnotesize
\setlength{\tabcolsep}{3pt}
\renewcommand{\arraystretch}{1.08}
\caption{Headline question types, inputs, and evidence thresholds in the annotation-grounded Embodied Memory release.
The dashed rule separates \textbf{semantic} probes consolidated across videos from \ERecall{}, which is \textbf{episodic} (video-specific).
The generator's JSON schema may tag \SHabit{}/\SRoutine{} with \texttt{cognitive\_level=procedural\_memory} because habits and routines describe \emph{how} someone acts; in our benchmark taxonomy they are nevertheless grouped with \SPref{} as declarative, cross-session semantic knowledge about the participant.
Each type also has \textbf{multi-hop} variants ($\texttt{reasoning\_depth}\in\{2,3\}$): the same cognitive probe, but the answer requires chaining 2--3 within-video actions (not a separate question family).}
\label{tab:embodied-memory-types}
\begin{tabularx}{\textwidth}{@{}l l >{\raggedright\arraybackslash}X >{\raggedright\arraybackslash}X >{\raggedright\arraybackslash}X@{}}
\toprule
\textbf{Type} & \textbf{Class} & \textbf{What is tested} & \textbf{Primary inputs} & \textbf{Evidence required} \\
\midrule
\SPref     & Semantic    & Cross-video tool/method/location preference
& Verb/object statistics; cross-video digests; sparse action sample
& Preference supported in ${\geq}2$ videos with repeated observations \\
\SHabit    & Semantic    & After trigger $X$, typical next action (contextual habit)
& Trigger$\rightarrow$next patterns; recurring action sequences; digests
& Trigger in ${\geq}2$ videos; habit repeated often enough to be habitual \\
\SRoutine  & Semantic    & Multi-step workflow or kitchen strategy (routine)
& Recurring multi-step workflows; verb-transition regularities; digests
& Strategy seen in ${\geq}2$ videos; distractors are distinct approaches \\\cdashline{1-5}[0.6pt/3pt]
\ERecall   & Episodic    & Main activity in a \emph{specific} video
& Per-video narrations $\rightarrow$ short activity label; plausible wrong activities
& One episodic item per video; answer tied to that \texttt{video\_id} only \\
\bottomrule
\end{tabularx}
\end{table*}

\paragraph{Scope of this appendix.}
This appendix specifies the annotation-grounded pipeline that produces the four reported EAM-QA types in Table~\ref{tab:embodied-memory-types}.
For each participant, we aggregate in-scope narrations, summarize them into compact statistics and per-video digests, prompt a generator LLM for structured multiple-choice items, and filter with a separate verifier model.
Two earlier pixel-grounded types (object localization \ELoc, temporal ordering \EOrder) are excluded from the released benchmark because parametric and perception-heavy baselines already saturate those probes.

\paragraph{Construction steps.}
For each participant, the pipeline applies the following steps; each step is selected to keep generation \emph{grounded} while fitting in context windows.

\textbf{(i) Index.}
We aggregate verb/noun marginals, verb--object pairs, consecutive verb bigrams, and trigger$\rightarrow$next histograms (within-video only).
Marginals surface stable tool/object preferences for \SPref{};
trigger--response tables directly support \SHabit{} (``after $X$, what next?''); verb bigrams expose coarse workflow regularities for \SRoutine{} without sending the full action log to the LLM.
These statistics are a \emph{lossy but faithful} summary: they highlight recurring structure the generator should mine, while staying far smaller than raw narrations.

\textbf{(ii) Summarize.}
Each in-scope video is compressed into a short Goal / key-sequence / object digest so cross-video types see thematic context (e.g., ``checks pan by stirring'') without loading thousands of raw rows.

\textbf{(iii) Single-hop generate.}
Per-type prompts combine statistics, a sparse action sample, video digests, and shared instructions requiring timestamp and \texttt{video\_id} citations, balanced distractors, and guards against tautological or commonsense-only items.
\SPref{}, \SHabit{}, and \SRoutine{} use a lower decoding temperature than multi-hop variants to limit hallucinated counts while preserving phrasing diversity.

\textbf{(iv) Multi-hop variants (within each type).}
Multi-hop is \emph{not} a fifth question family: every headline type also receives 2-hop and 3-hop items.
The pipeline first extracts 4--6 step action chains within a video, then prompts the generator to select a 2- or 3-step subchain and phrase a question whose answer depends on that subchain only.
Deterministic chain extraction guarantees the hops exist in the log; the model selects and words the probe.

\textbf{(v) Post-process.}
Rule deduplication removes exact duplicates; keyword topic deduplication is disabled in the reported release because it was too aggressive (${\sim}22\%$ loss).
LLM curation and contradiction resolution on \SPref/\SHabit{} drop near-duplicate or conflicting claims.
A per-participant cap (50\% per type) keeps the release balanced across cognitive classes.
Long runs can resume from intermediate save points after interruption.

\textbf{(vi) Construct E-correct controls.}
After fixing the $2{,}212$ answerable items, we create $551$ unanswerable
controls (approximately 20\% of the final panel) by deterministically sampling
a question from a different donor participant, rewriting participant mentions
to the target participant, and assigning option E as the correct answer. The
answer agent receives only the target participant's memory bank, so none of the
four donor-derived content options is supported there. The release records
\texttt{is\_unanswerable}, \texttt{source\_participant}, and
\texttt{target\_participant} for auditing this construction.

\paragraph{Per-type inputs, flow, and prompts.}

\begin{figure*}[t]
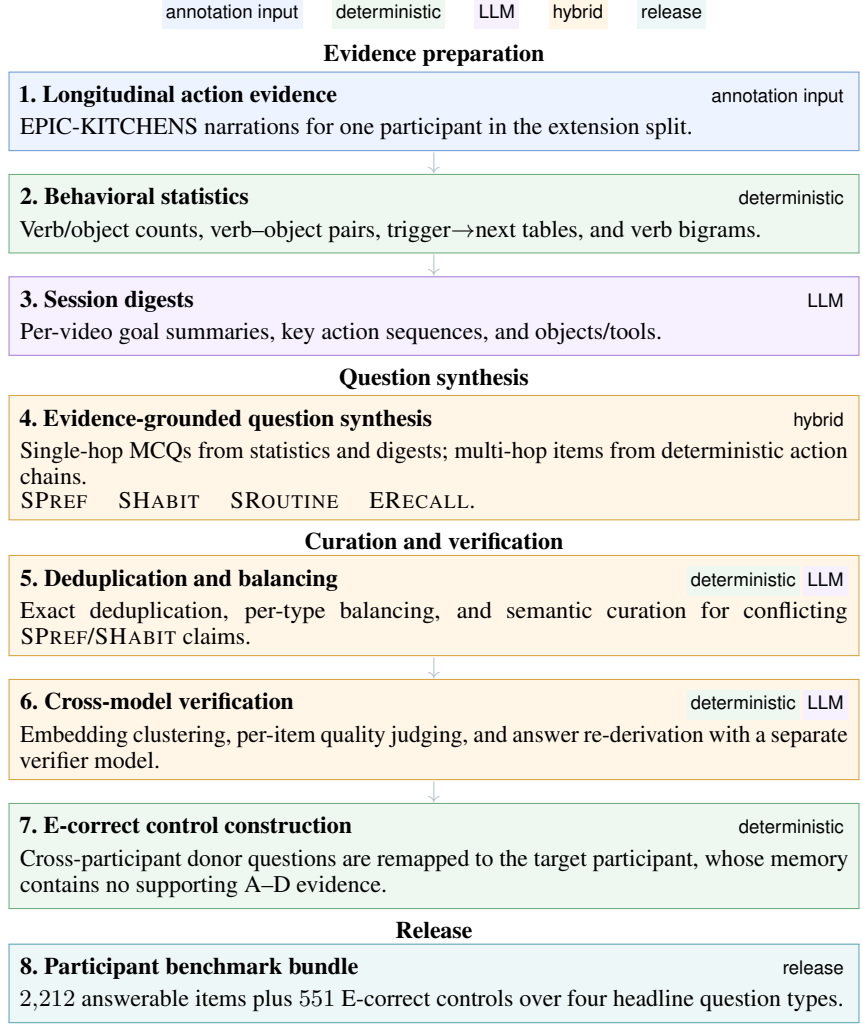

\centering
\small
\setlength{\fboxsep}{5pt}
\begin{minipage}{0.90\textwidth}
\centering
{\footnotesize
\flowpill{flowInputBg}{annotation input}
\quad \flowpill{flowRuleBg}{deterministic}
\quad \flowpill{flowLLMBg}{LLM}
\quad \flowpill{flowHybridBg}{hybrid}
\quad \flowpill{flowOutputBg}{release}
}
\\[0.45em]

\flowphase{Evidence preparation}
\flowstagecolor{flowInputBg}{flowInputFrame}{1.\ Longitudinal action evidence}
  {\flowpill{flowInputBg}{annotation input}}
  {EPIC-KITCHENS narrations for one participant in the extension split.}
\flowarrow
\flowstagecolor{flowRuleBg}{flowRuleFrame}{2.\ Behavioral statistics}
  {\flowpill{flowRuleBg}{deterministic}}
  {Verb/object counts, verb--object pairs, trigger$\rightarrow$next tables, and verb bigrams.}
\flowarrow
\flowstagecolor{flowLLMBg}{flowLLMFrame}{3.\ Session digests}
  {\flowpill{flowLLMBg}{LLM}}
  {Per-video goal summaries, key action sequences, and objects/tools.}

\vspace{0.28em}
\flowphase{Question synthesis}
\flowstagecolor{flowHybridBg}{flowHybridFrame}{4.\ Evidence-grounded question synthesis}
  {\flowpill{flowHybridBg}{hybrid}}
  {  Single-hop MCQs from statistics and digests; multi-hop items from deterministic action chains.\\[-0.1em]
  \SPref{} \quad \SHabit{} \quad \SRoutine{} \quad \ERecall{}.}

\vspace{0.28em}
\flowphase{Curation and verification}
\flowstagecolor{flowHybridBg}{flowHybridFrame}{5.\ Deduplication and balancing}
  {\flowpill{flowRuleBg}{deterministic}\,\flowpill{flowLLMBg}{LLM}}
  {Exact deduplication, per-type balancing, and semantic curation for conflicting \SPref{}/\SHabit{} claims.}
\flowarrow
\flowstagecolor{flowHybridBg}{flowHybridFrame}{6.\ Cross-model verification}
  {\flowpill{flowRuleBg}{deterministic}\,\flowpill{flowLLMBg}{LLM}}
  {Embedding clustering, per-item quality judging, and answer re-derivation with a separate verifier model.}
\flowarrow
\flowstagecolor{flowRuleBg}{flowRuleFrame}{7.\ E-correct control construction}
  {\flowpill{flowRuleBg}{deterministic}}
  {Cross-participant donor questions are remapped to the target participant, whose memory contains no supporting A--D evidence.}

\vspace{0.28em}
\flowphase{Release}
\flowstagecolor{flowOutputBg}{flowOutputFrame}{8.\ Participant benchmark bundle}
  {\flowpill{flowOutputBg}{release}}
  {$2{,}212$ answerable items plus $551$ E-correct controls over four headline question types.}
\end{minipage}
\caption{\textbf{MEMORA-Embodied Memory benchmark construction pipeline.}
MEMORA-Bench converts longitudinal action annotations into participant-specific Embodied Memory questions through four phases: evidence preparation, question synthesis, curation and verification, and release.
Colors indicate whether each stage is annotation input, deterministic processing, LLM-based synthesis, hybrid checking, or the released benchmark bundle.}
\label{fig:gen-pipeline-overview}
\end{figure*}

Figure~\ref{fig:gen-pipeline-overview} and Table~\ref{tab:embodied-memory-types} summarize the pipeline and per-type inputs and evidence thresholds for each headline type.
Every generation prompt appends the same three instruction blocks: \textbf{verifiability} (mandatory timestamps, \texttt{video\_id}s, verb--noun citations), \textbf{answer format} (balanced distractors, varied correct-letter positions), and \textbf{anti-failure} guards (tautology, triviality, hallucination, stem leakage).
Figure~\ref{fig:embodied-memory-prompt-skeletons} gives the auditable prompt skeletons.

\begin{figure*}[t]
\centering
\begin{minipage}[t]{0.48\textwidth}
\textbf{Shared verifiability rules}\vspace{0.2em}\par
\promptpanelbox{0.95\linewidth}{%
VERIFIABILITY REQUIREMENTS\\
In reasoning, cite:\\
1. timestamp(s): HH:MM:SS[.ms]\\
2. video\_id(s) where evidence appears\\
3. exact verb-noun labels from input\\
4. counts for ``typically'' claims\\
Never fabricate unseen actions or times.}
\end{minipage}
\hfill
\begin{minipage}[t]{0.48\textwidth}
\textbf{\SPref{}: preference}\vspace{0.2em}\par
\promptpanelbox{0.95\linewidth}{%
Goal: discover PERSONAL PREFERENCES.\\
Input: action statistics, sample actions, per-video digests.\\
Require: preference in >=2 videos.\\
Discover: tool/order/location/method/timing.\\
Output fields: question, preference\_type, activity\_topic, evidence.count, evidence.video\_ids, choices, correct\_answer, reasoning.}
\end{minipage}

\vspace{0.6em}

\begin{minipage}[t]{0.48\textwidth}
\textbf{\SHabit{}: contextual habit}\vspace{0.2em}\par
\promptpanelbox{0.95\linewidth}{%
Goal: find trigger -> response habits.\\
Input: statistics, common action sequences, video digests.\\
Require: trigger in >=2 videos; prefer trigger\_count>=4.\\
Question form: ``After [trigger], what does P typically do next?''\\
Output: trigger\_action, habit\_count, total\_trigger\_occurrences, video\_ids, choices, answer.}
\end{minipage}
\hfill
\begin{minipage}[t]{0.48\textwidth}
\textbf{\SRoutine{}: workflow strategy}\vspace{0.2em}\par
\promptpanelbox{0.95\linewidth}{%
Goal: identify strategic approaches, not generic step lists.\\
Input: recurring sequences, statistics, video digests.\\
Reject: universal flows such as pick-up->put-down or open->close.\\
Require: seen\_in\_videos >= 2.\\
Choices: four fundamentally different strategies.\\
Output: task\_topic, correct\_sequence, evidence, choices, answer.}
\end{minipage}

\vspace{0.6em}

\begin{minipage}[t]{0.98\textwidth}
\textbf{\ERecall{}: video-specific episodic recall}\vspace{0.2em}\par
\promptpanelbox{0.98\linewidth}{%
Step A: summarize each video from narrations into short memory\_summary + main\_activity.\\
Step B: generate three plausible but wrong activities from varied categories.\\
Assemble one 4-way MCQ per video\_id. Multi-hop \ERecall{} uses the same schema but targets an outcome chain of depth 2 or 3 inside a single video.}
\end{minipage}
\caption{\textbf{Prompt skeletons for Embodied Memory question generation.}
All generation prompts append the shared verifiability rules; \SPref{}, \SHabit{}, \SRoutine{}, and \ERecall{} differ in the evidence pattern they require and the schema fields they emit. Placeholders \texttt{\{P\}} and \texttt{\{N\}} are filled per run, and multi-hop variants reuse the same framing with supplied action chains.}
\label{fig:embodied-memory-prompt-skeletons}
\end{figure*}

\subsection{Quality Control and Diagnostics}

\paragraph{Summarization and multi-hop wording.}
The summarization prompt requires each video digest to list a one-sentence goal, 2--5 titled arrow sequences, and key objects/tools.
For multi-hop items in \SPref{}, \SHabit{}, \SRoutine{}, and \ERecall{}, the generator receives a timestamped within-video chain and must select a 2- or 3-step subchain before writing the question; released items are tagged with \texttt{reasoning\_depth}${\in}\{2,3\}$ but keep the parent type's schema.

\paragraph{Automated verification before release.}
Every candidate passes three filter stages, each targeting a different failure mode:
\textbf{(1)~Semantic dedup} (embedder in Table~\ref{tab:embodied-memory-generation}; cosine $\geq 0.95$) groups near-duplicate stems before expensive judging; an LLM picks which cluster members to keep so we remove redundancy without collapsing legitimately distinct probes.
\textbf{(2)~Quality filter} ($T{=}0$): a type-aware \textsc{yes}/\textsc{no} prompt rejects unclear stems, weak distractors, or answers not supported by cited evidence; a soft retry at $T{=}0.3$ rescues borderline items if the first pass is overly strict (${<}30\%$ pass rate).
\textbf{(3)~Answer re-derivation}: the judge sees the question, choices, and evidence but \emph{not} the stored label and must independently recover the same letter---catching cases where the generator assigned a key inconsistent with its own evidence.
Rule-based gates at generation time enforce minimum support (e.g., $\geq 2$ videos for cross-video types, minimum trigger counts for \SHabit{}) so obviously weak candidates never reach the judge.
On a reference run for P06, 173 candidates survived in-generator steps and 129 were released after filtering (27 removed by the quality judge alone).
Representative numeric thresholds are listed in Table~\ref{tab:embodied-memory-generation}.

\paragraph{OpenAI API configuration.}
The reported annotation-grounded release is produced through the \textbf{OpenAI Chat Completions API} (\texttt{https://api.openai.com/v1}).
Each request is a single-turn user message (no system role, no dialogue history).
We use \texttt{gpt-5.5} for generation and \texttt{gpt-4o-mini} for verification so acceptance is not self-judged.
Structured outputs use JSON mode when the prompt requests JSON; the verifier uses low-temperature decoding ($T{=}0$, \texttt{top\_p}$=0.9$).
The generator is a reasoning-style model: the API ignores custom temperature and instead allocates a larger completion budget so hidden reasoning tokens do not crowd out the visible answer.
Operational settings are 3 concurrent requests, 240\,s timeout, up to 3 retries with exponential backoff, and a 128k-token context budget with prompt truncation when needed.

\begin{table*}[t]
\centering
\footnotesize
\setlength{\tabcolsep}{3pt}
\renewcommand{\arraystretch}{1.06}
\caption{Models and decoding settings for Embodied Memory benchmark generation.
For \texttt{gpt-5.5}, listed $T$ values document intent; the API uses its built-in sampling and an expanded completion cap.}
\label{tab:embodied-memory-generation}
\begin{tabularx}{\textwidth}{@{}l l c c X@{}}
\toprule
\textbf{Stage / role} & \textbf{Model} & \textbf{$T$} & \textbf{Max out.} & \textbf{Notes} \\
\midrule
Generator   & \texttt{gpt-5.5}     & ---  & ---   & Question synthesis and post-gen.\ curation \\
Verifier    & \texttt{gpt-4o-mini} & 0.0  & varies& Per-item quality check and answer re-derivation \\
Embed dedup & \texttt{e5-base-v2}~\citep{wang2024textembeddingsweaklysupervisedcontrastive} & --- & --- & Local embedding model (not an API call) \\
\midrule
Video summarization           & \texttt{gpt-5.5} & 0.1 & 512  & Generation stage \\
\SPref{}/\SHabit{}/\SRoutine{} (single-hop) & \texttt{gpt-5.5} & 0.3 & 6144 & Generation stage \\
\ERecall{} activity label     & \texttt{gpt-5.5} & 0.4 & 400  & Generation stage \\
\ERecall{} distractors        & \texttt{gpt-5.5} & 0.7 & 2048 & Generation stage \\
Multi-hop (all four types)    & \texttt{gpt-5.5} & 0.7 & 6144 & Generation stage \\
Post-generation curation      & \texttt{gpt-5.5} & 0.1 & 1000 & Generation stage \\
Verifier: semantic-dedup pick & \texttt{gpt-4o-mini} & 0.0 & 32 & Verification stage \\
Verifier: quality filter      & \texttt{gpt-4o-mini} & 0.0 & 16 & Verification stage \\
Verifier: answer re-derivation& \texttt{gpt-4o-mini} & 0.0 & 32 & Verification stage \\
\bottomrule
\end{tabularx}
\end{table*}

Embedding dedup uses the embedder in Table~\ref{tab:embodied-memory-generation} before any verifier API calls.

\paragraph{Reproducibility.}
We generate one bundle per participant on the extension video whitelist; per-type targets are 8 single-hop and 8 multi-hop items per semantic type (\SPref{}, \SHabit{}, \SRoutine{}, before $\times 3$ overproduction) and one \ERecall{} item per in-scope video.
Full launch parameters and record schemas are recorded in the MEMORA-Bench reproduction appendix.

\subsection{Evaluation Protocol}
\label{app:mem-bench-eval}

\paragraph{Task and metrics.}
For each question $q$, the agent must choose one of five options $\{\mathrm{A},\ldots,\mathrm{E}\}$.
Options A--D are contentful answers; \textbf{E} denotes ``information not available,'' so a model must recognize missing memory rather than guess among four plausible kitchen actions---otherwise parametric priors inflate accuracy on underspecified probes.
\textbf{Accuracy} is reported per type and macro-aggregated over participants; the Pool column pools all headline items with count weighting.
Each question is routed to the fixed memory bank of its declared target
participant. Answerable items are grounded in that participant's observed
corpus; E-correct controls are donor-derived questions for which the target
memory intentionally contains no supporting content option. The released
headline records do not define a per-question \texttt{ask\_turn\_id}, so this
evaluation makes no claim of within-video prefix rollback.

\paragraph{Controlled comparison.}
All conditions share the same perception front-end, dense retrieval backend, answer-agent backbone, and ReAct budget within a column (Appendix~\ref{app:impl}), so differences are not confounded by vision, embedding capacity, or inference budget.
The compared memory interfaces are Parametric (no external memory), \textbf{Flat-1D}~\citep{zeng2022socraticmodelscomposingzeroshot} (chronological text memory exposed via a flat \texttt{search} tool), \textbf{Graph-2D} (an entity-relation graph memory instantiated in the style of M3-Agent~\citep{long2025seeinglisteningrememberingreasoning}), and \textbf{MEMORA} (typed substrates with consolidation variants).
Cross-family gaps therefore reflect memory organization together with its intended read interface; raw/processed and episodic/full pairs provide the controlled tests of editing and consolidation.

\paragraph{Adapted memory-organization check.}
We additionally compare two recent organizational designs inside a matched
MEMORA-Bench harness. Graph-2D is the entity-centric graph instantiation in the
style of M3-Agent; the adapted MEMENTO organization~\citep{kwon2026embodiedagentsmeetpersonalization}
partitions memory by participant and groups evidence around object semantics
and recurring user patterns. Both adaptations reuse the same upstream
perception records, retrieval backend, answer backbone, and evaluation budget;
the changed component is the organization presented to retrieval. This is an
in-domain architectural adaptation, not an evaluation on the external MEMENTO
benchmark, whose input consists of already processed memory entries.
Table~\ref{tab:adapted-memory-organizations} reports the matched P01--P04
EAM-QA slice and the 153-task Generalize planning set.

\begin{table}[t]
\centering
\small
\setlength{\tabcolsep}{4pt}
\caption{Matched adapted memory organizations. EAM-QA uses P01--P04 and the
four released EAM-QA types; E-rate is option-E selection. Planning reports
RGP on 153 Generalize tasks. Graph-2D is the M3-Agent-style organization used
throughout our figures and tables, not an additional baseline.}
\label{tab:adapted-memory-organizations}
\begin{tabular}{@{}lrrr@{}}
\toprule
\textbf{Organization} & \textbf{EAM-QA Acc.} & \textbf{E-rate} & \textbf{RGP} \\
\midrule
Flat-1D                & 37.2 & 65.8 & .386 \\
Graph-2D               & 34.9 & 70.6 & .336 \\
MEMENTO (adapted)      & 39.8 & 48.3 & .370 \\
MEMORA                 & \textbf{42.0} & \textbf{35.7} & \textbf{.450} \\
\bottomrule
\end{tabular}
\end{table}

\paragraph{Agent loop.}
The answering agent follows the shared Embodied Memory ReAct budget in Appendix~\ref{app:impl} (up to five \texttt{search} calls, then one letter).
For Embodied Memory evaluation, the harness loads the complete fixed memory
bank for the question's target participant. No per-question temporal slicing
is applied to the released headline panel; protection against unsupported
answers is instead measured by the E-correct controls described above.
Tool schemas and per-condition system prompts are fixed within each block; Graph-2D receives schema-equivalent per-type prompts so routing verbosity does not confound substrate comparisons.

\paragraph{Reporting diagnostics.}
A \textbf{Parametric Memory} column is always reported alongside structured memory: types where Parametric already exceeds ${\sim}70\%$ are flagged as prior-saturated in footnotes rather than excluded post hoc.
E-selection rate is tracked separately because flat retrieval can overproduce information-not-available answers on some commercial backbones.

\paragraph{Illustrative items.}
Figure~\ref{fig:embodied-memory-examples} shows the construction-time audit
record corresponding to one released answerable item per headline type from
participant P06. Evidence counts and reasoning depth are used during
construction and quality filtering but are stripped from the public task
record. The answering agent receives the question and the five released
choices A--E, without the evidence or reasoning metadata.

\begin{figure*}[t]
\centering
\begin{minipage}[t]{0.48\textwidth}
\textbf{\SPref{}: cross-video preference}\vspace{0.2em}\par
\begin{lstlisting}[style=compactjson]
{
  "qa_type": "user_preference",
  "participant_id": "P06",
  "question": "Across videos, when P06 checks food cooking in a pan, what observable method do they typically use?",
  "choices": [
    "A) Stir the pan contents",
    "B) Shake the pan briefly",
    "C) Leave pan untouched",
    "D) Pour food into bowl"
  ],
  "correct_answer": "A",
  "evidence": {
    "correct_option_count": 26,
    "video_ids": ["P06_101", "P06_102", "P06_103", "P06_104", "P06_105"]
  },
  "reasoning_depth": 1
}
\end{lstlisting}
\end{minipage}
\hfill
\begin{minipage}[t]{0.48\textwidth}
\textbf{\SHabit{}: trigger$\rightarrow$response habit}\vspace{0.2em}\par
\begin{lstlisting}[style=compactjson]
{
  "qa_type": "contextual_habit",
  "participant_id": "P06",
  "question": "Across pan-checking moments, what does P06 typically do immediately after picking up the spoon?",
  "choices": [
    "A) open drawer for another utensil",
    "B) stir pan before touching the lid",
    "C) take-off lid from the pan",
    "D) put-down spoon beside the hob"
  ],
  "correct_answer": "C",
  "evidence": {
    "correct_habit_count": 5,
    "total_trigger_occurrences": 15,
    "video_ids": ["P06_103", "P06_102", "P06_101"]
  },
  "reasoning_depth": 1
}
\end{lstlisting}
\end{minipage}

\vspace{0.6em}

\begin{minipage}[t]{0.48\textwidth}
\textbf{\SRoutine{}: cross-video workflow}\vspace{0.2em}\par
\begin{lstlisting}[style=compactjson]
{
  "qa_type": "action_pattern",
  "participant_id": "P06",
  "question": "During active hob cooking, what is P06's typical control routine?",
  "choices": [
    "A) Uses a spoon-led lid check, stirs, re-covers, then tastes.",
    "B) Shakes the pan with the lid on and avoids spoon testing.",
    "C) Transfers the contents to a bowl before checking doneness.",
    "D) Leaves the pan uncovered and adjusts heat without sampling."
  ],
  "correct_answer": "A",
  "evidence": {
    "seen_in_videos": ["P06_103", "P06_102", "P06_101"],
    "occurrences": 3
  },
  "reasoning_depth": 1
}
\end{lstlisting}
\end{minipage}
\hfill
\begin{minipage}[t]{0.48\textwidth}
\textbf{\ERecall{}: video-specific episodic recall ($d{=}2$)}\vspace{0.2em}\par
\begin{lstlisting}[style=compactjson]
{
  "qa_type": "event_recall",
  "participant_id": "P06",
  "video_id": "P06_104",
  "question": "In video P06_104, P06 handled the sweet potato. What did they do with it?",
  "choices": [
    "A) They mashed the sweet potato and put it into the pan.",
    "B) They cut the sweet potato and left it beside the pan.",
    "C) They cut the sweet potato and put it in the pan.",
    "D) They pierced the sweet potato and kept it on the fork."
  ],
  "correct_answer": "C",
  "evidence": {
    "video_id": "P06_104",
    "core_chain": [
      {"action": "cut potato:sweet", "idx": 5},
      {"action": "put-in potato:sweet", "idx": 6}
    ]
  },
  "reasoning_depth": 2
}
\end{lstlisting}
\end{minipage}
\caption{Construction-time audit views of released Embodied Memory items (P06).
Left column: \SPref{} and \SRoutine{}; right column: \SHabit{} and \ERecall{}.
For compact display the figure shows only content options A--D; every released
JSON record already contains option~E (information not available), while the
displayed evidence metadata is omitted from the public task input.}
\label{fig:embodied-memory-examples}
\end{figure*}

\subsection{Full and Conditional EAM-QA Evaluation Views}
\label{app:eamqa-evaluation-views}

The Embodied Memory benchmark mixes \emph{answerable} questions (ground truth $\in\{\mathrm{A,B,C,D}\}$) with \emph{unanswerable} probes (ground truth $\mathrm{E}$), the latter required so a model cannot inflate accuracy by guessing on underspecified items.
We report three views with explicit predicates. The \textbf{full benchmark}
contains all $2{,}763$ items. The \textbf{MEMORA-committed} view retains items
for which MEMORA selects A--D. The \textbf{parametric-missed +
MEMORA-committed} view additionally requires the Parametric condition to be
incorrect. All compared conditions are scored on the same retained IDs within
each backbone.

These latter two views are conditional diagnostics, not alternative benchmark
definitions. Parametric failure is a proxy for weak closed-book support, not a
proof that a question is experience-dependent; likewise, selecting A--D is a
commitment event, not by itself proof that the answer is grounded. The filter
does not condition on MEMORA correctness, so an E-correct control on which
MEMORA commits incorrectly remains in the diagnostic and is scored as an
error. This makes the selection rule auditable while avoiding a correctness-
conditioned subset. Table~\ref{tab:filter-sensitivity} reports all three views;
the Parametric row is $0\%$ in the final conditional view by construction and
is therefore omitted from the per-backbone table.

\paragraph{Commit rate and commit-conditioned accuracy.}
Within the parametric-missed pool, we report the MEMORA commit rate
$\mathrm{P}(\text{MEMORA picks A--D}\mid\text{parametric wrong})$ and
commit-conditioned accuracy
$\mathrm{P}(\text{correct}\mid\text{A--D},\text{parametric wrong})$.
The latter is the final conditional-view accuracy by construction. It rises
from Gemma-4-26B-A4B $54.1\%$ to Qwen3.6-27B $69.1\%$ and Gemma-4-31B
$74.5\%$, whereas commit rate varies non-monotonically (respectively $67.5\%$,
$70.3\%$, and $50.7\%$). We therefore report both quantities rather than
interpreting A--D selection alone as successful grounding.
We read this as a \emph{capacity-bound} rather than \emph{pipeline-bound} cross-backbone pattern: absolute differences across backbones reflect the answer-agent's discriminative capacity once it has enough support to answer, not a failure of the memory pipeline that feeds it.

\begin{table*}[t]
\centering
\caption{%
\textbf{EAM-QA evaluation views: MEMORA overall accuracy (\%) and gap to strongest non-MEMORA condition} (\textsc{Flat-1D}/\textsc{Graph-2D}, raw or after MEMORA's online editing pass; MEMORA-Episodic excluded as it is a MEMORA ablation rather than a baseline).
The full-benchmark view reports all EAM-QA questions; the MEMORA-committed view
retains cases where MEMORA selects A--D; the final conditional view additionally
requires the parametric baseline to be incorrect.
}
\label{tab:filter-sensitivity}
\small
\setlength{\tabcolsep}{2.4pt}
\renewcommand{\arraystretch}{1.10}
\begin{tabular}{@{}l rr rr rr rr@{}}
\toprule
& \multicolumn{2}{c}{\textbf{Q3.6-35B-A3B}$^{\sharp}$} & \multicolumn{2}{c}{\textbf{G4-26B-A4B}} & \multicolumn{2}{c}{\textbf{Q3.6-27B}} & \multicolumn{2}{c}{\textbf{G4-31B}} \\
\cmidrule(lr){2-3}\cmidrule(lr){4-5}\cmidrule(lr){6-7}\cmidrule(lr){8-9}
\textbf{Evaluation view}                                             & MEMORA & $\Delta$best  & MEMORA & $\Delta$best  & MEMORA & $\Delta$best  & MEMORA & $\Delta$best \\
\midrule
Full benchmark                                                        & 50.2   & $-3.4$        & 39.3   & $-9.9$        & 51.7   & $+1.5$        & 46.1   & $-0.5$ \\
MEMORA-committed                                                      & 56.5   & $-0.3$        & 45.8   & $-7.2$        & 61.3   & $+5.9$        & 67.7   & $+9.6$ \\
\rowcolor{black!4}
Parametric-missed + MEMORA-committed                                 & \textbf{63.7} & $\mathbf{+11.0}$ & \textbf{54.1} & $\mathbf{+9.6}$ & \textbf{69.1} & $\mathbf{+19.3}$ & \textbf{74.5} & $\mathbf{+20.5}$ \\
\midrule
Net swing (full $\to$ headline)                                       & \multicolumn{2}{c}{$+14.4$ points} & \multicolumn{2}{c}{$+19.5$ points} & \multicolumn{2}{c}{$+17.8$ points} & \multicolumn{2}{c}{$+21.0$ points} \\
\bottomrule
\end{tabular}

\vspace{2pt}
\raggedright
{\footnotesize
Subset sizes per backbone ($n_{\text{full}} / n_{\text{answerable}} / n_{\text{headline}}$):
Q3.6-35B-A3B $1{,}202 / 819 / 721$;
G4-26B-A4B $2{,}763 / 1{,}770 / 1{,}497$;
Q3.6-27B $2{,}763 / 1{,}753 / 1{,}556$;
G4-31B $2{,}763 / 1{,}234 / 1{,}122$.
$\Delta$best $= $ MEMORA $-$ best of \{\textsc{Flat-1D} raw/processed, \textsc{Graph-2D} raw/processed\} on the same subset.
\textsuperscript{$\sharp$}\,Q3.6-35B-A3B panel is reported on the $8$-PID intersection ($n_{\text{full}}{=}1{,}202$) for which this diagnostic was first computed; the headline row agrees with the $12$-PID Qwen3.6-35B-A3B intersection in Table~\ref{tab:eamqa-per-backbone} (MEMORA $63.5\%$, $+11.8$ points over the strongest non-MEMORA condition) within $\pm 0.8$ points.
}
\end{table*}

\subsection{Per-backbone $\times$ per-type Results}
\label{app:per-backbone-per-type}

The main paper headlines a $2{\times}2$ open-source cross-backbone design (Qwen 3.6 / Gemma 4 $\times$ MoE / dense; Table~\ref{tab:memorization-main} displays the Gemma-4-31B-it column on the full $18$-PID panel).
This subsection unpacks the remaining three backbones --- Qwen3.6-35B-A3B, Qwen3.6-27B, and Gemma-4-26B-A4B-it --- under the unified parametric-missed, MEMORA-committed diagnostic of Appendix~\ref{app:eamqa-evaluation-views} (Table~\ref{tab:eamqa-per-backbone}; the per-backbone $\Delta_{\text{consol}}$ rows summarize the per-type lift).

\paragraph{Cross-backbone consistency of the consolidation lift.}
Under the unified parametric-missed, MEMORA-committed EAM-QA diagnostic (Table~\ref{tab:eamqa-per-backbone}, $\Delta_{\text{consol}}$ rows), the consolidation lift $\Delta_{\text{consol}} = $ MEMORA $-$ MEMORA-Episodic is positive on every question type and every backbone.
The \SRoutine{} lift is the most robust as expected for the aggregation-requiring type: $+4.9$ on Qwen3.6-35B-A3B, $+9.9$ on Gemma-4-26B-A4B, $+10.7$ on Qwen3.6-27B, and $+24.6$ points on Gemma-4-31B.
\SHabit{}, the other cross-event aggregation type, lifts by $+6.1$ to $+13.7$ points across the four backbones; the single-fact retrieval types lift by $+7.0$ to $+11.3$ points (\SPref{}) and $+4.7$ to $+10.5$ points (\ERecall{}).
The four-backbone means are $+12.5$/$+10.1$/$+9.2$/$+7.4$ points on \SRoutine{}/\SHabit{}/\SPref{}/\ERecall{}, with every per-backbone, per-type cell positive, consistent with consolidation strengthening retrieval whenever the question can be committed to a contentful answer.
\begin{table*}[t]
\centering
\caption{%
\textbf{MEMORA-Embodied Memory Assessment accuracy (\%): per-backbone, per-type panel}.
All four backbones report under the parametric-missed, MEMORA-committed EAM-QA diagnostic (Appendix~\ref{app:eamqa-evaluation-views}):
\textbf{Gemma-4-31B-it} on the full $18$-PID panel ($N{=}1{,}122$; mirrors Table~\ref{tab:memorization-main});
\textbf{Qwen3.6-35B-A3B} on the $12$-PID intersection where all seven conditions are currently materialized ($N{=}1{,}141$);
\textbf{Qwen3.6-27B} on the full $18$-PID panel ($N{=}1{,}556$);
\textbf{Gemma-4-26B-A4B-it} on the full $18$-PID panel ($N{=}1{,}497$).
$\Delta_{\text{consol}}$ row: MEMORA minus MEMORA-Episodic.
The Parametric (no-memory) row is omitted because the diagnostic retains only questions it answers incorrectly ($0\%$ by construction); full-benchmark and MEMORA-committed results are reported in Table~\ref{tab:filter-sensitivity}.
}
\label{tab:eamqa-per-backbone}
\small
\setlength{\tabcolsep}{3.6pt}
\renewcommand{\arraystretch}{1.08}
\begin{tabular}{@{}l l rrr r r@{}}
\toprule
                                                        &                                                        & \multicolumn{3}{c}{\textbf{Cross-video (semantic)}}                              & \textbf{Single-video}                       &                  \\
\cmidrule(lr){3-5}\cmidrule(lr){6-6}
\textbf{Backbone}                                       & \textbf{Memory condition}                              & \textbf{\SPref{}} & \textbf{\SHabit{}} & \textbf{\SRoutine{}} & \textbf{\ERecall{}} & \textbf{Overall} \\
\midrule
\multirow{7}{*}{\shortstack[l]{Qwen3.6-35B-A3B$^{\flat}$\\ \emph{(MoE, $64$K ctx)}\\ \emph{($N{=}1{,}141$)}}}
& Flat-1D (raw)                       & 51.2           & 33.5           & 53.2           & 12.5           & 41.6           \\
& Flat-1D (processed)                 & 53.1           & 34.2           & 52.9           & \phantom{0}5.3 & 41.3           \\
& Graph-2D (raw)                      & 50.6           & 27.4           & 54.7           & 48.7           & 45.0           \\
& Graph-2D (processed)                & 61.5           & 34.2           & 59.6           & 50.7           & 51.7           \\
& MEMORA-Episodic                     & 64.3           & 42.2           & 63.5           & 53.3           & 56.3           \\
& \cellcolor{black!4}\textbf{MEMORA}  & \cellcolor{black!4}\textbf{73.3} & \cellcolor{black!4}\textbf{48.3} & \cellcolor{black!4}\textbf{68.4} & \cellcolor{black!4}\textbf{63.8} & \cellcolor{black!4}\textbf{63.5} \\
& $\Delta_{\text{consol}}$            & $\mathbf{+9.0}$ & $\mathbf{+6.1}$ & $\mathbf{+4.9}$ & $\mathbf{+10.5}$ & $\mathbf{+7.2}$ \\
\midrule
\multirow{7}{*}{\shortstack[l]{Qwen3.6-27B$^{\flat}$\\ \emph{(dense, $64$K ctx)}\\ \emph{($N{=}1{,}556$)}}}
& Flat-1D (raw)                       & 48.9           & 41.6           & 49.6           & 65.7           & 49.6           \\
& Flat-1D (processed)                 & 49.8           & 38.7           & 49.4           & 61.0           & 48.5           \\
& Graph-2D (raw)                      & 51.1           & 29.6           & 53.3           & 53.5           & 46.8           \\
& Graph-2D (processed)                & 55.1           & 32.7           & 55.4           & 55.9           & 49.7           \\
& MEMORA-Episodic                     & 64.1           & 48.8           & 64.0           & 68.5           & 60.9           \\
& \cellcolor{black!4}\textbf{MEMORA}  & \cellcolor{black!4}\textbf{71.1} & \cellcolor{black!4}\textbf{57.1} & \cellcolor{black!4}\textbf{74.8} & \cellcolor{black!4}\textbf{73.2} & \cellcolor{black!4}\textbf{69.1} \\
& $\Delta_{\text{consol}}$            & $\mathbf{+7.0}$ & $\mathbf{+8.3}$ & $\mathbf{+10.7}$ & $+4.7$         & $\mathbf{+8.2}$ \\
\midrule
\multirow{7}{*}{\shortstack[l]{Gemma-4-26B-A4B$^{\flat}$\\ \emph{(MoE, $64$K ctx)}\\ \emph{($N{=}1{,}497$)}}}
& Flat-1D (raw)                       & 51.0           & 28.0           & 42.0           & 59.4           & 42.4           \\
& Flat-1D (processed)                 & 53.9           & 31.3           & 42.2           & 62.4           & 44.5           \\
& Graph-2D (raw)                      & 40.4           & 16.3           & 31.6           & 60.2           & 32.5           \\
& Graph-2D (processed)                & 48.8           & 23.4           & 40.8           & 60.2           & 40.1           \\
& MEMORA-Episodic                     & 45.3           & 39.2           & 39.6           & 62.4           & 43.2           \\
& \cellcolor{black!4}\textbf{MEMORA}  & \cellcolor{black!4}\textbf{56.5} & \cellcolor{black!4}\textbf{51.7} & \cellcolor{black!4}\textbf{49.5} & \cellcolor{black!4}\textbf{70.7} & \cellcolor{black!4}\textbf{54.1} \\
& $\Delta_{\text{consol}}$            & $\mathbf{+11.3}$ & $\mathbf{+12.4}$ & $\mathbf{+9.9}$ & $\mathbf{+8.3}$ & $\mathbf{+10.9}$ \\
\midrule
\multirow{7}{*}{\shortstack[l]{Gemma-4-31B$^{\flat}$\\ \emph{(dense, $64$K ctx)}\\ \emph{($N{=}1{,}122$)}}}
& Flat-1D (raw)                       & 63.2           & 36.8           & 53.8           & 51.2           & 52.0           \\
& Flat-1D (processed)                 & 62.4           & 41.4           & 56.8           & 53.0           & 54.0           \\
& Graph-2D (raw)                      & 50.9           & 21.4           & 37.1           & 55.4           & 40.0           \\
& Graph-2D (processed)                & 57.1           & 27.0           & 48.6           & 57.1           & 47.0           \\
& MEMORA-Episodic                     & 67.4           & 47.0           & 56.8           & 73.8           & 60.1           \\
& \cellcolor{black!4}\textbf{MEMORA}  & \cellcolor{black!4}\textbf{76.8} & \cellcolor{black!4}\textbf{60.7} & \cellcolor{black!4}\textbf{81.5} & \cellcolor{black!4}\textbf{79.8} & \cellcolor{black!4}\textbf{74.5} \\
& $\Delta_{\text{consol}}$            & $\mathbf{+9.4}$ & $\mathbf{+13.7}$ & $\mathbf{+24.6}$ & $\mathbf{+6.0}$ & $\mathbf{+14.4}$ \\
\bottomrule
\end{tabular}

\vspace{2pt}
\raggedright
{\footnotesize
\textbf{Bold}: best per column within each backbone block.
$\Delta_{\text{consol}}$ values are bolded when they exceed the $\pm 4$--$5$ points per-type binomial $95\%$ CI.
\textsuperscript{$\flat$}\,Parametric-missed, MEMORA-committed diagnostic (Appendix~\ref{app:eamqa-evaluation-views}): Gemma-4-31B-it on the full $18$-PID EAM-QA panel ($N{=}1{,}122$; mirrors Table~\ref{tab:memorization-main}); Qwen3.6-35B-A3B on the $12$-PID intersection ($N{=}1{,}141$); Qwen3.6-27B on the full $18$-PID panel ($N{=}1{,}556$); Gemma-4-26B-A4B-it on the full $18$-PID panel ($N{=}1{,}497$).
The Qwen3.6-35B-A3B row is reported on the $12$-PID intersection because that is the largest subset in the public release on which all seven memory conditions are materialized. The $8$-PID sensitivity subset reported in the $^{\sharp}$ footnote of Table~\ref{tab:filter-sensitivity} confirms that MEMORA's accuracy and its $+11.8$-point lead over the strongest controlled baseline on Qwen3.6-35B-A3B are preserved within $\pm 0.8$ points across the $8$-PID and $12$-PID subsets.
}
\end{table*}

\paragraph{Scaling to larger corpora.}
The pipeline in Figure~\ref{fig:gen-pipeline-overview} is \emph{participant-agnostic}: adding more benchmark volume means repeating steps 1--7 for additional people or sessions, not redesigning the question types.
To extend beyond the current extension split, a new corpus must supply at minimum:
\begin{itemize}\setlength{\itemsep}{0.1em}
\item \textbf{Participant identifiers} that are stable across sessions (so cross-video \SPref{}, \SHabit{}, and \SRoutine{} items remain meaningful);
\item \textbf{Video/session identifiers} mapping each file to exactly one participant;
\item \textbf{Time-ordered action annotations} per video: timestamps plus verb--object (or equivalent) labels, and preferably short natural-language descriptions for audit trails;
\item A defined \textbf{video whitelist} (or split policy) stating which sessions enter generation;
\item For \textbf{evaluation} on new footage: synchronized RGB video and the same perception/retrieval stack as in the main experiments (generation can remain annotation-only).
\end{itemize}
Optional enrichments---frame-level bounding boxes, object tracks, or a parallel VLM-augmented track---are not required for the headline annotation-grounded release but would enable spatial types (\ELoc) or pixel-grounded verification at larger scale.
As annotation density and participant count grow, the same overgenerate-and-filter recipe applies; cost scales with LLM verifier calls per candidate, which is why semantic dedup and rule-based gates precede the cross-model judge.

%% file: sections_new/Appendix_sections/MEMORA_Planning.tex

\section{MEMORA-Planning: Construction and Evaluation Protocol}
\label{app:planning}
\label{app:planning-full}

Appendix~\ref{app:planning} specifies the prospective planning evaluation in MEMORA-Bench.
\textbf{MEMORA-Planning} evaluates whether the constructed memory bank supports future action, rather than only retrospective question answering.
In contrast to the Embodied Memory assessment, which asks an agent to select a discrete answer about past
experience, MEMORA-Planning requires the agent to synthesize an ordered kitchen plan
for a specific participant. Each item is grounded in EPIC-KITCHENS-100
annotations~\citep{damen2022epic}: the reference sequence is an observed
segment of narrated actions, but the planner is evaluated on whether it can
produce an executable, participant-specific procedure rather than reproduce the
annotation verbatim.

The reported benchmark uses the released MEMORA-Planning split.  Its tasks are
generated deterministically from EPIC action annotations, restricted to videos
available in the participant's consolidated \InfMem{}, and enriched with object
metadata from the same memory.  No LLM is used to generate task labels or task
queries in this release.  The evaluation compares no-memory, linear-memory, and
structured-memory agents under the same planner model, and reports rule-based
metrics to avoid judging an LLM agent with another LLM.

\begin{figure*}[t]
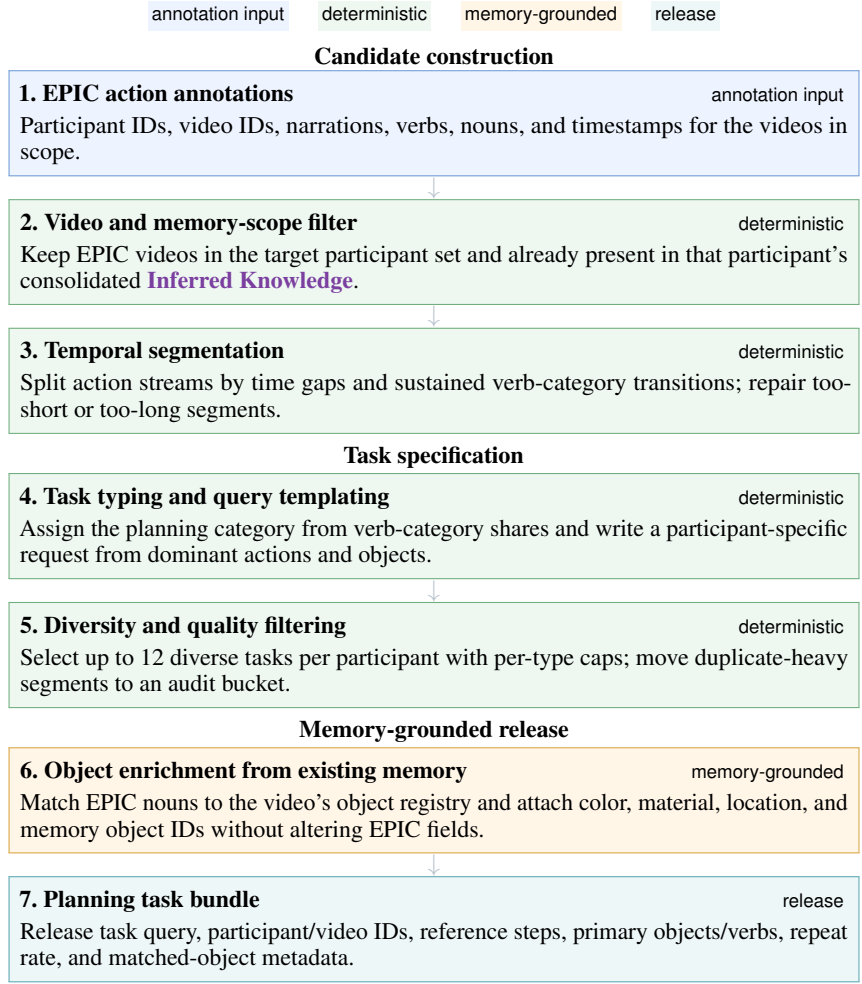

\centering
\small
\setlength{\fboxsep}{5pt}
\begin{minipage}{0.90\textwidth}
\centering
{\footnotesize
\flowpill{flowInputBg}{annotation input}
\quad \flowpill{flowRuleBg}{deterministic}
\quad \flowpill{flowHybridBg}{memory-grounded}
\quad \flowpill{flowOutputBg}{release}
}
\\[0.50em]

\flowphase{Candidate construction}
\flowstagecolor{flowInputBg}{flowInputFrame}{1.\ EPIC action annotations}
  {\flowpill{flowInputBg}{annotation input}}
  {Participant IDs, video IDs, narrations, verbs, nouns, and timestamps for the videos in scope.}
\flowarrow
\flowstagecolor{flowRuleBg}{flowRuleFrame}{2.\ Video and memory-scope filter}
  {\flowpill{flowRuleBg}{deterministic}}
  {Keep EPIC videos in the target participant set and already present in that participant's consolidated \InfMem{}.}
\flowarrow
\flowstagecolor{flowRuleBg}{flowRuleFrame}{3.\ Temporal segmentation}
  {\flowpill{flowRuleBg}{deterministic}}
  {Split action streams by time gaps and sustained verb-category transitions; repair too-short or too-long segments.}

\vspace{0.28em}
\flowphase{Task specification}
\flowstagecolor{flowRuleBg}{flowRuleFrame}{4.\ Task typing and query templating}
  {\flowpill{flowRuleBg}{deterministic}}
  {Assign the planning category from verb-category shares and write a participant-specific request from dominant actions and objects.}
\flowarrow
\flowstagecolor{flowRuleBg}{flowRuleFrame}{5.\ Diversity and quality filtering}
  {\flowpill{flowRuleBg}{deterministic}}
  {Select up to 12 diverse tasks per participant with per-type caps; move duplicate-heavy segments to an audit bucket.}

\vspace{0.28em}
\flowphase{Memory-grounded release}
\flowstagecolor{flowHybridBg}{flowHybridFrame}{6.\ Object enrichment from existing memory}
  {\flowpill{flowHybridBg}{memory-grounded}}
  {Match EPIC nouns to the video's object registry and attach color, material, location, and memory object IDs without altering EPIC fields.}
\flowarrow
\flowstagecolor{flowOutputBg}{flowOutputFrame}{7.\ Planning task bundle}
  {\flowpill{flowOutputBg}{release}}
  {Release task query, participant/video IDs, reference steps, primary objects/verbs, repeat rate, and matched-object metadata.}
\end{minipage}
\caption{\textbf{MEMORA-Planning benchmark construction pipeline.}
The flow converts EPIC action annotations into prospective, participant-specific planning requests.
Unlike the memory-bank construction flow in Figure~\ref{fig:memory-bank-flow}, this figure describes held-out benchmark construction; the memory bank is only read during video scope checking and object enrichment.}
\label{fig:planning-benchmark-flow}
\end{figure*}

\subsection{Task Construction}

\paragraph{Task definition.}
An instance consists of a natural-language request $x$, a participant ID, a
video ID, and a reference sequence of EPIC actions.  The request has the form of
a user instruction to a household robot, e.g., ``Help P03 clean the washing up
bowl in the sink.''  The reference sequence stores the ordered EPIC narrations,
verbs, nouns, and timestamps for the segment from which the request was derived.
The benchmark therefore tests procedural generation under verifiable evidence:
every released item can be audited against the annotation table, but the model's
output remains open-ended.

\paragraph{Design principles.}
Planning items are \emph{participant-specific}: they should reward agents that
retrieve the person's observed routines, object locations, and preferences, not
agents that merely know generic cooking scripts.  The generation process is
\emph{model-agnostic}: segmentation, classification, query templating,
diversity selection, filtering, and memory grounding are rule-based in the reported
release.  Finally, the benchmark is \emph{memory-scoped}: a task is generated
only from videos present in that participant's consolidated \InfMem{}, so a
memory-enabled agent is not penalized for lacking access to the relevant
episode.
Figure~\ref{fig:planning-benchmark-flow} summarizes the deterministic generation stages.

\paragraph{Segmentation and task typing.}
Within each video, action rows are sorted by timestamp.  A hard split is placed
when adjacent narrations are separated by more than 30 seconds. Soft splits are
deterministic: verbs map to \texttt{cleanup}, \texttt{cooking}, \texttt{prep},
or \texttt{general}, and action $i$ starts a new segment only if the
$i{-}1\!\to\!i$ gap exceeds 10 seconds, the category changes between two
non-\texttt{general} categories, and action $i{+}1$ confirms the new category.
Segments with fewer than three actions are merged with a neighbor, segments
with more than 15 actions are divided into near-even chunks, and the resulting
fixed task set is shared by all memory conditions. The candidates are then
classified into the five task types in Table~\ref{tab:planning-taxonomy}.
Filler pick-and-place verbs are removed before classification so that
superficial object motion does not dominate the task label.

\Needspace{0.34\textheight}
\begin{table*}[t]
\centering
\footnotesize
\setlength{\tabcolsep}{5pt}
\renewcommand{\arraystretch}{1.08}
\caption{MEMORA-Planning task taxonomy. Categories are assigned by deterministic verb-category thresholds.}
\label{tab:planning-taxonomy}
\begin{tabularx}{0.82\textwidth}{@{} >{\raggedright\hyphenpenalty=10000\exhyphenpenalty=10000\arraybackslash}p{0.20\textwidth} >{\raggedright\arraybackslash}p{0.34\textwidth} >{\raggedright\arraybackslash}X @{}}
\toprule
\textbf{Task type} & \textbf{Trigger} & \textbf{Example request} \\
\midrule
\texttt{meal\_}\newline\texttt{preparation}
& preparation verbs $>30\%$
& Help P02 peel and prepare the onion. \\
\texttt{cleanup\_}\newline\texttt{organization}
& cleanup verbs $>35\%$
& Help P03 clean the washing up bowl in the sink. \\
\texttt{object\_retrieval\_}\newline\texttt{setup}
& setup verbs $>30\%$, or storage nouns with no meaningful verb
& Help P06 get items from storage and set up. \\
\texttt{multi\_step\_}\newline\texttt{cooking}
& cooking verbs $>30\%$
& Help P12 stir-fry the vegetables. \\
\texttt{routine\_}\newline\texttt{reproduction}
& no dominant meaningful category
& Help P11 complete the kitchen routine. \\
\bottomrule
\end{tabularx}
\end{table*}

\paragraph{Selection and memory-grounded enrichment.}
Candidate tasks are scored by proximity to the target length, verb diversity,
and the fraction of non-filler actions.  For each participant we greedily select
up to 12 diverse tasks, with a per-type cap of four tasks under the main-paper
configuration.  A subsequent improvement pass computes duplicate-narration
rate, $1-|\mathrm{unique\ narrations}|/|\mathrm{steps}|$, and moves tasks above
0.70 to an audit-only filtered bucket.  For retained tasks, each EPIC noun is
matched against the same video's \texttt{object\_registry} using exact,
substring, token-overlap, and synonym matching.  Successful matches add an
enriched narration and a structured \texttt{matched\_object} record containing
memory object ID, name, color, material, and location.  This enrichment is strictly
additive: the original EPIC fields remain unchanged.

\Needspace{0.34\textheight}
\begin{center}
\captionsetup{type=table}
\centering
\scriptsize
\setlength{\tabcolsep}{4pt}
\renewcommand{\arraystretch}{1.1}
\caption{Core schema for an improved planning task.  The original EPIC fields
remain available after memory enrichment.}
\label{tab:planning-schema}
\begin{tabularx}{\linewidth}{@{}p{0.31\linewidth}p{0.17\linewidth}X@{}}
\toprule
\textbf{Field} & \textbf{Source} & \textbf{Description} \\
\midrule
\texttt{task\_id} & rule & \texttt{plan\_\{video\_id\}\_\{segment\}} \\
\texttt{participant\_id}, \texttt{video\_id} & EPIC & Participant and session identifiers \\
\texttt{task\_type} & rule & One of the five labels in Table~\ref{tab:planning-taxonomy} \\
\texttt{task\_query} & rule & Natural request produced from task type and dominant objects/actions \\
\texttt{ground\_truth\_steps} & EPIC & Ordered action rows with narration, verb, noun, start, and stop times \\
\texttt{primary\_objects}, \texttt{primary\_verbs} & rule & Top segment nouns and verbs \\
\texttt{repeat\_rate} & rule & Duplicate-narration fraction used by the filter \\
\texttt{enriched\_narration} & memory + rule & Optional step text augmented with object attributes \\
\texttt{matched\_object} & memory + rule & Optional memory object ID, name, color, material, and location \\
\bottomrule
\end{tabularx}
\end{center}

\paragraph{Consolidated \InfMem{}.}
The MEMORA conditions consume a per-participant consolidated \InfMem{} bank.  For planning,
the most important fields are \texttt{skill\_memory.routine\_skills}, which
stores routine goals, canonical steps, supporting episodes, and key objects;
and \texttt{skill\_memory.generated\_preferences}, which stores evidence-grounded
preferences, keywords, confidence scores, and supporting episodes.  These fields
make the consolidated \InfMem{} more than a searchable transcript: they expose consolidated
procedures and preferences that a planner can retrieve and adapt.

\paragraph{Participant coverage.}
The reported generation covers 18 EPIC participants with released consolidated \InfMem{}
banks: P01, P02, P03, P04, P06, P07, P09, P11, P12, P22, P23, P25, P26, P27,
P28, P30, P35, and P37.  P05 is excluded because its EPIC identifier convention
does not match the three-digit suffix pipeline used for this release.

\subsection{Memory Conditions and Tools}

\paragraph{Evaluation settings.}
The evaluation compares seven settings while holding the planning model,
benchmark tasks, perception stream, agent budget, and shared runtime stack
(Appendix~\ref{app:impl}) fixed. Each representation is paired with its intended
read tools and retrieval instructions. Cross-family results therefore compare
complete memory interfaces; the raw/processed and episodic/full pairs provide
the more tightly controlled tests of online editing and offline consolidation,
respectively. The linear-memory baseline follows the Socratic-models style of
exposing a text memory through language-model interaction~\citep{zeng2022socraticmodelscomposingzeroshot}.
Table~\ref{tab:planning-conditions} summarizes the conditions.

\Needspace{0.34\textheight}
\begin{center}
\captionsetup{type=table}
\centering
\scriptsize
\setlength{\tabcolsep}{3pt}
\renewcommand{\arraystretch}{1.08}
\caption{Planning evaluation settings. Cross-family rows differ in memory
representation and read interface; paired rows within a family isolate editing
or consolidation.}
\label{tab:planning-conditions}
\begin{tabularx}{\linewidth}{@{}p{0.20\linewidth}p{0.22\linewidth}p{0.25\linewidth}X@{}}
\toprule
\textbf{Condition} & \textbf{Memory} & \textbf{Tools} & \textbf{Prompt behavior} \\
\midrule
Parametric
& none
& none
& Single-shot planning from the request only. \\
Flat-1D (raw)
& chronological observations
& \texttt{search}
& Search-first access to unedited text memory. \\
Flat-1D (processed)
& chronological edited observations
& \texttt{search}
& Same Flat-1D interface after online memory editing. \\
Graph-2D (raw)
& entity-relation graph
& graph-oriented entity, activity, and temporal retrieval
& Procedure retrieval followed by entity grounding. \\
Graph-2D (processed)
& graph rebuilt from edited observations
& same Graph-2D tools
& Same Graph-2D interface after online memory editing. \\
MEMORA-Episodic
& \EnvMem{}, \EntMem{}, and \ActMem{}
& entity, activity, temporal, and unified retrieval
& Episodic evidence followed by entity grounding. \\
MEMORA (full)
& all four typed stores
& episodic retrieval plus routine and preference retrieval
& Routine-first retrieval followed by preference and entity grounding. \\
\bottomrule
\end{tabularx}
\end{center}

\paragraph{Planner and runtime.}
Planning-specific runtime settings beyond the shared stack in Appendix~\ref{app:impl} are:
\begin{itemize}\itemsep1pt\topsep2pt
  \item \textbf{Task-list loading:} the planner is loaded once per 72-specification run so weights stay resident for the full batch.
  \item \textbf{ReAct budget:} at most $8$ iterations, including a forced-answer final iteration with tools disabled.
  \item \textbf{Tool parsing:} a local parser handles Qwen-style XML/tool-call formats inside the agent wrapper.
  \item \textbf{Optional judge:} \texttt{gpt-4o-mini} is available for ablations but is not used in the reported rule-based panel.
\end{itemize}

\paragraph{ReAct loop and plan parser.}
Memory-backed conditions run a ReAct-style loop~\citep{yao2023reactsynergizingreasoningacting} under the shared agent-loop budget in Appendix~\ref{app:impl} (Planning: $8$ iterations including forced answer).
At each iteration the environment supplies conversation state and registered tools; valid tool calls are executed and appended as observations.
On the final iteration, the environment enters forced-answer mode: tools are
cleared, previous tool calls are summarized as text, and the prompt explicitly
requires a final \texttt{Plan:} rather than another tool call.
The parser extracts the generated plan from numbered lines, bullet lists, or an explicit \texttt{Plan:} block.

\paragraph{Structured planning tools.}
The MEMORA prompt is designed to make the consolidated \InfMem{} operational.  It asks the
agent to begin with \InfTok{get\_routine\_skill(goal\_query, top\_k)}, optionally
query \InfTok{get\_preferences}, and then ground concrete objects through \EntMem{}
search.  \InfTok{get\_routine\_skill} retrieves routine memories by embedding
similarity over routine goals, canonical steps, and key objects.  \InfTok{get\_preferences}
retrieves evidence-grounded preferences by query similarity or confidence.

\begin{table*}[t]
\centering
\caption{\textbf{Expanded memory-formation-to-planning examples.}
This is the complete version of main-paper Table~\ref{tab:benchmark-examples}: each row preserves the input evidence, MEMORA record or retrieval sequence, resulting output, planning relevance, and controlled contrast.}
\label{tab:benchmark-examples-expanded}
\footnotesize
\setlength{\tabcolsep}{3pt}
\renewcommand{\arraystretch}{1.06}
\newcommand{\apptabcell}[1]{\parbox[t]{\linewidth}{\raggedright #1}}
\begin{tabularx}{\textwidth}{@{}>{\raggedright\arraybackslash}p{0.33\textwidth}
>{\raggedright\arraybackslash}X
>{\raggedright\arraybackslash}p{0.24\textwidth}@{}}
\toprule
\textbf{Capability tested} & \textbf{Real MEMORA evidence/output} & \textbf{Why it matters} \\
\midrule
\cellcolor{entTint!10}
\apptabcell{\textbf{Maintain an object through time}\\
\textit{Input evidence:} the same dirty plate is observed at 40\,s, 50\,s, and 540\,s.} &
\cellcolor{entTint!10}
\apptabcell{\textbf{Entity Memory record:} \texttt{plate\_with\_food\_residue}\\
\textbf{State history:} \texttt{idle@countertop} (40\,s) $\rightarrow$ \texttt{idle@countertop} (50\,s) $\rightarrow$ \texttt{being\_used@countertop} (540\,s).\\
The editor keeps one object identity while recording state changes.} &
\cellcolor{entTint!10}
\apptabcell{\textbf{Planning relevance:} future actions need the current state of a persistent object, not a pile of repeated sightings.\\
\textbf{Controlled contrast:} append-only memory stores the three sightings separately. Online editing reduces entity records by median $\approx18\times$ (Figure~\ref{fig:memory-maintenance}).} \\
\cellcolor{memoraResultTint}
\apptabcell{\textbf{Turn remembered evidence into action}\\
\textit{Goal:} Help P03 clean the washing up bowl in the sink.} &
\cellcolor{memoraResultTint}
\apptabcell{\textbf{Typed retrieval}
\begin{enumerate}[leftmargin=1.15em,itemsep=0pt,topsep=0pt,partopsep=0pt,parsep=0pt]
\item \texttt{get\_routine\_skill("clean a bowl")} returns an under-specific cookware routine, which MEMORA rejects.
\item \texttt{get\_routine\_skill("wash bowl")} retrieves the dishwashing routine.
\item \texttt{search\_objects(name="bowl")} grounds the black cooking bowl, green sponge, chrome faucet, and metal drying rack.
\end{enumerate}
\textbf{Grounded plan:} clean the black cooking bowl with the green sponge under the chrome faucet, then place it on the metal drying rack.} &
\cellcolor{memoraResultTint}
\apptabcell{\textbf{Planning relevance:} the agent needs to bind a generic goal to experience-grounded objects, tools, and places.\\
\textbf{Controlled contrast:} No Memory fills in generic objects; Flat-1D retrieves plausible but mismatched bowl/towel evidence.} \\
\cellcolor{processedBaselineTint}
\apptabcell{\textbf{Consolidate repeated experience}\\
\textit{EAM-QA:} \textsc{SPref}.\\
When preparing to cook eggs, does P04 prefer a spatula or fork?
Choices: A) random, B) fork, C) spatula, D) whisk, E) unavailable.
Gold: C.} &
\cellcolor{processedBaselineTint}
\apptabcell{\textbf{Answer: C.}
\textbf{Evidence:} P04 repeatedly stirs yellow mixtures/eggs with a wooden spatula (e.g., P04\_116 turns 28, 39).} &
\cellcolor{processedBaselineTint}
\apptabcell{\textbf{Planning relevance:} plans can reflect stable participant preferences when repeated evidence supports them.\\
\textbf{Controlled contrast:} No Memory, Flat-1D, and Graph-2D select E despite repeated action evidence.} \\
\bottomrule
\end{tabularx}
\end{table*}

\paragraph{Planning prompt templates.}
Figure~\ref{fig:planning-prompt-templates} summarizes the operational differences
among the prompt families and the shared forced-answer mode.

\Needspace{0.34\textheight}
\begin{center}
\captionsetup{type=figure}
\centering
\begin{minipage}[t]{\linewidth}
\textbf{No memory}\vspace{0.2em}\par
\promptpanelbox{0.95\linewidth}{%
System: robot planning assistant.\\
Input: task\_query and participant\_id.\\
Constraint: do not assume participant-specific memory.\\
Required final answer: ordered executable plan with numbered steps.}
\end{minipage}
\hfill
\begin{minipage}[t]{\linewidth}
\textbf{Flat-1D memory}\vspace{0.2em}\par
\promptpanelbox{0.95\linewidth}{%
System: robot planning assistant with access to text memory.\\
Tool: \texttt{search(query)}.\\
Instruction: retrieve relevant observations before planning and ground the final plan in retrieved evidence when available.}
\end{minipage}

\vspace{0.6em}

\begin{minipage}[t]{\linewidth}
\textbf{Graph-2D memory}\vspace{0.2em}\par
\promptpanelbox{0.95\linewidth}{%
System: robot planning assistant with access to an entity-relation graph.\\
Tools: graph-oriented entity, activity, and temporal retrieval.\\
Instruction: retrieve procedural evidence, then ground named entities and locations.}
\end{minipage}

\vspace{0.6em}

\begin{minipage}[t]{\linewidth}
\textbf{MEMORA-Episodic}\vspace{0.2em}\par
\promptpanelbox{0.95\linewidth}{%
System: robot planning assistant with typed episodic memory.\\
Tools: entity, activity, temporal, and unified search.\\
Instruction: recover relevant episodes, then ground entities and locations.}
\end{minipage}

\vspace{0.6em}

\begin{minipage}[t]{\linewidth}
\textbf{MEMORA}\vspace{0.2em}\par
\promptpanelbox{0.95\linewidth}{%
System: robot planning assistant for a specific person's kitchen.\\
Priority: call \texttt{get\_routine\_skill} first; call \texttt{get\_preferences} when preferences may affect the plan; use object tools for grounding; use generic search only as fallback.}
\end{minipage}
\hfill
\begin{minipage}[t]{\linewidth}
\textbf{Forced final answer}\vspace{0.2em}\par
\promptpanelbox{0.95\linewidth}{%
Instruction: tool use is disabled. Do not emit tool-call XML or analysis tags. Return only a parseable final plan.\\
Output: \texttt{Plan: 1. ... 2. ...}}
\end{minipage}
\caption{\textbf{Planning prompt families.}
The panel records the operational differences in memory, tools, and retrieval
priority; every condition shares the same final-answer constraint and iteration budget.}
\label{fig:planning-prompt-templates}
\end{center}

\subsection{Metrics}

\paragraph{Rule-based metrics.}
The reported panel uses rule-based metrics rather than an LLM judge.  For a
generated plan $P=(p_1,\ldots,p_m)$, task query $x$, participant routines $R$,
and participant preferences $Q$, we compute four scores.

\Needspace{0.34\textheight}
\begin{center}
\captionsetup{type=table}
\centering
\scriptsize
\setlength{\tabcolsep}{2.5pt}
\renewcommand{\arraystretch}{1.12}
\caption{Rule-based planning metrics. Undefined routine- or preference-based scores are skipped when computing means.}
\label{tab:planning-metric-summary}
\begin{tabularx}{\linewidth}{@{}p{0.16\linewidth}>{\raggedright\arraybackslash}p{0.27\linewidth}>{\raggedright\arraybackslash}p{0.27\linewidth}>{\raggedright\arraybackslash}X@{}}
\toprule
\textbf{Metric} & \textbf{Question} & \textbf{Evidence} & \textbf{Undefined if} \\
\midrule
Exec & Is each step physically actionable? & Generated plan text & Plan is empty or unparseable \\
OrderExec & Are reference steps recovered in order? & Generated plan + reference step sequence (Replay routine or Generalize GT) & Reference unavailable \\
KeyObj & Does the plan mention key routine objects? & Matched routine key-object list & No matched routine or no key objects \\
KBGround & Are plan objects part of the participant's KB vocabulary? & Generated plan + KB noun set & Plan contains no kitchen-noun candidate \\
PrefAdh & Does the plan reflect relevant preferences? & Matched preference entries & No preference passes the relevance threshold \\
RGP & Composite mean of OrderExec, KeyObj, and PrefAdh & Condition-level means of the defined axes above & All component axes are undefined \\
\bottomrule
\end{tabularx}
\end{center}

\paragraph{Reference provenance.}
Replay asks whether an agent recovers a workflow already present in participant
memory, so OrderExec, KeyObj, and PrefAdh use matched consolidated routine,
object, and preference references. Generalize has no matching stored routine by
construction; procedure order is therefore scored against verified task-level
\texttt{ground\_truth\_steps}, with the selected source recorded in
\texttt{metric\_source}.

\textbf{Executability (Exec).}
A step is executable if it contains an action, an object, and a location phrase
according to rule-based predicates.  Let $A(p)$, $O(p)$, and $L(p)$ denote the
action, object, and location predicates:
\begin{equation}
\mathrm{Exec}(P)=
\frac{|\{p_i \in P: A(p_i)\wedge O(p_i)\wedge L(p_i)\}|}{|P|}.
\end{equation}

\textbf{Procedure order (OrderExec).}
The metric compares the normalized action-verb sequence of the generated plan with a per-task reference verb sequence.
On Replay, the reference is the matched routine's canonical-step verbs (selected as the participant's \texttt{routine\_skills} entry with highest Jaccard overlap with the task query, threshold $0.10$).
On Generalize, where no matching routine exists by construction, the reference is the task's verified \texttt{ground\_truth\_steps} verb sequence; the metric router (\texttt{scripts/compute\_rule\_metrics.py}) auto-detects Generalize tasks by their \texttt{denovo\_} prefix and records the chosen reference in a per-task \texttt{metric\_source} field.
\begin{equation}
\mathrm{OrderExec}(P,s)=
\frac{\mathrm{LCS}(v(P),v(s))}{|v(s)|},
\end{equation}
where $s$ is the per-task reference sequence and $v(\cdot)$ denotes a synonym- and inflection-normalized verb sequence.
Tasks with no qualifying reference are undefined and skipped in the metric mean.

\textbf{Key-object coverage (KeyObj).}
For the matched routine $r$ with key-object set $K_r$, we compute:
\begin{equation}
\mathrm{KeyObj}(P,r)=
\frac{|\{k\in K_r: k \in P\}|}{|K_r|}.
\end{equation}

\textbf{Preference adherence (PrefAdh).}
Preferences are first filtered by Jaccard overlap between the task query and the
preference text or keywords, with threshold 0.05.  For each relevant preference,
we measure the fraction of preference tokens appearing in the plan, then
average over relevant preferences.

\textbf{Robot-Grounded Plan score (RGP).}
The headline planning aggregate is the unweighted mean of three condition-level robot-followability scores. For condition $c$, each component is first averaged over the tasks on which that component is defined:
\begin{equation}
\begin{aligned}
\mathrm{RGP}(c)=\tfrac{1}{3}\big(&\overline{\mathrm{OrderExec}}_c
 + \overline{\mathrm{KeyObj}}_c\\[-2pt]
&+ \overline{\mathrm{PrefAdh}}_c\big).
\end{aligned}
\end{equation}
Each axis captures one execution requirement (order, task-relevant object coverage, personalization).
Raw Exec is reported separately because it rewards plan length regardless of correctness.
KBGround is reported as a separate diagnostic in \S\ref{app:planning-grounding-diag} rather than folded into RGP, for the metric reasons discussed there.

\textbf{Verb normalization.}
OrderExec uses the V2 verb extractor for both the plan and reference sequences.
This normalization corrects two systematic under-estimation errors: canonical memory steps often begin with ``Person ...'' rather than ``The person ...'', and routine or GT-step descriptions often use inflected verbs such as \textit{places}, \textit{washing}, or \textit{preparing}.
V2 strips article-free and pronominal subjects, maps common inflections to canonical verb classes, and suppresses non-verb fallback tokens such as \textit{person}.
The LCS scoring formula itself is unchanged.

\subsection{Results and Diagnostics}

\begin{figure*}[t]
\centering
\includegraphics[width=0.93\linewidth]{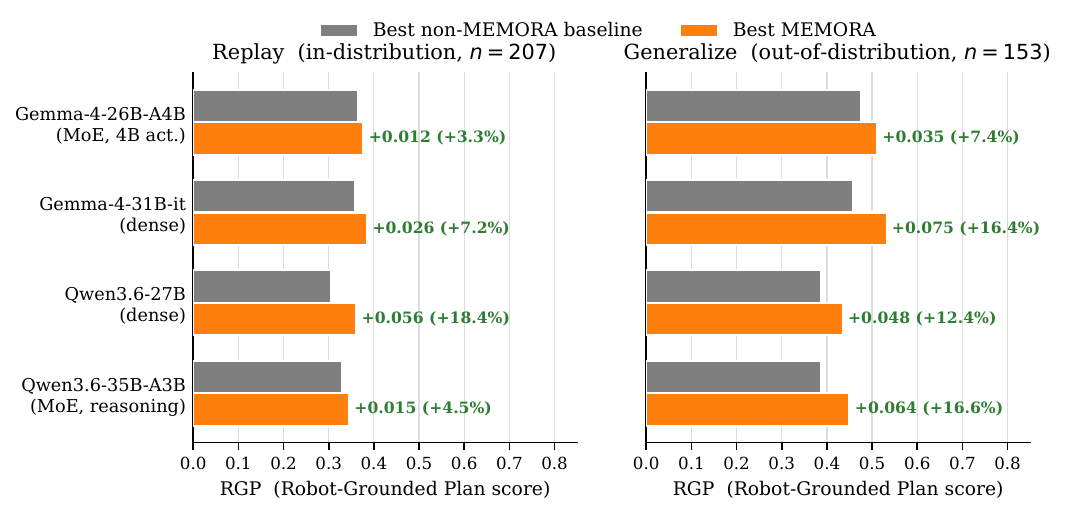}
\caption{%
\textbf{Cross-backbone RGP summary} (companion to Figure~\ref{fig:planning-main}).
Grey = best non-MEMORA baseline; orange = best MEMORA variant; annotations give $\Delta$~RGP and relative gain.
}
\label{fig:planning-xbackbone-appendix}
\end{figure*}

\paragraph{Cross-backbone summary.}
Figure~\ref{fig:planning-xbackbone-appendix} summarizes the eight (backbone, benchmark) cells from the main paper.
Grey = best non-MEMORA baseline; orange = best MEMORA variant.
MEMORA wins every cell, with $\Delta$~RGP rising from $+0.012\!-\!+0.056$ on Replay to $+0.035\!-\!+0.075$ on Generalize.
Two recurring patterns underlie the two observations summarised in \S\ref{sec:planning-results}.
First, MEMORA's gain is larger on Generalize than on Replay for three of the four backbones, suggesting that its benefits are often greatest when the model must compose beyond a matching observed episode.
Second, the two larger reasoning backbones (Gemma-4-31B-it and Qwen3.6-35B-A3B) extract more value from the \emph{same} memory state on Generalize, both delivering $\geq +0.064$ absolute $\Delta$~RGP, consistent with a reasoning-bound rather than memory-bound ceiling on the harder split.

\Needspace{0.32\textheight}

\paragraph{Per-condition numerical results: Gemma-4-31B-it (main-paper RGP figure).}
Figure~\ref{fig:planning-main} reports per-condition RGP on Gemma-4-31B-it for Replay ($n{=}207$) and Generalize ($n{=}153$); the per-axis OrderExec / KeyObj / PrefAdh decomposition referenced in \S\ref{sec:planning-results} is available in the figure companion panel.

\paragraph{Per-condition numerical results: Qwen3.6-35B-A3B.}
Table~\ref{tab:planning-perax-q35-full} reports the same per-condition decomposition on Qwen3.6-35B-A3B.
The qualitative pattern matches Gemma-4-31B-it: typed retrieval lifts the same three axes, with the magnitude consistently larger on Generalize, where chunk-retrieval baselines no longer benefit from verbatim segment recall.

\Needspace{0.40\textheight}
\begin{table*}[t]
\centering
\small
\setlength{\tabcolsep}{5pt}
\renewcommand{\arraystretch}{1.06}
\caption{Per-condition per-axis Planning detail on Qwen3.6-35B-A3B.
\emph{(proc.)} indicates a baseline substrate rebuilt after MEMORA's online editing pass.
\textbf{Bold}: best per column within each benchmark block.}
\label{tab:planning-perax-q35-full}
\begin{tabular}{@{}l rrrr c rrrr@{}}
\toprule
                                        & \multicolumn{4}{c}{\textbf{Replay} ($n{=}207$)} & & \multicolumn{4}{c}{\textbf{Generalize} ($n{=}153$)} \\
\cmidrule(lr){2-5}\cmidrule(lr){7-10}
\textbf{Condition}                      & OrderExec & KeyObj & PrefAdh & RGP        & & OrderExec & KeyObj & PrefAdh & RGP \\
\midrule
Parametric (No Memory)                  & .072 & .316 & .233 & .207                  & & .175 & .485 & .301 & .320 \\
Flat-1D (raw)                           & .087 & .473 & .382 & .314                  & & .166 & .570 & .422 & .386 \\
Graph-2D (raw)                          & .065 & .411 & .386 & .287                  & & .145 & .491 & .371 & .336 \\
Flat-1D (proc.)                         & \textbf{.096} & .494 & .400 & .330         & & .170 & .531 & .405 & .369 \\
Graph-2D (proc.)                        & .079 & .426 & .397 & .300                  & & .151 & .492 & .425 & .356 \\
MEMORA-Episodic                         & .085 & \textbf{.501} & .448 & \textbf{.345} & & .186 & \textbf{.651} & .440 & .425 \\
\textbf{MEMORA} (full)                  & .091 & .474 & \textbf{.450} & .338         & & \textbf{.243} & .650 & \textbf{.458} & \textbf{.450} \\
\bottomrule
\end{tabular}
\end{table*}

\begin{figure*}[t]
\centering
\includegraphics[width=0.94\linewidth]{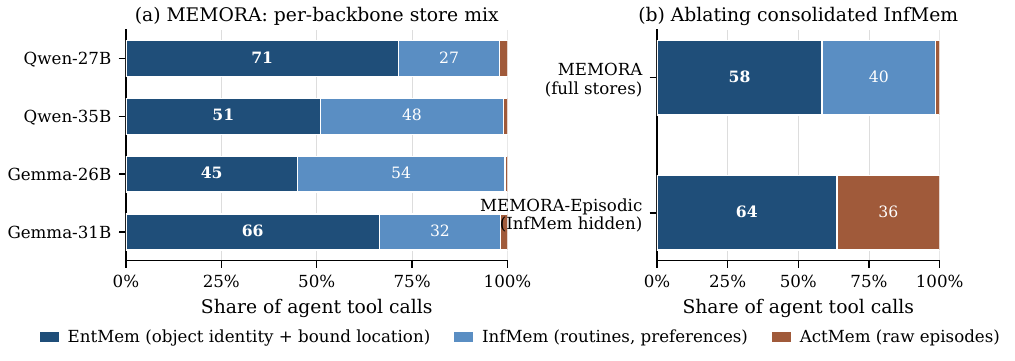}
\caption{\textbf{Per-store contribution of agent tool calls during planning} ($1{,}440$ runs $=$ $4$ backbones $\times$ $(207$ Replay $+153$ Generalize tasks$)$).
\textbf{(a)}~Under the main MEMORA configuration, every backbone draws on multiple typed stores, but with markedly different mixtures: Gemma4-26B-A4B-it routes $54\%$ of its tool budget to \InfMem{} (routines, preferences), while Qwen3.6-27B routes $71\%$ to \EntMem{} (object identity).
A single-store substrate may fit these backbone families unevenly.
\textbf{(b)}~In the MEMORA-Episodic ablation, agents query \EntMem{} and \ActMem{} only (\texttt{search\_objects}, \texttt{get\_object\_history}, \texttt{search\_activities}); those queries are \emph{not} eliminated --- $35\%$ of all calls migrate to \ActMem{} episode search, showing that removing consolidated routines redirects substantial tool use to episodic activity search.
\EnvMem{} is omitted from the figure because spatial information is bound to object records in the planning prompt (every \texttt{search\_objects} return includes \texttt{location} attributes), so a separate environment-search tool is not advertised.}
\label{fig:store-contribution}
\end{figure*}

\paragraph{Per-store tool-use distribution.}
Figure~\ref{fig:store-contribution} aggregates every logged \texttt{tool\_call} across all $1{,}440$ planning runs ($4$ backbones $\times$ $(207$ Replay $+153$ Generalize tasks$)$, $7{,}163$ calls under the main MEMORA configuration) and maps each tool to the typed store it queries (\EntMem{}: \texttt{search\_objects}, \texttt{get\_object\_history}; \InfMem{}: \texttt{get\_routine\_skill}, \texttt{get\_preferences}; \ActMem{}: \texttt{search\_activities}).
Two patterns recur.
First, backbones differ sharply in how they spend their tool budget: \InfMem{} share ranges from $27\%$ (Qwen3.6-27B) to $54\%$ (Gemma4-26B-A4B-it), suggesting that a single-store interface may fit backbone families unevenly.
Second, comparing MEMORA against the MEMORA-Episodic ablation (identical agent loop and KB, but agents rely on \EntMem{} and \ActMem{} tools only) shows that \InfMem{}'s $40\%$ share does \emph{not} simply vanish: $35\%$ of all tool calls instead migrate to \texttt{search\_activities} over \ActMem{}, consistent with consolidated routines absorbing queries that otherwise shift to episodic activity search.
Across both configurations the unified \texttt{search} dispatcher remains rare ($<\!4\%$); agents prefer the typed entry points the prompt advertises, and $\approx 98.5\%$ of MEMORA's tool calls go through the typed cascade (\texttt{search\_objects} / \texttt{get\_routine\_skill} / \texttt{get\_preferences} / \texttt{search\_activities}), with the cascade invoked on $96$--$100\%$ of Generalize tasks per backbone.
For contrast, Flat-1D conditions on the same Generalize split issue an average of $5.8$ \texttt{search} queries per task, overwhelmingly keyword lookups that can surface a verbatim chunk when the task matches an observed video but offer no structured affordance for composition.

\begin{table*}[t]
\centering
\footnotesize
\setlength{\tabcolsep}{4pt}
\renewcommand{\arraystretch}{1.05}
\caption{Planning tool-call counts by typed store and configuration.
\textit{MEMORA-Episodic} exposes \EntMem{} and \ActMem{} tools only; full \textit{MEMORA} additionally provides \texttt{get\_routine\_skill} and \texttt{get\_preferences} over consolidated \InfMem{}.
Percentages are normalized within each variant ($7{,}163$ vs.\ $8{,}331$ total calls); tools with $<\!1\%$ usage in both variants are omitted.}
\label{tab:per-tool-distribution}
\begin{tabularx}{\linewidth}{@{} >{\raggedright\arraybackslash}p{0.07\linewidth} >{\raggedright\arraybackslash}X >{\raggedleft\arraybackslash}p{0.08\linewidth} >{\raggedleft\arraybackslash}p{0.05\linewidth} >{\raggedleft\arraybackslash}p{0.08\linewidth} >{\raggedleft\arraybackslash}p{0.05\linewidth} @{}}
\toprule
\textbf{Store} & \textbf{Tool} &
\multicolumn{2}{c}{\textbf{MEMORA}} &
\multicolumn{2}{c}{\textbf{MEMORA-Episodic}} \\
\cmidrule(lr){3-4}\cmidrule(lr){5-6}
 & & \textbf{calls} & \textbf{\%} & \textbf{calls} & \textbf{\%} \\
\midrule
\multirow{1}{*}{\textsc{EntMem}}
  & \texttt{search\_objects}     & 4180 & 58.4 & 5103 & 61.3 \\
\midrule
\multirow{2}{*}{\textsc{InfMem}}
  & \texttt{get\_routine\_skill} & 2530 & 35.3 &    0 &  0.0 \\
  & \texttt{get\_preferences}    &  347 &  4.8 &    0 &  0.0 \\
\midrule
\multirow{1}{*}{\textsc{ActMem}}
  & \texttt{search\_activities}  &  103 &  1.4 & 2912 & 35.0 \\
\midrule
\multirow{1}{*}{\textsc{Unified}}
  & \texttt{search}              &    3 &  0.0 &  297 &  3.6 \\
\bottomrule
\end{tabularx}
\end{table*}

\paragraph{Illustrative trace.}
Figure~\ref{fig:planning-tool-trace} provides a compact excerpt from a real MEMORA
trace in the reported evaluation.  The task is \texttt{plan\_P03\_101\_019}:
``Help P03 clean the washing up bowl in the sink''
(\texttt{cleanup\_organization}, 4/8 iterations).  The agent first retrieves a
generic cookware-washing routine, notices that the retrieved key objects do not
match the requested bowl, refines the routine query, grounds the bowl through
object search, and then emits a grounded plan.

\Needspace{0.34\textheight}
\begin{center}
\captionsetup{type=figure}
\centering
\promptpanelbox{0.96\linewidth}{%
Task: Help P03 clean the washing up bowl in the sink.\\[0.2em]
Iter. 1: \texttt{get\_routine\_skill(goal\_query="clean a bowl in sink", top\_k=3)} returns a cookware-washing routine about pot/spoon/stove, so the agent rejects it as under-specific.\\
Iter. 2: \texttt{get\_routine\_skill(goal\_query="wash a bowl", top\_k=2)} retrieves a dishwashing routine mentioning a black cooking bowl and drying rack.\\
Iter. 3: \texttt{search\_objects(name="bowl")} returns bowl attributes and nearby grounding objects including green sponge, chrome faucet, and metal drying rack.\\[0.2em]
Final plan excerpt: pick up the black cooking bowl from the sink; pick up the green sponge; turn on the chrome faucet; clean the bowl; place it on the metal drying rack next to the sink.}
\caption{\textbf{Excerpt from a real \texttt{memora} ReAct trace for \texttt{plan\_P03\_101\_019}.}
The trace illustrates routine-first retrieval followed by object grounding before final plan emission.}
\label{fig:planning-tool-trace}
\end{center}

\subsection{Generation pipeline reproducibility}

\paragraph{Reproducibility.}
The generation pipeline is deterministic under the fixed EPIC CSV, participant
list, and released consolidated \InfMem{} banks.  Task extraction targets
12 tasks per participant, followed by a rule-based improvement pass that
computes duplicate-narration rate and moves high-repeat segments to an
audit-only bucket.  The evaluation task list contains one specification for each participant and
condition, for $18 \times 4 = 72$ specifications.  An optional
\texttt{gpt-4o-mini} judge can be run as a separate phase, but it is not used in
the reported rule-based panel.

\paragraph{Generalize benchmark.}
\label{app:planning-denovo}
The Generalize split is generated by a separate, OOD-by-design pipeline that combines three rule-based families with an LLM verification pass that filters out tasks the KB could have answered as Replay.
The first family, \emph{transfer}, draws on a participant's consolidated \InfMem{} to identify ``observed action $a$ on object $o_{\mathrm{obs}}$'' pairs (e.g.\ P01 has been observed cutting mushrooms three times), then substitutes a target object $o_{\mathrm{tgt}}$ that the same participant has been observed but never in combination with $a$ (e.g.\ \texttt{cutting potato}); the task query asks the agent to apply the skill to the substituted target.
The second family, \emph{composition}, samples a high-level goal that is \emph{not} present in \texttt{routine\_skills} but whose sub-procedures individually are (e.g.\ \texttt{prepare a vegetable stir-fry dish} decomposes into operations such as heating a pan, picking up vegetables, and stirring, each of which appears as a routine but never as a parent skill).
The third family, \emph{fully-novel}, asks for plans whose surface goal has neither a matching routine nor a matching sub-procedure decomposition in the KB.
An LLM verifier (\texttt{gpt-4o-mini}) is used \emph{at generation only}: it writes a $5$-step ground-truth plan for each candidate task and scores it on five binary criteria (novelty relative to KB routines, sub-procedure grounding, object grounding, physical correctness, and step-quality coherence); tasks that fail any criterion are dropped.
The released split contains $153$ verified tasks across the $18$-PID panel.
At evaluation time, the rule-based panel substitutes the task-level \emph{PlanFid\_GT} (verb-LCS against \texttt{ground\_truth\_steps}) for the KB-routine reference, because a Jaccard-matched nearest KB routine for a novel goal is only lexically close and penalizes agents that compose correctly; the metric router (\texttt{scripts/compute\_rule\_metrics.py}) auto-detects Generalize tasks by their \texttt{denovo\_} \texttt{task\_id} prefix and records the chosen reference in a per-task \texttt{metric\_source} field.

\subsection{Object-grounding diagnostic}
\label{app:planning-grounding-diag}

We separately report object grounding as a diagnostic because no token-overlap definition of object grounding cleanly captures what memory contributes to robot-followable planning.
This subsection (i) defines four candidate object-grounding metrics, (ii) reports them across all six completed (backbone, benchmark) cells, and (iii) discusses why we ultimately exclude object grounding from RGP while still tracking it.

\paragraph{Four candidate definitions.}
Let $P$ be the set of distinct kitchen-noun stems extracted from a generated plan, $G$ the corresponding set for the ground-truth plan, and $K$ the participant's KB-vocabulary noun set (\S\ref{app:planning-full}).
We consider:
{\footnotesize
\begin{align}
\mathrm{KBGround}(P,K) &= |P \cap K|/|P|, \\
\mathrm{ObjGT\text{-}R}(P,G) &= |P \cap G|/|G|, \\
\mathrm{ObjGT\text{-}P}(P,G) &= |P \cap G|/|P|, \\
\mathrm{ObjGT\text{-}F1}
  &= \frac{2\,\mathrm{ObjGT\text{-}P}\,\mathrm{ObjGT\text{-}R}}{\mathrm{ObjGT\text{-}P}+\mathrm{ObjGT\text{-}R}}, \\
\mathrm{GR}(P,K,G) &= \frac{|P \cap K \cap G|}{|K \cap G|}, \\
\mathrm{GP}(P,K,G) &= \frac{|P \cap K \cap G|}{|P|}, \\
\mathrm{GF1} &= \frac{2\,\mathrm{GP}\,\mathrm{GR}}{\mathrm{GP}+\mathrm{GR}}.
\end{align}
}
KBGround answers ``does the plan mention things known to the participant?''
ObjGT-\{R,P,F1\} answers ``does the plan mention the same objects as the GT plan?''
GR / GP / GF1 (Grounded Recall / Precision / F1) answer the more robot-relevant question ``does the plan mention the right task objects \emph{and} are those objects findable in the participant's kitchen?''

\paragraph{Results across all completed cells.}
Table~\ref{tab:planning-grounding-allcells} reports the four grounding metrics for every (backbone, benchmark) cell.
The pattern is consistent across cells: chunk-based retrieval baselines win KBGround by a wide margin (their plans inherit verbatim text from the chunks used to build the KB), and the winner under ObjGT-F1 or GF1 is typically Flat-1D or Graph-2D rather than MEMORA.

\Needspace{0.34\textheight}
\begin{table*}[t]
\centering
\small
\setlength{\tabcolsep}{5pt}
\renewcommand{\arraystretch}{1.08}
\caption{Object-grounding diagnostic across all completed (backbone, benchmark) cells.
For each cell we report the value for the best non-MEMORA baseline (``best base'') and for the best MEMORA variant (over MEMORA, MEMORA-Episodic), and the gap $\Delta$ (MEMORA $-$ best base, positive favors MEMORA).
ObjGT-F1 = GT-plan noun F1; GF1 = Grounded F1 (KB $\cap$ GT).
MEMORA does not consistently win on any token-overlap grounding metric; the chunk-based retrieval advantage is robust across metric variants.}
\label{tab:planning-grounding-allcells}
\begin{tabular}{@{}l l r r r r r r@{}}
\toprule
\textbf{Backbone} & \textbf{Bench} &
\multicolumn{2}{c}{\textbf{KBGround}} &
\multicolumn{2}{c}{\textbf{ObjGT-F1}} &
\multicolumn{2}{c}{\textbf{GF1 (KB$\cap$GT)}} \\
\cmidrule(lr){3-4}\cmidrule(lr){5-6}\cmidrule(lr){7-8}
 & & best base / MEM & $\Delta$ & best base / MEM & $\Delta$ & best base / MEM & $\Delta$ \\
\midrule
\multirow{2}{*}{\shortstack[l]{Gemma-4-26B-A4B-it\\ \emph{(MoE, 4B active)}}}
  & Replay     & .912 / .883 & $-.029$ & .481 / .482 & $+.001$ & .477 / .461 & $-.016$ \\
 & Generalize & .916 / .885 & $-.031$ & .663 / .610 & $-.053$ & .629 / .579 & $-.050$ \\
\addlinespace[0.15em]
\multirow{2}{*}{\shortstack[l]{Gemma-4-31B-it\\ \emph{(dense)}}}
  & Replay     & .904 / .901 & $-.003$ & .490 / .491 & $+.001$ & .490 / .491 & $+.001$ \\
 & Generalize & .918 / .905 & $-.013$ & .679 / .645 & $-.034$ & .638 / .616 & $-.022$ \\
\addlinespace[0.15em]
\multirow{2}{*}{\shortstack[l]{Qwen3.6-35B-A3B\\ \emph{(MoE, reasoning)}}}
  & Replay     & .909 / .883 & $-.026$ & .491 / .483 & $-.008$ & .494 / .452 & $-.042$ \\
 & Generalize & .925 / .892 & $-.033$ & .622 / .598 & $-.024$ & .608 / .566 & $-.042$ \\
\bottomrule
\end{tabular}
\end{table*}

\paragraph{The chunk-copying signature: KeyObj and KBGround disagree on baselines.}
The contrast between KeyObj (in RGP) and KBGround (this diagnostic) is informative.
On Qwen3.6-35B-A3B / Generalize, Flat-1D~(raw) scores KBGround $=.920$ but KeyObj $=.570$; MEMORA scores KBGround $=.883$ but KeyObj $=.650$ --- the chunk-copying signature.
The Pearson correlation between KeyObj and KBGround at the per-task level is essentially zero on this cell (range $r{=}{-}.116$ to ${+}.220$ across conditions), confirming that the two metrics measure complementary properties: KeyObj rewards plans that name the per-task gold objects, while KBGround rewards plans that mention any KB-known noun.
Folding both into a single composite would (i) double-weight ``object naming'' relative to procedure / preferences, and (ii) reward chunk-copying baselines on the KBGround side without rewarding task-relevant retrieval.
RGP therefore keeps only KeyObj.

\paragraph{Why no token-overlap metric cleanly favors MEMORA.}
Across all four grounding-metric variants, MEMORA wins outright on at most one of six cells (Gemma-4-31B-it / Replay, ObjGT-F1 and GF1 tied at $+.001$).
Three structural reasons explain this:
(1) The Generalize task query directly names the key objects (``Help P01 \emph{cut the cucumber}''), so any LLM with kitchen prior --- including the No-Memory parametric baseline --- produces a plan whose nouns largely match the GT plan; this is a verbosity confound that ObjGT-R rewards (No Memory wins ObjGT-R on every Generalize cell).
(2) Flat-1D copies EPIC narrations verbatim, so its plans inherit both KB-membership and GT-noun-membership ``for free'' on Replay; Graph-2D injects graph-resolved entities sparsely and wins ObjGT-Precision by being terse.
(3) MEMORA paraphrases retrieved objects into natural-language plan steps (e.g., ``the green cucumber that P01 keeps on the counter'') and adds procedural elaboration, both of which dilute token-overlap scores even when the underlying entity is correct.
None of these metrics tests whether the \emph{specific instance} named is correct (which cucumber, on which counter); answering that question requires execution-grounded evaluation in simulation or on a physical platform, which we leave to future work.

%% file: sections_new/Appendix_sections/robot.tex
\section{Physical Robot Demonstration}
\label{app:robot}

This appendix documents the qualitative robot case study summarized in Figure~\ref{fig:robot-demo-main} and referenced in Section~\ref{sec:robot-demo}; Figures~\ref{fig:robot-demo-task-a} and~\ref{fig:robot-demo-task-b} provide the complete traces.
The purpose is not to introduce a separate robot benchmark, but to check whether the object-grounding errors measured by MEMORA-Planning remain visible when a memory-conditioned instruction is executed in a physical scene.
The demonstration is a transfer setting: the memory is formed from a human demonstrator's egocentric video, then retrieved as procedural and object-grounding guidance for a robot acting in the same environment.
We therefore report the setup, task design, success criterion, and video traces for two memory-grounded robot tasks.

\subsection{Setup and Protocol}

\paragraph{Platform and scene.}
The demonstration uses a Unitree~G1 operating in a kitchen-like tabletop scene with visually similar distractor objects.
Before robot execution, the human demonstrator records two egocentric videos in the same scene: one for a drink-preparation routine and one for a breakfast routine.
These videos are processed before robot execution, so the robot plan is conditioned on memory formed from human first-person experience rather than on direct access to the demonstration video at execution time.

\paragraph{Task design.}
Each instruction is intentionally underspecified unless the system remembers the participant's prior choices.
\textbf{Task A: Prepare Drink} tests preference and grounding: the robot must choose bottled Coca-Cola over a yellow Fanta pineapple can, and must choose the demonstrator's orange plastic cup among cups with other colors.
\textbf{Task B: Breakfast} tests preference, grounding, and sequence: the robot must choose the yellow bowl rather than a gray bowl, place Cinnamon Tea rather than Lipton Black Tea into the bowl, place the yellow spoon rather than a metal spoon into the bowl, and then take the bread.
The intended test is whether a memory system can transfer participant-specific human choices into a robot plan that resolves object identity, attributes, preferences, and temporal order, rather than following a generic task script.

\paragraph{Evidence.}
The demonstration consists of four videos: the human egocentric demonstration for Task~A, the human egocentric demonstration for Task~B, the Unitree~G1 execution of Task~A from MEMORA's plan, and the Unitree~G1 execution of Task~B from MEMORA's plan.
The two robot videos provide the qualitative success traces reported below.

\paragraph{Execution interface.}
MEMORA produces a language-level plan for each participant-specific instruction.
The generated plan is then converted into grasp and placement primitives by a fixed rule-based low-level controller.
This design intentionally holds motor execution fixed, so the demonstration focuses on memory-grounded language planning rather than learned robot-control differences.

\begin{table*}[t]
\centering
\footnotesize
\setlength{\tabcolsep}{4pt}
\renewcommand{\arraystretch}{1.15}
\begin{tabularx}{0.98\textwidth}{@{}p{0.14\textwidth}p{0.24\textwidth}p{0.36\textwidth}X@{}}
\toprule
\textbf{Task} & \textbf{Distractors / ambiguity} & \textbf{Generated MEMORA plan used for execution} & \textbf{Success criterion} \\
\midrule
\textsc{Prepare Drink} &
Fanta and other cups are visible, so the instruction alone does not specify the participant's drink setup. &
(1) retrieve the orange cup; (2) retrieve the Coca-Cola bottle; (3) position the orange cup with the Coca-Cola bottle as demonstrated. &
The robot selects Coca-Cola rather than Fanta and the orange cup rather than other cups. \\
\addlinespace[0.25em]
\textsc{Breakfast} &
Multiple bowls, tea items, utensils, and side items are present, so both choices and order depend on prior experience. &
(1) select the yellow bowl; (2) add Cinnamon Tea; (3) add the yellow spoon; (4) bring bread. &
The robot selects the remembered breakfast entities and preserves the demonstrated assembly order. \\
\bottomrule
\end{tabularx}
\caption{\textbf{Robot-transfer task specification and generated plans.}
The plans are language-level MEMORA outputs used to drive a fixed rule-based controller.}
\label{tab:robot-appendix-plans}
\end{table*}

\subsection{Results and Traces}

\begin{figure*}[!t]
\centering
\includegraphics[width=\textwidth]{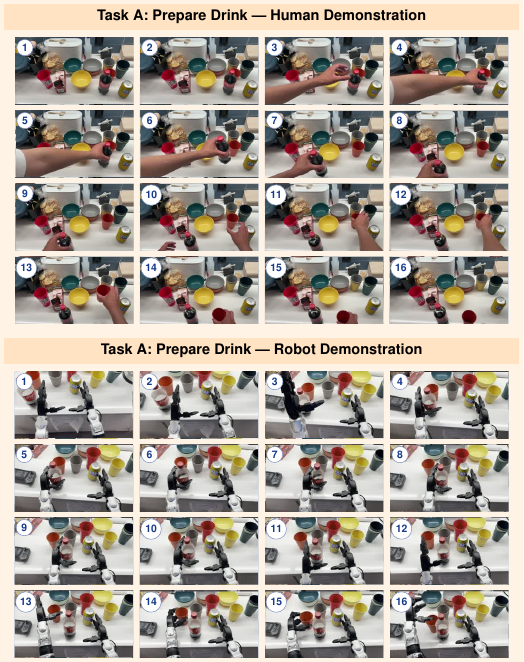}
\caption{\textbf{Task~A: Prepare Drink (full trace).}
The contact sheet shows the human egocentric demonstration and the Unitree~G1 execution guided by MEMORA's memory-grounded language plan and a fixed rule-based controller.
This task tests preference and object grounding: the robot must select bottled Coca-Cola over a visually salient Fanta can and the demonstrator's orange plastic cup among multiple colored cups.}
\label{fig:robot-demo-task-a}
\end{figure*}

\begin{figure*}[!t]
\centering
\includegraphics[width=\textwidth]{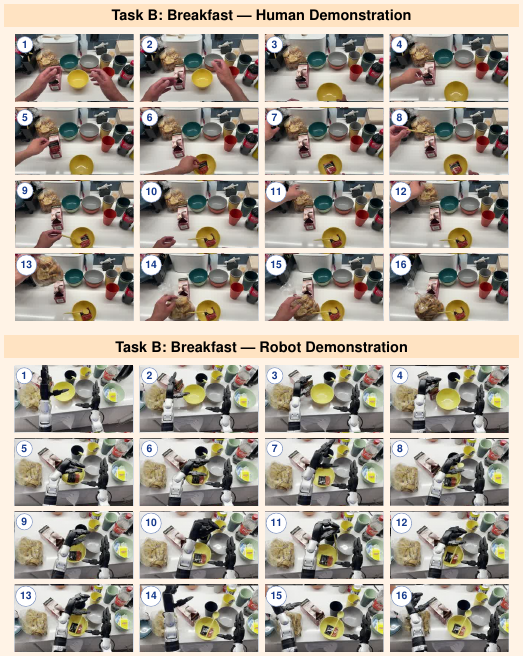}
\caption{\textbf{Task~B: Breakfast (full trace).}
The contact sheet shows the human egocentric demonstration and the Unitree~G1 execution guided by MEMORA's memory-grounded language plan and a fixed rule-based controller.
This task adds ordered procedural execution: the robot must select the yellow bowl, Cinnamon Tea, yellow spoon, and bread in the remembered breakfast sequence.}
\label{fig:robot-demo-task-b}
\end{figure*}

Table~\ref{tab:robot-appendix-plans} records the task-level plan and success criterion, and Figures~\ref{fig:robot-demo-task-a} and~\ref{fig:robot-demo-task-b} show the full visual traces used to construct the compact main-paper figure.
Each panel includes the human egocentric demonstration and the corresponding Unitree~G1 execution, making explicit which observed preferences and object choices are transferred into robot behavior.

\clearpage
\clearpage

\clearpage

Both tasks succeed on the first attempt (2/2): Task~A selects bottled Coca-Cola and the orange plastic cup; Task~B selects the yellow bowl, Cinnamon Tea, yellow spoon, and bread in the remembered order.

\paragraph{Limitations.}
This demonstration should be interpreted as a deployment-oriented sanity check.
It involves two tasks in one physical scene and is not intended to establish statistical robot generalization.
Its role is to make the planning metric concrete by showing that memory formed from human egocentric observation can guide physical object selection and sequencing in a robot execution setting.

\paragraph{Beyond the fixed-controller demonstration.}
The rule-based Unitree~G1 controller used above is one simple instantiation of a broader real-robot stack; it is not a requirement imposed by MEMORA.
MEMORA supplies a high-level goal or plan grounded in participant-specific memory, which can then be executed by a conventional feedback controller, a learned state- or vision-conditioned policy such as Diffusion Policy~\citep{chi2023diffusionpolicy}, or a general vision--language--action policy such as RT-2 or $\pi_0$~\citep{brohan2023rt2visionlanguageactionmodelstransfer,black2026pi0visionlanguageactionflowmodel}.
When adaptation to a target robot or task is needed, parameter-efficient updates such as LoRA can specialize a pretrained policy~\citep{hu2022lora}, while online interaction offers an additional route for refinement: Policy Decorator adds a model-agnostic residual policy to a fixed imitation policy through controlled online exploration~\citep{yuan2025policydecorator}, whereas RL Token (RLT) exposes a compact representation from a pretrained vision--language--action model to a lightweight actor--critic for sample-efficient real-robot online reinforcement learning~\citep{xu2026rltokenbootstrappingonline}.
For long-horizon execution, a progress monitor can assess whether each intended state transition has occurred before the system advances or replans, using vision--language-model state grounding and replanning~\citep{skreta2024replanroboticreplanningperception}, milestone-based progress monitoring and recovery~\citep{dai2026seeplanrewindprogressaware}, or action-conditioned world-model verification~\citep{liu2026checkvlaexecutiontimeverificationactionconditioned}; human confirmation provides a conservative fallback when automated verification is unreliable.
Under this decomposition, MEMORA provides persistent participant-specific context and high-level grounding, the downstream policy realizes the selected action, and the monitor closes the loop by deciding whether to continue, replan, or request intervention.
These are practical pathways for broader evaluation rather than claims established by our two-task demonstration.

%% file: sections_new/Appendix_sections/ablation.tex
\section{Controlled Ablation Studies}
\label{app:ablation}

Appendix~\ref{app:ablation} defines three orthogonal controlled ablations over the MEMORA memory-bank construction pipeline and its downstream answer agents.
\textbf{Reference} uses the main-paper processor settings in Appendix~\ref{app:impl} (processor contracts: Table~\ref{tab:kb-processors}); the released pipeline contains no closed-weight model in any role.
\textbf{Cohort split.}
We partition the 18-participant MEMORA-Bench panel into a \emph{calibration} cohort and a \emph{held-out evaluation} cohort, and run Axes~1--2 on the held-out cohort to avoid leakage of any processor tuning or prompt iteration into the ablation result:
\begin{itemize}\itemsep1pt\topsep1pt
  \item \textbf{Calibration cohort (P01--P04, 4 PIDs, $\sim$80 EPIC-KITCHENS extension videos, $746$ EAM-QA questions = $27.0\%$ of the full EAM-QA panel).}
        This subset was used during pipeline development to iterate perception- and memory-processor prompt templates and to fix the MEMORA pipeline; it is therefore not a clean test set for processor swaps.
  \item \textbf{Held-out ablation cohort (P06, P07, P09, P11, P12, P22, P23, P25, P26, P27, P28, P30, P35, P37 -- 14 PIDs, $\sim$121 videos, $2{,}017$ EAM-QA questions = $73.0\%$ of the benchmark).}
        Memory-processor outputs for the three Axis-1/Axis-2 arms (Omni+Q30B, Omni+Q14B, VL+Q30B) were materialized in a single ``rest'' merge over exactly this PID set, so the comparison covers a fixed video population for every arm.
\end{itemize}
\textbf{Isolation protocol:} each study varies exactly one of \{\emph{online memory processor}, \emph{perception processor (and audio)}, \emph{answer agent}\}; the other two axes remain at the main-paper settings in Appendix~\ref{app:impl} and the \textbf{Reference} row of Table~\ref{tab:ablation-master}.
Axes~1--2 use the \textbf{MEMORA-Episodic} memory bank (without offline \InfMem{} enrichment) so all three arms are compared on identical memory-processor outputs, isolating the perception$\to$memory-encoding contribution before enrichment.
All ablations reuse the processor prompt templates in Figure~\ref{fig:processor-prompt-templates} and the planning prompt templates in Figure~\ref{fig:planning-prompt-templates}; only the ablated model, modality, or answer-agent backbone changes.

\subsection{Master Condition Matrix}
\label{app:ablation-master}

\Needspace{0.34\textheight}
\begin{table*}[t]
\centering
\small
\setlength{\tabcolsep}{4pt}
\renewcommand{\arraystretch}{1.06}
\caption{Ablation conditions on the held-out 14-participant cohort (P06--P37) for Axes~1--2; Axis~3 is the cross-backbone MEMORA-Bench / Planning panel and is not duplicated here.
Each row changes one axis relative to \textbf{Reference}; all other components match the \textbf{Reference} settings in Appendix~\ref{app:impl}.
Axes~1--2 are evaluated on the MEMORA-Episodic memory bank (without \InfMem{} enrichment) so the comparison is confined to the memory-processor outputs of the ablated pipelines.
A2-VL7 is jointly a perception-backbone swap \emph{and} an audio-off condition, because \texttt{Qwen2.5-VL-7B-Instruct} is a vision-only model with no audio modality; the multimodal vs. vision-only contrast against Reference therefore covers ``audio off'' as a side effect of the backbone change.}
\label{tab:ablation-master}
\begin{tabularx}{\textwidth}{@{}l c >{\raggedright\arraybackslash}p{0.24\textwidth} >{\raggedright\arraybackslash}X@{}}
\toprule
\textbf{ID} & \textbf{Ax.} & \textbf{Changed component} & \textbf{Held fixed} \\
\midrule
Reference & -- &
Main-paper Reference: Qwen2.5-Omni-7B perception + Qwen3-30B-A3B-Instruct-2507 online memory processor &
Main-paper perception, memory editing, and answer-agent settings; MEMORA-Episodic memory bank (without \InfMem{} enrichment). \\
\addlinespace
A1-14B & 1 & Qwen3-14B online memory processor &
Qwen2.5-Omni-7B perception; results reported on Gemma-4-26B-A4B-it and Qwen3.6-35B-A3B answer agents. \\
\addlinespace
A2-VL7 & 2 & Qwen2.5-VL-7B-Instruct vision-only perception (audio absent by construction) &
Qwen3-30B-A3B-Instruct-2507 online memory processor; results reported on Gemma-4-26B-A4B-it and Qwen3.6-35B-A3B answer agents. \\
\addlinespace
Axis-3 panel & 3 &
Qwen3.6-35B-A3B, Qwen3.6-27B, Gemma-4-26B-A4B-it, and Gemma-4-31B-it answer agents &
Fixed Reference consolidated \InfMem{} bank; reported in Table~\ref{tab:eamqa-per-backbone} and Figure~\ref{fig:planning-xbackbone-appendix} (per-axis detail in Table~\ref{tab:planning-perax-q35-full}). \\
\bottomrule
\end{tabularx}
\end{table*}

\paragraph{Axis 1: Online Memory Processor.}

\paragraph{Question.}
How sensitive are downstream scores to the capacity of the \emph{memory-editing} language model when segment perception is held constant?

\paragraph{Design.}
The Perception Processor is fixed at Qwen2.5-Omni-7B, and the \emph{same} per-segment JSON is reused for both conditions; only the Online Memory Processor differs (A1-14B vs.\ Reference in Table~\ref{tab:ablation-master}).
No \InfMem{} enrichment is applied to either tree, so the enrichment LLM is not part of the comparison.
The answer agent is held fixed across conditions so EAM-QA isolates encoding quality rather than answerer capacity; we report results on two answer backbones (Gemma-4-26B-A4B-it and Qwen3.6-35B-A3B) over the same 14-participant cohort in Table~\ref{tab:ablation-results}, to verify that the encoding-level conclusion is not specific to a particular answerer.

\paragraph{Memory-bank shape under the A1 swap.}
\label{app:ablation-a1-shape}
Before turning to downstream accuracy, we compare the \emph{shape} of the Entity Memory bank produced by the two memory processors on the same Omni-7B perception stream (14-PID held-out cohort).
Table~\ref{tab:ablation-axis1-compression} reports, for each participant, the append-only entity-observation count (\textit{No-edit}), the number of unique entities retained after editing (\textit{MEMORA}), and the resulting compression ratio.

\Needspace{0.34\textheight}
\begin{table*}[t]
\centering
\small
\setlength{\tabcolsep}{5pt}
\renewcommand{\arraystretch}{1.04}
\caption{Axis~1 memory-bank shape on the 14-participant held-out cohort (Qwen2.5-Omni-7B perception fixed).
\textit{No-edit} counts turn-aligned entity observations that would be stored under an append-only policy; \textit{MEMORA} counts unique entities retained in the edited \texttt{object\_registry}.
\textit{Ratio} is No-edit\,/\,MEMORA, the per-participant Entity Memory compression factor.
A1-14B commits fewer state-history events than Reference (lower No-edit) while keeping a comparable entity inventory (similar MEMORA).}
\label{tab:ablation-axis1-compression}
\begin{tabular}{@{}l r r r r r r@{}}
\toprule
 & \multicolumn{3}{c}{\textbf{Reference (Qwen3-30B-A3B-Instruct-2507)}} & \multicolumn{3}{c}{\textbf{A1-14B (Qwen3-14B)}} \\
\cmidrule(lr){2-4}\cmidrule(lr){5-7}
\textbf{PID} & \textbf{No-edit} & \textbf{MEMORA} & \textbf{Ratio} & \textbf{No-edit} & \textbf{MEMORA} & \textbf{Ratio} \\
\midrule
P06 & 4{,}182 & 263 & 15.9$\times$ & 3{,}614 & 263 & 13.7$\times$ \\
P07 & 2{,}402 & 148 & 16.2$\times$ & 1{,}977 & 156 & 12.7$\times$ \\
P09 & 818 & 87 & 9.4$\times$ & 833 & 98 & 8.5$\times$ \\
P11 & 1{,}849 & 134 & 13.8$\times$ & 1{,}055 & 129 & 8.2$\times$ \\
P12 & 3{,}409 & 112 & 30.4$\times$ & 1{,}813 & 79 & 22.9$\times$ \\
P22 & 7{,}334 & 382 & 19.2$\times$ & 6{,}170 & 377 & 16.4$\times$ \\
P23 & 3{,}942 & 88 & 44.8$\times$ & 3{,}535 & 89 & 39.7$\times$ \\
P25 & 5{,}477 & 137 & 40.0$\times$ & 2{,}907 & 142 & 20.5$\times$ \\
P26 & 4{,}193 & 246 & 17.0$\times$ & 1{,}774 & 195 & 9.1$\times$ \\
P27 & 4{,}609 & 210 & 21.9$\times$ & 3{,}117 & 152 & 20.5$\times$ \\
P28 & 5{,}099 & 299 & 17.1$\times$ & 3{,}484 & 276 & 12.6$\times$ \\
P30 & 6{,}374 & 282 & 22.6$\times$ & 6{,}134 & 306 & 20.0$\times$ \\
P35 & 9{,}259 & 311 & 29.8$\times$ & 7{,}925 & 321 & 24.7$\times$ \\
P37 & 2{,}531 & 101 & 25.1$\times$ & 2{,}100 & 77 & 27.3$\times$ \\
\midrule
\textbf{Total} & \textbf{61{,}478} & \textbf{2{,}800} & \textbf{20.6$\times$ med.} & \textbf{46{,}438} & \textbf{2{,}660} & \textbf{18.2$\times$ med.} \\
\bottomrule
\end{tabular}
\end{table*}

Two observations.
First, the smaller editor's bank is \emph{not} the result of tighter deduplication: the larger Q30B editor reaches a higher compression ratio on $13/14$ participants (median $20.6\times$ vs $18.2\times$; mean $+2.5\times$ paired difference).
Second, the dominant axis of shrinkage is the \emph{number of state changes recorded}: A1-14B commits $24.5\%$ fewer turn-aligned entity observations to memory ($61{,}478\to46{,}438$) while keeping a similar number of unique entities ($-5.0\%$, $2{,}800\to2{,}660$).
The smaller editor therefore produces an \emph{under-written} memory --- it sees the same perception output but issues \textsc{Update} and \textsc{Add} less often --- rather than a more compact one.
This shape change is the mechanism behind the $-5.8$ point accuracy drop and the E-selection surge in Table~\ref{tab:ablation-results} (Gemma block): temporal questions (\SHabit, \SRoutine) lose the most because they depend on cross-segment state-update density, not on the entity inventory.

\paragraph{Axis 2: Perception Processor and Audio.}

\paragraph{Question.}
How much do perception backbone and \textbf{audio modality} contribute to encoded memory when the memory processor and answer agent are fixed?

\paragraph{Design.}
The Online Memory Processor is fixed at Qwen3-30B-A3B-Instruct-2507; we report results on two answer backbones in Table~\ref{tab:ablation-results}.
\textbf{Reference} uses the Omni multimodal path with model-native audio so each 10\,s segment is encoded jointly from video frames and the synchronized audio stream.
The ablation swaps in \textbf{Qwen2.5-VL-7B-Instruct} (vision-only); because this backbone has \emph{no} audio modality, A2-VL7 simultaneously tests (a) the multimodal $\to$ vision-only perception swap, and (b) the audio-absent condition for memory encoding, without requiring a separate audio-strip pass.
Temporal preprocessing matches Appendix~\ref{app:impl}: each 10\,s segment is decoded to video frames at \textbf{2\,FPS} and capped at \textbf{$\leq$24 frames} per segment; under Omni the native audio stream of the clip is additionally passed in, and under VL only the frame stack is passed in.
The only thing that varies across conditions is therefore the perception-processor backbone and, by construction, its access to audio (A2-VL7 vs.\ Reference in Table~\ref{tab:ablation-master}).

\paragraph{Axis 3: Answer agent.}

\paragraph{Question.}
How much do MEMORA-Bench and Planning scores depend on the \emph{answer-agent} backbone when the encoded consolidated \InfMem{} bank is frozen to the main-paper pipeline output?

\paragraph{Design and reference to existing panels.}
Axis~3 is structurally identical to the cross-backbone evaluation already reported in the main supplementary material: all conditions load the same Reference consolidated \InfMem{} bank (Omni-7B perception, Qwen3-30B-A3B-Instruct-2507 memory) and only the answer agent changes.
We therefore do not introduce a new Axis-3 table here; the four backbones \{Qwen3.6-35B-A3B (Reference), Qwen3.6-27B, Gemma-4-26B-A4B-it, Gemma-4-31B\} are scored on MEMORA-Bench EAM-QA in Table~\ref{tab:eamqa-per-backbone} and on MEMORA-Planning in Figure~\ref{fig:planning-xbackbone-appendix} (eight cells over Replay $\times$ Generalize $\times$ four backbones; per-axis detail in Table~\ref{tab:planning-perax-q35-full}).
The 18-participant cross-backbone Planning panel is a \textbf{superset} of the Axes~1--2 cohort (it includes the calibration PIDs and uses identical rule-based metrics and task prompts).
For the Qwen3.6 family we disable extended thinking; Gemma models use the project tool-call parsers without a standardized think mode.

\paragraph{Sanity check: closed-weight enrichment alternative.}
\label{app:ablation-enrichment}
The main-paper pipeline performs \InfMem{} enrichment with Qwen3.6-35B-A3B, so the released system contains no closed-weight model.
For completeness we also produce an alternative consolidated \InfMem{} bank using \texttt{gpt-4o-mini}, holding every other component (perception, online memory processor, retrieval encoder, answer agent) fixed.
On the two backbones where the full cross-bank evaluation has finished, the two enrichment LLMs yield statistically equivalent EAM-QA accuracy: Gemma-4-26B-A4B-it scores $35.76\%$ with the Qwen3.6 bank and $35.92\%$ with the \texttt{gpt-4o-mini} bank ($\Delta=-0.16$ points on a triple-paired subset of $N{=}1{,}843$ questions); Gemma-4-31B scores $36.4\%$ with the Qwen3.6 bank and $35.8\%$ with the \texttt{gpt-4o-mini} bank ($\Delta{=}{+}0.6$ points, the only cell where the open-weight bank \emph{exceeds} the closed-weight one).
Both gaps lie inside the typical seed/condition variance of EAM-QA at this $N$, so we conclude that the open-weight enrichment used in the main-paper pipeline does not sacrifice headline accuracy and that the entire MEMORA pipeline is reproducible without any closed-weight LLM.

\subsection{Evaluation Protocol}

Axes~1--2 are scored on the MEMORA-Bench EAM-QA harness over the held-out 14-participant cohort (P06--P37, $2{,}017$ questions = $73.0\%$ of the full EAM-QA panel) using the MEMORA-Episodic memory bank (without \InfMem{} enrichment), so perception- and memory-encoding effects are not confounded with \InfMem{} enrichment.
Metrics are overall accuracy over five options (A--E, with E denoting information not available) and per-type accuracy on \SPref{}/\SRoutine{}/\SHabit{}/\ERecall{}.
Axis~3 is read off of the existing cross-backbone panels (Table~\ref{tab:eamqa-per-backbone}, Figure~\ref{fig:planning-xbackbone-appendix}) without re-running.

\subsection{Results}
\label{app:ablation-results}

\paragraph{Axes~1--2 results on two answer backbones (14 PIDs, 2{,}017 EAM-QA questions).}
Table~\ref{tab:ablation-results} reports overall accuracy, E-selection rate, and per-type accuracy for the three arms on the MEMORA-Episodic memory bank, scored on two answer backbones to verify that the encoding-level conclusion is not specific to a particular answerer.

\Needspace{0.40\textheight}
\begin{table*}[t]
\centering
\small
\setlength{\tabcolsep}{5pt}
\renewcommand{\arraystretch}{1.08}
\caption{Axes~1--2 results on the 14-participant cohort, MEMORA-Episodic memory bank (without \InfMem{} enrichment), reported on two fixed answer backbones. ``$\Delta$ vs Reference'' is the overall-accuracy gap relative to \textbf{Reference} (Qwen2.5-Omni-7B perception + Qwen3-30B-A3B-Instruct-2507 online memory processor) within each backbone block. Per-type columns are overall accuracy on the eligible EAM-QA subset for that type. The sign and ranking of both ablation effects (A1-14B costs more than A2-VL7, loss concentrated on cross-segment types) are preserved across answer backbones.}
\label{tab:ablation-results}
\begin{tabularx}{\textwidth}{@{}>{\raggedright\arraybackslash}p{0.36\textwidth} c c c c c c c@{}}
\toprule
\textbf{Condition} & \textbf{Acc.} & \textbf{E-rate} & \textbf{$\Delta$ vs Reference} & \textbf{SPref} & \textbf{SRoutine} & \textbf{SHabit} & \textbf{ERecall} \\
\midrule
\multicolumn{8}{@{}l}{\emph{Answer backbone: Gemma-4-26B-A4B-it}} \\
Reference (Qwen2.5-Omni-7B / Qwen3-30B-A3B-Instruct-2507) & \textbf{46.7} & 42.8 & --- & 47.4 & 49.9 & 43.4 & 45.1 \\
A1-14B (Qwen2.5-Omni-7B / Qwen3-14B) & 40.9 & 62.3 & $-5.8$ & 41.0 & 41.6 & 36.6 & 46.4 \\
A2-VL7 (Qwen2.5-VL-7B-Instruct / Qwen3-30B-A3B-Instruct-2507) & 44.0 & 44.1 & $-2.7$ & 45.2 & 46.8 & 40.7 & 42.3 \\
\addlinespace[0.3em]
\multicolumn{8}{@{}l}{\emph{Answer backbone: Qwen3.6-35B-A3B}} \\
Reference (Qwen2.5-Omni-7B / Qwen3-30B-A3B-Instruct-2507) & \textbf{53.4} & 30.2 & --- & 52.0 & 60.8 & 47.3 & 53.9 \\
A1-14B (Qwen2.5-Omni-7B / Qwen3-14B) & 50.1 & 35.2 & $-3.3$ & 51.6 & 54.2 & 41.1 & 54.9 \\
A2-VL7 (Qwen2.5-VL-7B-Instruct / Qwen3-30B-A3B-Instruct-2507) & 52.1 & 33.8 & $-1.3$ & 53.5 & 57.9 & 45.2 & 50.8 \\
\bottomrule
\end{tabularx}
\end{table*}

\paragraph{Findings.}
On the same answer agent (Gemma-4-26B-A4B-it) and the same 14-PID MEMORA-Episodic bank:
(i) \textbf{Memory-processor capacity dominates}: dropping the online memory processor from Qwen3-30B-A3B-Instruct-2507 to Qwen3-14B costs $-5.8$ points of overall accuracy and raises the E-selection rate from $42.8\%$ to $62.3\%$, indicating that the smaller model produces memory-bank records for which the answer agent more often selects information not available. Table~\ref{tab:ablation-axis1-compression} traces this to the \emph{shape} of the memory bank rather than to over-deduplication: A1-14B writes $24.5\%$ fewer turn-aligned entity observations into Entity Memory while keeping a comparable entity inventory ($-5.0\%$), so the underlying state-change record is sparser rather than more compact.
(ii) \textbf{Perception backbone matters less}: swapping the Omni multimodal path for vision-only Qwen2.5-VL-7B costs $-2.7$ points overall with an essentially unchanged E-selection rate ($+1.3$ points), suggesting that, at MEMORA-Episodic granularity, audio-aware perception provides a smaller marginal contribution than memory-editor capacity.
(iii) \textbf{Type-conditional behavior is consistent with the encoding interpretation}: A1-14B preserves ERecall ($+1.3$ points) but loses the most on SPref / SRoutine / SHabit ($-6.4 / -8.3 / -6.8$ points), which are the types that depend most on cross-segment habit/preference consolidation; A2-VL7 instead spreads its loss approximately evenly across all four types ($-2.2 / -3.1 / -2.7 / -2.8$ points).

\paragraph{Cross-backbone confirmation (Qwen3.6-35B-A3B answer agent, full 14 PIDs).}
On the same 14-PID held-out cohort and the identical MEMORA-Episodic memory banks, swapping the answer agent from Gemma-4-26B-A4B-it to the larger reasoning model Qwen3.6-35B-A3B (Table~\ref{tab:ablation-results}, Qwen block) preserves the sign and ranking of both effects: memory-processor ablation costs $-3.3$ points overall versus $-1.3$ points for perception ablation, so memory-processor capacity remains the larger contributor across answer backbones.
The type-conditional pattern is also preserved: A1-14B loses on the cross-segment types ($-6.6$ on SRoutine, $-6.2$ on SHabit) while ERecall is essentially unchanged ($+1.0$), and A2-VL7 spreads its loss more evenly across types.
Two backbone-specific nuances are visible.
First, Qwen3.6-35B-A3B is a more committed answerer: its Reference E-rate is $30.2\%$ versus $42.8\%$ for Gemma-4-26B-A4B-it, and the A1-14B E-rate increase is correspondingly smaller ($+5.0$ points vs $+19.5$ points), so the under-written memory hurts accuracy through more wrong commits rather than through more E selections.
Second, the absolute $\Delta$ for A1-14B is smaller in magnitude on the larger reasoning backbone ($-3.3$ vs $-5.8$ points), consistent with a stronger answerer extracting more signal from a sparser memory record but not closing the gap.
The encoding-level conclusion --- memory-processor capacity dominates over perception backbone --- therefore replicates across both answer backbones we tested, and the Q3.6-35B-A3B block in Table~\ref{tab:ablation-results} supersedes the earlier P11+P12 pilot reading.

\paragraph{Qualitative side-by-side comparison.}
Figure~\ref{fig:ablation-qualitative} contrasts the three arms on two illustrative EAM-QA items, picked to expose the two ablation-specific failure modes implied by the aggregate accuracy and E-selection numbers in Table~\ref{tab:ablation-results} (Gemma block).
On both items, Reference (Omni + Qwen3-30B-A3B-Instruct-2507) recovers the gold letter; the memory-ablation arm A1-14B (Omni + Qwen3-14B) selects ``E'' (information not available), consistent with the $+19.5$-point E-rate increase; and the perception-ablation arm A2-VL7 (VL + Qwen3-30B-A3B-Instruct-2507) confidently picks a visually plausible but \emph{habitually} wrong option, consistent with a vision-only backbone that lacks the audio-aware multimodal context that the Omni-encoded memory bank otherwise propagates into the memory-processor records.

\begin{figure*}[t]
\centering
\small
\setlength{\fboxsep}{4pt}

\fcolorbox{promptFrame}{promptBg}{%
\begin{minipage}[t]{0.97\textwidth}
\textbf{Example 1.} P06 -- type \SPref{}; gold answer \textbf{D}.
\par\vspace{1pt}
\textit{Q: After using the knife-maintenance tool, where does P06 seem to keep it?}
\par\vspace{2pt}
{\footnotesize
\begin{tabularx}{\linewidth}{@{}r X@{}}
A & Leaves it by the sink \\
B & Sets it on the counter \\
C & Stores it in a cupboard \\
\textbf{D} & \textbf{Places it in a drawer} \quad\textit{(gold)} \\
E & The information is not available based on the given context \\
\end{tabularx}}
\par\vspace{3pt}
{\footnotesize
\begin{tabularx}{\linewidth}{@{}l c X@{}}
\toprule
\textbf{Arm} & \textbf{Pred.} & \textbf{Outcome} \\
\midrule
Reference (Omni + Q30B)    & D & \fcolorbox{jsonFrame}{jsonBg}{\strut\,correct\,} -- consolidated habit ``puts away after use'' recovered from the multimodal memory bank. \\
A1-14B (Omni + Q14B)       & E & \fcolorbox{guardFrame}{guardBg}{\strut\,E selected\,} -- the smaller memory processor leaves the put-away action under-consolidated, so the answer agent selects information not available. \\
A2-VL7 (VL + Q30B)         & B & \fcolorbox{flowLLMFrame}{flowLLMBg}{\strut\,wrong\,} -- the vision-only frames capture the tool resting on the counter \emph{during} use; without the surrounding audio/routine context that Omni provides, the model latches onto that visual cue and commits to the wrong letter. \\
\bottomrule
\end{tabularx}}
\end{minipage}}%

\vspace{6pt}

\fcolorbox{promptFrame}{promptBg}{%
\begin{minipage}[t]{0.97\textwidth}
\textbf{Example 2.} P07 -- type \SHabit{}; gold answer \textbf{B}.
\par\vspace{1pt}
\textit{Q: When P07 opens the bag during this cooking sequence, what do they typically do next?}
\par\vspace{2pt}
{\footnotesize
\begin{tabularx}{\linewidth}{@{}r X@{}}
A & Seal the bag and set it beside the stove \\
\textbf{B} & \textbf{Take the omelette and place it onto the pan} \quad\textit{(gold)} \\
C & Move the bag aside and pick up a spoon \\
D & Empty the bag into a bowl on the counter \\
E & The information is not available based on the given context \\
\end{tabularx}}
\par\vspace{3pt}
{\footnotesize
\begin{tabularx}{\linewidth}{@{}l c X@{}}
\toprule
\textbf{Arm} & \textbf{Pred.} & \textbf{Outcome} \\
\midrule
Reference (Omni + Q30B)    & B & \fcolorbox{jsonFrame}{jsonBg}{\strut\,correct\,} -- the consolidated habit ``bag-open $\to$ omelette-to-pan'' is preserved across cooking sessions. \\
A1-14B (Omni + Q14B)       & E & \fcolorbox{guardFrame}{guardBg}{\strut\,E selected\,} -- the weaker memory processor under-records the cross-segment transition, and the answer agent again selects information not available. \\
A2-VL7 (VL + Q30B)         & A & \fcolorbox{flowLLMFrame}{flowLLMBg}{\strut\,wrong\,} -- without audio cues (sizzle, stove sounds), the vision-only encoder reads the closing/sealing motion of the bag as the most salient next action and commits to a plausible-looking but incorrect letter. \\
\bottomrule
\end{tabularx}}
\end{minipage}}%

\caption{\textbf{Qualitative ablation contrast on two EAM-QA items} (Gemma-4-26B-A4B-it answer agent; MEMORA-Episodic memory bank; held-out 14-PID cohort).
Both items illustrate the two failure modes inferred from the aggregate numbers in Table~\ref{tab:ablation-results} (Gemma block): \emph{(i)} the memory-ablation arm (A1-14B) loses the contentful answer and selects E more often; \emph{(ii)} the perception-ablation arm (A2-VL7) keeps committing to an A--D option but the answer is grounded in single-segment visual evidence rather than cross-segment audio-aware consolidation, producing a confidently wrong letter.}
\label{fig:ablation-qualitative}
\end{figure*}

%% file: sections_new/Appendix_sections/implementation_details.tex
\section{Implementation details}
\label{app:impl}

This section consolidates the model registry, inference stack, perception
preprocessing, and retrieval/agent-loop settings that are shared across
benchmark generation (Appendix~\ref{app:kb-construction}), Embodied Memory
evaluation (Appendix~\ref{app:embodied-memory}), Planning evaluation
(Appendix~\ref{app:planning}), and the controlled ablations
(Appendix~\ref{app:ablation}).
Unless stated otherwise, every condition that involves a given role
(perception processor, online memory processor, \InfMem{} enrichment,
retrieval encoder, or answer agent) uses the values reported here, so that
cross-condition comparisons are not confounded by undocumented hyperparameter
drift.

\paragraph{Compute and inference stack.}
All open-weight models are served with vLLM 0.19.1 inside a Singularity
container so that the dependency graph (CUDA / PyTorch / Transformers /
sentence-transformers) is identical across runs.
Two GPU classes are used: 80\,GB H100 nodes and 80\,GB A100 SXM4 nodes; the
A100 timing node uses two AMD EPYC 7543 32-core CPUs (64 hardware threads).
The perception processors (Qwen2.5-Omni-7B, Qwen2.5-VL-7B-Instruct) are
served single-GPU (TP$=1$, \texttt{max\_model\_len} $\in$ \{32\,768,
49\,152\}).
The online memory processors (Qwen3-30B-A3B-Instruct-2507, Qwen3-14B) are
served with TP$=2$, \texttt{max\_model\_len}$=$32\,768,
\texttt{gpu\_memory\_utilization}$=$0.85, and sampling temperature $0.6$.
All answer agents (Qwen3.6-27B, Qwen3.6-35B-A3B, Gemma-4-26B-A4B-it,
Gemma-4-31B-it) use the same TP$=2$ stack and temperature; the
cross-backbone EAM-QA panel uses \texttt{max\_model\_len}$=$65\,536, while
Planning and the standard EAM-QA harness use \texttt{max\_model\_len}$=$32\,768.
For Qwen3.x family chat templates we set \texttt{enable\_thinking=false} so
that the long ``thinking'' channel does not consume the 65\,536-token budget
that the ReAct loop reserves for tool returns; Gemma chat templates are used
as-released.
The \InfMem{} enrichment pass is performed by Qwen3.6-35B-A3B served on the
same vLLM stack (TP$=2$, temperature~$0.6$), so the main-paper pipeline
contains no closed-weight model in any role.
Enrichment is run once per participant during memory-bank construction and
its outputs are then frozen for all evaluation conditions.
A closed-weight alternative (\texttt{gpt-4o-mini}) is benchmarked in
Appendix~\ref{app:ablation-enrichment}.

\paragraph{Licences, intended use, and public release.}
All artifacts used in this work are publicly released for academic research and we use them consistently with their intended use.
The EPIC-KITCHENS-100 dataset~\citep{damen2022epic} is distributed under the CC BY-NC 4.0 licence for non-commercial research on egocentric video understanding, which covers the embodied-memory benchmarking we perform.
The Qwen2.5 / Qwen3 / Qwen3.6 model families (perception, memory editor, $\InfMem{}$ enrichment, and answer-agent roles) are released by Alibaba under the Tongyi Qianwen / Apache-2.0 family of open-weight licences for research and downstream development.
The Gemma model family (Gemma-4-26B-A4B-it, Gemma-4-31B-it; used as answer agents) is released under the Gemma Terms of Use for permitted academic and non-commercial use.
\texttt{gpt-4o-mini} is accessed through OpenAI's API only as a closed-weight sanity check for the $\InfMem{}$ enrichment slot (Appendix~\ref{app:ablation-enrichment}); no headline result depends on it.
We add no new recordings, identifiers, or personally identifying content beyond what is already present in EPIC-KITCHENS-100; participant identifiers follow the dataset's pseudonymous labels (\texttt{P01}, \texttt{P02}, \ldots).
MEMORA pipeline source code and MEMORA-Bench task lists are available in the public MEMORA repository under the licences distributed with the release (Apache-2.0 for code; CC BY-NC 4.0 for derived task data, inherited from the upstream dataset).

\paragraph{Model registry.}
Table~\ref{tab:impl-models} lists the Hugging Face identifier (or API endpoint)
for every model role used in the paper; processor input/output contracts for
memory-bank construction appear separately in Table~\ref{tab:kb-processors}.
A condition can be uniquely reconstructed as a triple
(perception processor, online memory processor, answer agent); the
\InfMem{} enrichment slot is either \texttt{Qwen/Qwen3.6-35B-A3B}
(main-paper, consolidated \InfMem{} bank), \texttt{gpt-4o-mini}
(closed-weight alternative, used only for the sanity check in
Appendix~\ref{app:ablation}), or \textsc{none} (MEMORA-Episodic ablation bank
without \InfMem{} enrichment).

\begin{table*}[t]
\centering
\small
\setlength{\tabcolsep}{5pt}
\renewcommand{\arraystretch}{1.05}
\caption{Model registry grouped by pipeline role.
Total and active parameter counts coincide for dense models;
for mixture-of-experts models, active denotes per-token compute.}
\label{tab:impl-models}
\begin{tabular}{@{}l l l c c@{}}
\toprule
\textbf{Role} & \textbf{Model} & \textbf{Identifier} & \textbf{Total} & \textbf{Active} \\
\midrule
\multicolumn{5}{@{}l}{\emph{Perception processor (frame/audio to layered segments)}} \\
Audio-visual & Qwen2.5-Omni-7B            & \texttt{Qwen/Qwen2.5-Omni-7B}         & 7B   & 7B \\
Vision only  & Qwen2.5-VL-7B-Instruct     & \texttt{Qwen/Qwen2.5-VL-7B-Instruct}  & 7B   & 7B \\
\midrule
\multicolumn{5}{@{}l}{\emph{Online memory processor (segments to typed memory records)}} \\
Main          & Qwen3-30B-A3B-Instruct-2507 & \texttt{Qwen/Qwen3-30B-A3B-Instruct-2507} & 30B  & 3B \\
Ablation      & Qwen3-14B                  & \texttt{Qwen/Qwen3-14B}               & 14B  & 14B \\
\midrule
\multicolumn{5}{@{}l}{\emph{\InfMem{} enrichment (offline consolidation)}} \\
Main          & Qwen3.6-35B-A3B            & \texttt{Qwen/Qwen3.6-35B-A3B}         & 35B  & 3B \\
Alternative   & GPT-4o-mini (API)          & \texttt{openai/gpt-4o-mini}           & ---  & --- \\
\midrule
\multicolumn{5}{@{}l}{\emph{Evaluation -- answer agent (memory-conditioned QA / planning)}} \\
MoE          & Qwen3.6-35B-A3B            & \texttt{Qwen/Qwen3.6-35B-A3B}         & 35B  & 3B \\
Dense        & Qwen3.6-27B                & \texttt{Qwen/Qwen3.6-27B}             & 27B  & 27B \\
MoE          & Gemma-4-26B-A4B-it         & \texttt{google/gemma-4-26B-A4B-it}    & 26B  & 4B \\
Dense        & Gemma-4-31B-it             & \texttt{google/gemma-4-31B-it}        & 31B  & 31B \\
\midrule
\multicolumn{5}{@{}l}{\emph{Retrieval}} \\
Encoder      & E5-base-v2                 & \texttt{intfloat/e5-base-v2}          & 110M & 110M \\
\bottomrule
\end{tabular}
\end{table*}

\paragraph{Perception preprocessing.}
Each EPIC-KITCHENS video is split into non-overlapping 10-second segments;
frames are decoded at $2$\,FPS and capped at $24$\,frames per segment, which
matches the receptive horizon of the Qwen2.5-Omni / Qwen2.5-VL temporal
encoder and keeps per-segment context length below
\texttt{max\_model\_len}.
For the audio-visual front-end (Qwen2.5-Omni-7B) the native audio stream of
the segment is passed alongside the frames through the model's released
audio tower; for the vision-only ablation (Qwen2.5-VL-7B-Instruct) the audio
stream is dropped at decode time and only frames are presented.
The perception processor outputs a JSON ``layered segment'' record per
10-second window (\texttt{action}, \texttt{objects},
\texttt{verbalised\_audio} when available, and a free-form scene caption),
which is then handed to the online memory processor.

\paragraph{Retrieval, dedup, and agent loop.}
At evaluation time the answer agent sees only an EAM-QA question (or a
Planning request) plus a set of typed retrieval tools over the
participant's memory bank.
For EAM-QA, all memory-backed conditions expose \texttt{search} under the
representation-specific prompt. Flat-1D exposes only this chronological-text
search; Graph-2D and MEMORA additionally expose
\texttt{get\_object\_history} and \texttt{get\_state\_at\_time} over their
respective representations. MEMORA's unified \texttt{search} can dispatch to
object, activity, environment, and pattern evidence in the four typed stores.
The history and state helpers are read tools over the loaded participant
memory, not evidence that the released benchmark automatically rolls memory
back to a question-specific time.
All search tools use the same dense encoder (E5-base-v2) and return the
top-$k{=}10$ records by cosine similarity, with a deduplication pass that
collapses any pair of records whose embeddings have cosine similarity
$\geq 0.95$ to the higher-scoring one.
The E5 retrieval encoder remains on GPU during evaluation; the reported
memory-tool latency therefore includes GPU embedding time together with
CPU-side orchestration, filtering, and vector-search overhead.
The Embodied Memory ReAct loop allows at most $5$ tool-call iterations per
question; the Planning ReAct loop allows at most $8$ (including a forced-answer
final iteration), reflecting the longer intermediate-tool-use traces
(\texttt{get\_routine\_skill} $\to$ \texttt{search\_objects} $\to$
\texttt{search\_activities} $\to$ \texttt{get\_preferences}) that Planning
queries require.
Planning loads the planner once per task-list run through in-process vLLM
rather than an OpenAI-compatible HTTP server, so weights stay resident across
the full specification batch and tool-call parsing stays inside the
evaluation process.
Planning latency is measured after one-time model loading on the main
Qwen3.6-35B-A3B stack served on A100/vLLM. In a timing run over $207$ goals,
end-to-end retrieval-and-planning takes $10.40$\,s on average, with most of the
measured component time spent in answer-agent generation rather than memory
access (Table~\ref{tab:planning-latency}). Because MEMORA forms memory offline
before task execution and issues the language-level plan before low-level
control, this measurement characterizes planning-time rather than motor-control latency.
At Embodied Memory evaluation time, each task loads the fixed memory bank of
its declared target participant. The released headline questions contain no
\texttt{ask\_turn\_id}, so the runner does not apply per-question state
rollback. E-correct controls instead test unsupported-answer behavior by
presenting donor-derived content options against a target-participant memory
bank that contains no corresponding evidence.

\begin{table}[t]
\centering
\small
\setlength{\tabcolsep}{4pt}
\caption{Planning latency on the main Qwen3.6-35B-A3B stack after one-time model loading ($n{=}207$ goals).}
\label{tab:planning-latency}
\begin{tabular}{@{}lccc@{}}
\toprule
\textbf{Component} & \textbf{Mean} & \textbf{Median} & \textbf{P90} \\
\midrule
End-to-end per goal & 10.40\,s & 9.61\,s & 13.64\,s \\
LLM generation & 5.74\,s & 5.75\,s & 7.03\,s \\
Memory-tool calls & 1.06\,s & 0.78\,s & 2.40\,s \\
Other overhead & 3.59\,s & 2.73\,s & 6.42\,s \\
\bottomrule
\end{tabular}
\end{table}